\pdfoutput=1
\documentclass{article}

\usepackage[main, final]{neurips_2026}

\usepackage[utf8]{inputenc}
\usepackage[T1]{fontenc}
\usepackage{xcolor}

\definecolor{ourred}{HTML}{B22222}
\definecolor{ourgreen}{HTML}{2E8B57}
\definecolor{ourblue}{HTML}{0047AB}
\definecolor{ourorange}{HTML}{D35400}
\definecolor{ourlink}{HTML}{004A6C}
\colorlet{refcolor}{blue!50!black}

\usepackage{hyperref}
\hypersetup{
    colorlinks=true,
    linkcolor={refcolor},
    citecolor={refcolor},
    urlcolor={black},
}
\usepackage{url}
\usepackage{booktabs}
\usepackage{amsfonts}
\usepackage{nicefrac}
\usepackage{microtype}

\usepackage{graphicx}
\usepackage{tabularx}
\usepackage{arydshln}
\usepackage{multirow}
\usepackage{pifont}
\usepackage[labelfont=bf]{caption}
\usepackage{subcaption}
\usepackage{titletoc}
\usepackage{amsmath}
\usepackage{float}
\usepackage{enumitem}
\usepackage[most]{tcolorbox}
\usepackage[export]{adjustbox}
\usepackage[ruled,vlined]{algorithm2e}

\usepackage[capitalise,noabbrev,nameinlink]{cleveref}

\newcommand{\ulref}[1]{%
  \begingroup\setbox0\hbox{#1}%
  \rlap{\color{refcolor}\rule[-0.2em]{\wd0}{0.4pt}}\box0\endgroup}

\makeatletter
\renewcommand\NAT@hyper@[1]{%
  \hyper@natlinkstart{\@citeb\@extra@b@citeb}\ulref{#1}\hyper@natlinkend}
\def\hyper@natlinkbreak#1#2{#1}
\makeatother

\newcommand{\setrefstyle}[3]{%
  \crefformat{#1}{##2\ulref{#2~##1}##3}%
  \Crefformat{#1}{##2\ulref{#2~##1}##3}%
  \crefrangeformat{#1}{##3\ulref{#3~##1}##4\,--\,##5\ulref{##2}##6}%
  \Crefrangeformat{#1}{##3\ulref{#3~##1}##4\,--\,##5\ulref{##2}##6}%
  \crefmultiformat{#1}{##2\ulref{#3~##1}##3}{ and~##2\ulref{##1}##3}{, ##2\ulref{##1}##3}{, and~##2\ulref{##1}##3}%
  \Crefmultiformat{#1}{##2\ulref{#3~##1}##3}{ and~##2\ulref{##1}##3}{, ##2\ulref{##1}##3}{, and~##2\ulref{##1}##3}%
}

\setrefstyle{figure}{Figure}{Figures}
\setrefstyle{subfigure}{Figure}{Figures}
\setrefstyle{table}{Table}{Tables}
\setrefstyle{section}{Section}{Sections}
\setrefstyle{subsection}{Section}{Sections}
\setrefstyle{subsubsection}{Section}{Sections}
\setrefstyle{appendix}{Appendix}{Appendices}
\setrefstyle{subappendix}{Appendix}{Appendices}
\setrefstyle{algocf}{Algorithm}{Algorithms}

\crefformat{equation}{#2\ulref{Eq.~(#1)}#3}
\Crefformat{equation}{#2\ulref{Equation~(#1)}#3}

\setrefstyle{algorithm}{Algorithm}{Algorithms}
\setrefstyle{algocfline}{Line}{Lines}

\usepackage[normalem]{ulem}

\renewcommand{\ulref}[1]{\uline{#1}}

\newcolumntype{C}{>{\centering\arraybackslash}X}

\newcommand{\LINE}{\textup{\textsc{LINE}}}

\title{\LINE{}: LLM-based Iterative Neuron Explanations for Vision Models}

\author{%
  Vladimir Zaigrajew$^{1,2}$ \quad
  Michał Piechota$^{2,}$\thanks{Equal contribution.} \quad
  Gaspar Sekula$^{2,}$\footnotemark[1] \\
  \bfseries Paweł Gelar$^{1,2}$ \quad
  Przemysław Biecek$^{1,2,3}$ \\[4pt]
  \mdseries $^{1}$Centre for Credible AI \quad
            $^{2}$Warsaw University of Technology \quad
            $^{3}$University of Warsaw \\[4pt]
  \mdseries\small\texttt{\{vladimir.zaigrajew.dokt,pawel.gelar\}@pw.edu.pl} \\
}

\begin{document}

\maketitle

\begin{abstract}
  Interpreting the concepts encoded by individual neurons in deep neural networks is a crucial step towards understanding their complex decision-making processes and ensuring AI safety. Despite recent progress in neuron labeling, existing methods often limit the search space to predefined concept vocabularies or produce overly specific descriptions that fail to capture higher-order, global concepts. We introduce \LINE{}, a novel, training-free iterative approach tailored for open-vocabulary concept labeling in vision models.
  Operating in a \emph{strictly gray-box} setting, \LINE{} leverages a large language model and a text-to-image generator to iteratively propose and refine concepts in a closed loop, guided by activation history. We demonstrate that \LINE{} achieves state-of-the-art performance across multiple model architectures, yielding AUC improvements of up to 0.15 on ImageNet and 0.05 on Places365, while discovering, on average, 25\% of new concepts missed by predefined vocabularies. Beyond identifying the top concept, \LINE{} provides a complete generation history, enabling polysemanticity evaluation and producing supporting visual explanations that rival gradient-dependent activation maximization methods. The \LINE{} source code is available at {\hypersetup{allcolors=ourlink}{\url{https://github.com/CredibleAI/LINE}}}.
\end{abstract}

\section{Introduction}
\label{sec:introduction}

 Deep Neural Networks (DNNs) have been widely adopted across various domains due to their performance and capabilities. However, limited progress in comprehending their opaque decision-making processes remains a major obstacle to their use in safety-critical applications, such as healthcare or the justice system. To address this problem, numerous explainable AI (XAI) methods~\citep{Selvaraju_2017_ICCV,tjoa2020survey,holzinger2022explainable} have been developed over the years to uncover how these models ``think'' and decide. Further need for more precise control over model behavior led to the development of the field of mechanistic interpretability~\citep{bereska2024mechanistic}, which aims to surgically uncover the inner workings of DNNs, focusing on identifying localized decision-making circuits~\citep{conmy2023towards} or specific neuron activation patterns~\citep{feldhus2025interpreting}. A significant problem in this area is to automatically translate neural activity into human-understandable semantic concepts. While neuron labeling has seen success in both vision~\citep{bau2017CVPR,hernandez2022natural,kalibhat2023identifying,oikarinen2023clip,bykov2024labeling,bai2025interpreting} and text~\citep{bills2023language,kopf2025capturing} domains, crucial limitations remain. Foremost, most concept assignments rely on predefined vocabularies~\citep{bau2017CVPR,kalibhat2023identifying,oikarinen2023clip,bykov2024labeling} that often lack the precise ground-truth concept. Alternatively, methods that rely on text-generative models, such as MILAN~\citep{hernandez2022natural}, frequently struggle to capture broad concepts, producing overly crisp descriptions.

\begin{figure}[t]
    \centering
    \includegraphics[width=0.98\linewidth]{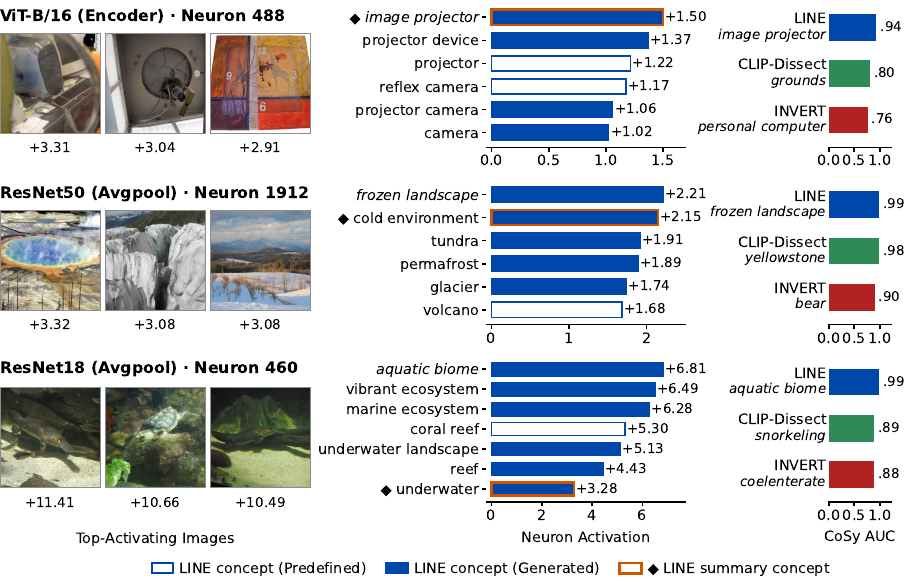}
    \caption{\textbf{\LINE{} neuron descriptions.} For three neurons from different models, we show (\textbf{from the left}): the top three activating images from the model-trained dataset, the top concepts \LINE{} registered in the final scoreboard with their origin, CoSy benchmark~\citep{kopf2024cosy} AUC scores for \LINE{} and the baseline methods. The images and the \LINE{} concepts are provided with their neuron activation scores. Extended results for different layers and models are in \Cref{app:extended_visual}.}
    \label{fig:main_qualitive}
\end{figure}

\textbf{This work: A novel LLM-based iterative pipeline for automated neuron labeling.}
We introduce \LINE{}, the first \textbf{L}LM-based \textbf{I}terative \textbf{N}euron \textbf{E}xplanation framework for providing textual explanations of vision models (see \Cref{fig:schema}, with corresponding visual results in \Cref{fig:main_qualitive}). \LINE{} operates in a strictly \textbf{training-free, gray-box \footnote{\LINE{} requires only access to the inspected neurons' activations, without access to the model weights or gradients.}} setting, leveraging the generative capabilities of text-to-image (T2I) models and the reasoning power of Large Language Models (LLMs). At its core, \LINE{} utilizes an LLM to propose new potential concepts' descriptions based on a neuron's concept scoreboard, initialized from a predefined vocabulary. For instance, in \Cref{app:single_loop}, while an initial step might incorrectly assign a \textit{pool table} label to a neuron, \LINE's iterative refinement reveals that the neuron activates more strongly to the concept \textit{weightlifting}. To connect LLMs with vision models and identify concept activations, we synthesize concepts via a T2I model and leverage the resulting images to extract neuron activations. Finally, these activations are evaluated by a scoring function to quantify the correctness of the proposed concept and update the scoreboard with the new concept. To overcome the limitation of overly specific concepts, an additional summary step is added at the end of the cycle to suggest and score a higher-order concept from the final history (e.g., highlighting a global concept like \textit{physical exercise}). Through its highly transparent framework, \LINE{} provides not only the top-scoring concept but also the generation history, the reasoning behind each new concept, and the generated visual samples, enabling polysemanticity evaluation. Furthermore, by optimizing textual descriptions, \LINE{} provides a gradient-free alternative to activation maximization approaches commonly used for bias discovery~\citep{pennisi2025diffexplainer,carnemolla2025dexter}.

\textbf{Contribution.} In this work, we present \LINE{}, a gray-box, training-free iterative pipeline for providing textual and visual explanations for vision model neurons. Our core contributions include:
\textbf{State-of-the-art Neuron Labeling:} Evaluated across four diverse CNN and ViT architectures, \LINE{} significantly outperforms prior methods on the CoSy neuron labeling benchmark~\citep{kopf2024cosy} and achieves the highest overall correlation score between label descriptions and image activations.
\textbf{Beyond Predefined Vocabularies:} We demonstrate that relying solely on predefined vocabularies leads to suboptimal label descriptions. Through the iterative refinement process, \LINE{} discovers up to $44\%$ new, highly human-interpretable concepts missed even by standard predefined vocabularies.
\textbf{Visual Explanations:} Compared to activation maximization methods, \LINE{} identifies analogous model biases without requiring access to the gradients, resulting in more interpretable and natural visuals than rival approaches.
We evaluate \LINE{} across three dimensions: \textbf{(1)} quantitative superiority on the CoSy benchmark across three architectures and three datasets, and top performance on the correlation evaluation, \textbf{(2)} qualitative and causal validation via image ablation, and \textbf{(3)} a comparison of visual explanations against activation maximization methods for bias discovery tasks. Demonstrating strong performance across these areas, \LINE{} offers a transparent and scalable approach for auditing DNNs.
In this way, \LINE{} supports the model science agenda \citep{biecek2025model} of treating trained models as objects of empirical study that can be explored and verified.

\section{Related work}
\label{sec:related_work}

\textbf{Neuron Labeling.}
Early methods for automatic neuron labeling, pioneered by Network Dissection~\citep{bau2017CVPR}, linked neurons to concepts by computing the Intersection over Union (IoU) between neuron activation maps and ground-truth segmentation masks. INVERT~\citep{bykov2024labeling} expanded this by incorporating compositional logic and optimizing for a higher Area Under the Curve (AUC). To decouple descriptions from rigid datasets, MILAN~\citep{hernandez2022natural} trained an LSTM-based captioning model on the human-annotated custom dataset (MILANNOTATIONS) to generate free-form descriptions. To further overcome the need for labeled corpora or segmentation masks like Broden~\citep{bau2017CVPR}, methods such as CLIP-Dissect~\citep{oikarinen2023clip}, SAND~\citep{srinivas2025sand}, and FALCON~\citep{kalibhat2023identifying} leverage multimodal models like CLIP~\citep{radford2021learning} to match top-activation images against large-scale, open-vocabulary captions. More recently, Describe-and-Dissect (DnD)~\citep{bai2025interpreting} introduced a training-free framework that shares the localization-based approach of FALCON and SAND. It uses an image-to-text (I2T) model to describe highly activating image crops, an LLM to summarize I2T captions into neuron labels, and a T2I model to synthesize them. While \LINE{} shares DnD's components, such as LLMs and T2I models, it fundamentally departs from standard single-forward-pass methods through an iterative refinement process. Evaluated on the recent, architecture-agnostic unified benchmark, Concept Synthesis (CoSy)~\citep{kopf2024cosy}, \LINE{} achieves superior performance. Finally, because neurons are often highly polysemantic~\citep{bricken2023monosemanticity}, \LINE{} enables rapid assessment of multi-faceted neuron behavior by maintaining a history of all evaluated concepts.

\textbf{Activation Maximization (AM).}
AM interprets neural networks by generating inputs that maximize targeted neuron activations~\citep{erhan2009visualizing,zhu2025representation}. While early methods relying on the direct optimization of input pixels often produced unrealistic, hard-to-interpret images~\citep{simonyan2013deep,olah2017feature}, recent approaches like DiffExplainer~\citep{pennisi2025diffexplainer} and later DEXTER~\citep{carnemolla2025dexter} have modernized AM by optimizing the text prompts for diffusion models that generate AM images. However, as these methods predominantly rely on gradient-based optimization, they impose significant memory overhead and require the entire pipeline to be fully differentiable, often limiting their architectural choices to older T2I models like Stable Diffusion 1.5~\citep{Rombach_2022_CVPR}. Since \LINE{} does not require differentiability, it can seamlessly integrate modern T2I models such as SDXL~\citep{podell2023sdxl} and FLUX.1[dev]~\citep{flux_dev_2024}. This flexibility enables the discovery of model biases by producing AM samples that are visually cleaner, more natural, and less out-of-distribution (OOD).

\textbf{Generative Models for Explainability.}
The integration of generative models has significantly advanced XAI by bridging visual and textual reasoning. The reasoning capabilities of LLMs have enabled their deployment as iterative optimizers in tasks that can be described in natural language~\citep{yang2023large} or in multimodal feedback loops for refining image caption descriptions~\citep{ashutosh2025llms}. Concurrently, generative image models have enabled the creation of visual counterfactuals~\citep{sobieski2024global,sobieski2024rethinking} and synthetic evaluation sets~\citep{kopf2024cosy}, thereby mitigating reliance on extensive human-labeled corpora. Recent advancements in agentic frameworks have introduced the first agent-based neuron interpretability pipelines, such as MAIA~\citep{shaham2024a} and OpenMAIA~\citep{camunas2025openmaia}, which equip multimodal LLMs with unconstrained tools for exemplar selection, synthetic generation, and image editing. While these systems create complex, less interpretable, multi-billion-parameter workflows, \LINE{} offers a simpler, highly transparent, and controllable architecture. It demonstrates that the iterative reasoning of small-scale LLMs (e.g., Llama 3.1 8B~\citep{grattafiori2024llama}), when paired only with a T2I generative model, achieves state-of-the-art performance on neuron labeling tasks.

\section{Method}
\label{sec:method}

\begin{figure}[t]
    \centering
    \includegraphics[width=1\linewidth]{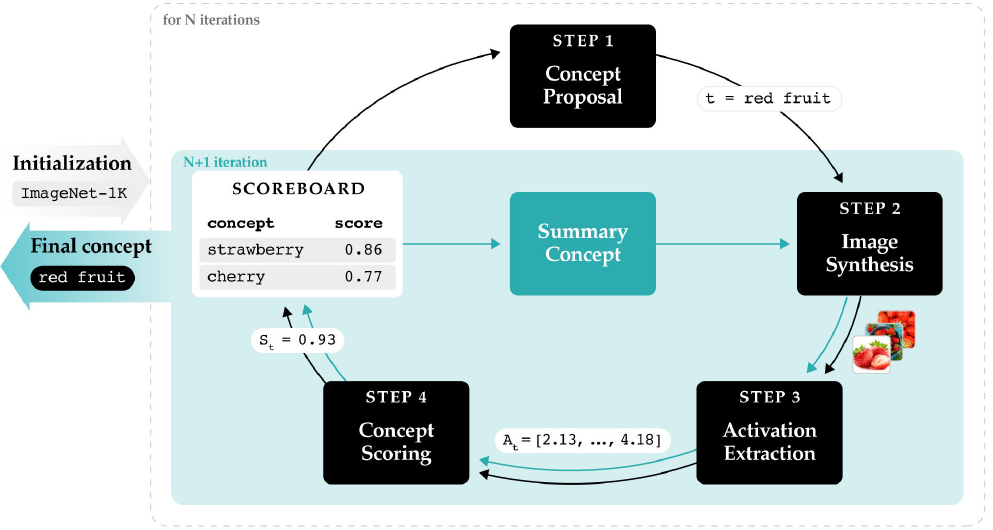}
    \caption{\textbf{Overview of the \LINE{} iterative framework.} \texttt{Step 1:} An LLM proposes a new concept $t$ based on the descriptions from the scoreboard $\mathcal{H}$ (e.g., generalizing \textit{strawberry} and \textit{cherry} into \textit{red fruit}). \texttt{Step 2:} A Text-to-Image (T2I) model generates a batch of diverse synthetic images illustrating the concept $t$. \texttt{Step 3:} These images are processed by the target vision model, extracting concept activations $A_t$. \texttt{Step 4:} A scoring function ($\psi_{avg}(A)$) converts $A_t$ into score $s_t$, and the $\mathcal{H}$ is updated with the result. Upon reaching the final iteration $N$, an additional final summary iteration ($N+1$) evaluates the higher-order concept reasoned from top-scoreboard descriptions; here, \texttt{Step 1} is replaced by a \texttt{Summary Concept} step, while the remaining steps are unchanged. The highest-scoring concept from $\mathcal{H}$ is then returned. The \LINE{} algorithm is summarized in \Cref{app:pseudocode}.}
    \label{fig:schema}
\end{figure}

In this section, we describe \LINE{}, an automatic, training-free, gray-box framework for labeling vision model neurons. An overview of the \LINE{} algorithm is illustrated in \Cref{fig:schema} and detailed in \Cref{sec:line}. We establish formal notation in \Cref{sec:preliminaries} (with full detailed notation in \Cref{app:notation}) and provide a comprehensive characteristic comparison against baseline methods in \Cref{app:method_comparison}.

\subsection{Preliminaries}
\label{sec:preliminaries}

Let $f: \mathcal{X} \to \mathcal{A}$ denote a target vision network, where $\mathcal{X} \subset \mathbb{R}^{C \times H \times W}$ represents the input image domain and $\mathcal{A}$ denotes the activation space of a specific model layer under inspection. For Convolutional Neural Networks (CNNs), the activation space typically resides in $\mathbb{R}^{C' \times H' \times W'}$, where $C'$ is the number of channels and $H' \times W'$ is the spatial resolution. To obtain a vector representation, we apply a spatial global pooling function $p(\cdot)$, yielding pooled activations $a = p(f(x))$, such that $a \in \mathbb{R}^{D}$, where $D$ denotes the number of neurons. For layers that are inherently one-dimensional, such as global average pooling (\texttt{avgpool}), the identity mapping is used. For Vision Transformers (ViT-B/16) trained on ImageNet, we focus on the \texttt{encoder} layer, and for DINOv3~\citep{simeoni2025cijo}, on the \texttt{norm} layer, which are natively one-dimensional. The objective of neuron labeling is to define an explanation function $E: [D] \to \mathcal{T}$ that assigns semantic textual descriptions from a set $\mathcal{T}$ to each neuron, providing a human-interpretable concept for each of the $D$ dimensions in $\mathcal{A}$.

\subsection{\LINE{}: LLM-based Iterative Neuron Explanations}
\label{sec:line}

\LINE{} is a training-free iterative approach that leverages four core components in a loop (see \Cref{fig:schema}). Similar to prior methods, \LINE{} analyzes one neuron $n$ at a time, allowing for parallel and independent explanations of each neuron in the model. Following an initialization phase that creates an initial concept scoreboard $\mathcal{H}$, each iteration proceeds through the stages of concept proposal, image synthesis, activation extraction, and concept scoring, allowing evaluation of a newly added concept to the scoreboard. This process repeats for $N$ iterations, concluding with a final summary step and yielding the highest-scoring concept from $\mathcal{H}$. The algorithm in \Cref{app:pseudocode} summarizes the proposed method. All the prompts used in the pipeline are detailed in \Cref{app:prompts}.

\textbf{Initialization.}
To avoid the slow convergence associated with uniform or zero-score distributions, we initialize the concept scoreboard $\mathcal{H}$ with labels from the ImageNet-1K validation set~\citep{ILSVRC15}, which provides a diverse vocabulary that closely resembles the target model's training distribution. For each of the $K=1000$ classes, we pass $M=50$ images through the target model and extract the neuron's activations to form a control activation matrix $A_{init} \in \mathbb{R}^{K \times M}$. These activations are evaluated by the scorer on natural images from the corresponding class, producing an initial score vector $s_{init} \in \mathbb{R}^{K}$. Finally, to balance exploitation and exploration, the scoreboard $\mathcal{H}$ is populated with the top $5$ highest-scoring class labels alongside $5$ randomly selected ones, and their scores are calculated on synthesized images to ensure consistency with the iterative process. \\

\textbf{Concept Proposal.}
 Like MILS~\citep{ashutosh2025llms}, we employ an LLM to reason and propose a new concept $t$ based on the already evaluated concepts in $\mathcal{H}$. After $N$ iterations, a final summary step is executed. The LLM is provided with the top three concepts from $\mathcal{H}$ to generate a generalized explanation. This summary concept is evaluated via the standard pipeline, appending the results to $\mathcal{H}$.

\textbf{Image Synthesis.}
We use a T2I model to generate a set $\mathcal{P}_t$ of synthetic images for the proposed concept. To ensure visual diversity, we use the prompt template: \texttt{"A realistic photo of a \{concept\}, \{angle\}, \{lighting\}"}, where sampling of environment features is described in \Cref{app:prompts}.

\textbf{Activation Extraction.}
The generated set of images $\mathcal{P}_t$ is processed by the target model $f$ to obtain the activation vector set $A_t$ for the specific neuron $n$:
\begin{equation}
    A_t = [a_n(x_1^t), a_n(x_2^t), \dots, a_n(x_{|\mathcal{P}_t|}^t)],
\end{equation}
where $a_n(x_i^t)$ is the activation of neuron $n$ for the $i$-th synthetic image of concept $t$.

\textbf{Concept Scoring.}
While prior work often used Mean Activation Difference (MAD)~\citep{kopf2024cosy} or the Area Under the Curve (AUC) relative to a reference set, our preliminary analysis indicates that the mean over $A_t$ provides a more granular and continuous scoring signal $s$. Specifically, AUC and MAD require a control set to be computed, and AUC frequently saturates, limiting its discriminative power. By maximizing the neuron's response, this continuous objective steers the LLM toward increasingly relevant semantic labels. Formally, given a concept $t$ with activation set $A_t$, the scalar score $s_t$ is computed via scoring function $\psi$ as: $s_t = \psi_{\mathrm{avg}}(A_t) = \frac{1}{|A_t|} \sum_{a \in A_t} a$.

\section{Experiments}
\label{sec:experiments}

In this section, we evaluate \LINE{} across diverse architectures and tasks. We focus specifically on the \emph{deeper layers} of the networks---these layers capture the high-level semantics critical for safety auditing, in contrast to the low-level features (e.g., color or texture) found in earlier layers \citep{lecun2015deep}. For evaluations of the lower layers of the evaluated models, we refer the reader to \Cref{app:lower_evaluations}. \LINE{} uses Stable Diffusion XL (SDXL) with $|\mathcal{P}_t|=5$ images generated per concept and Llama 3.1 8B~\citep{grattafiori2024llama} with temperature of $0.5$ and nucleus sampling of $0.9$. The number of refinement iterations is set to $N=10$, unless specified otherwise.

\Cref{sec:quantitative} presents a quantitative assessment on the CoSy benchmark~\citep{kopf2024cosy} across ResNet18/50, ViT-B/16, trained on either ImageNet or Places365 and DINOv3~\citep{simeoni2025cijo}, whose training dataset (LVD-1689M) is not publicly available. We compare \LINE{} against top CoSy baselines INVERT~\citep{bykov2024labeling} and CLIP-Dissect~\citep{oikarinen2023clip}, alongside the recently proposed DnD~\citep{bai2025interpreting}. Full setup details are provided in \Cref{app:reproducibility}.
\Cref{sec:qualitative} later focuses on a qualitative comparison with the baselines and introduces a causal validation framework that leverages an image-to-image generative model to ablate labeled concepts and measure the resulting drop in neuron activation. In \Cref{sec:visual_explanations}, we evaluate our visual explanations (best concept synthesized images) against established activation maximization methods, following the Salient ImageNet~\citep{singla2022salient} evaluation protocol.
Finally, Sections~\ref{sec:ablation_optimization} and~\ref{sec:ablation_t2i} analyze how the number of iterative steps and T2I model choice affect \LINE{} performance.

\subsection{Quantitative Evaluation}
\label{sec:quantitative}

\begin{table}[t]
    \centering
    \caption{\textbf{Evaluation on the CoSy benchmark and VLM correlation.} We compare \LINE{} against CLIP-Dissect, INVERT, and DnD. Evaluations are performed on the following layers: \texttt{avgpool} for ResNet50 and ResNet18 (trained on ImageNet-1k and Places365), \texttt{encoder} for ViT-B/16 (ImageNet-1k), and \texttt{norm} for DINOv3. DnD results on ViT-based models are affected by a bug in the original implementation (see \Cref{app:baseline_implementation}). We report CoSy metrics MAD and AUC, as well as the Pearson correlation ($\times 10^{-1}$) between neuron activations and the SigLIP similarity of each image to the neuron's predicted label. For all metrics, higher is better ($\uparrow$). Scores are averaged over 100 neurons. The best and second-best results in each row are highlighted in \textbf{bold} and \underline{underlined}, respectively.}
    \label{tab:comparison_table}
    \vspace{0.5em}
    \begin{tabular}{l l l c c c c}
    \toprule
    \textbf{Model} & \textbf{Dataset} & \textbf{Metric} & \textbf{\LINE} & \textbf{CLIP-Dissect} & \textbf{INVERT} & \textbf{DnD} \\
    \midrule
    \multirow{3}{*}{ResNet18} & \multirow{3}{*}{ImageNet}
      & AUC & $\mathbf{0.97}_{\pm 0.04}$ & $\underline{0.90}_{\pm 0.14}$ & $0.85_{\pm 0.14}$ & $0.77_{\pm 0.21}$ \\
     & & MAD & $\mathbf{4.76}_{\pm 1.91}$ & $\underline{3.42}_{\pm 2.17}$ & $2.16_{\pm 1.61}$ & $1.97_{\pm 1.99}$ \\
     & & Corr & $\mathbf{0.57}_{\pm 0.49}$ & $0.41_{\pm 0.39}$ & $\underline{0.47}_{\pm 0.45}$ & $0.44_{\pm 0.46}$ \\
    \midrule
    \multirow{3}{*}{ResNet18} & \multirow{3}{*}{Places365}
      & AUC & $\mathbf{0.93}_{\pm 0.10}$ & $\underline{0.89}_{\pm 0.15}$ & $0.80_{\pm 0.18}$ & $0.85_{\pm 0.19}$ \\
     & & MAD & $\mathbf{3.77}_{\pm 2.31}$ & $\underline{3.63}_{\pm 2.78}$ & $1.85_{\pm 1.70}$ & $3.16_{\pm 2.92}$ \\
     & & Corr & $\underline{0.51}_{\pm 0.54}$ & $0.29_{\pm 0.43}$ & $0.20_{\pm 0.33}$ & $\mathbf{0.64}_{\pm 0.58}$ \\
    \midrule
    \multirow{3}{*}{ResNet50} & \multirow{3}{*}{ImageNet}
      & AUC & $\mathbf{0.97}_{\pm 0.07}$ & $0.87_{\pm 0.18}$ & $\underline{0.88}_{\pm 0.15}$ & $0.85_{\pm 0.18}$ \\
     & & MAD & $\mathbf{5.28}_{\pm 2.10}$ & $\underline{3.56}_{\pm 2.52}$ & $2.81_{\pm 1.95}$ & $3.40_{\pm 3.02}$ \\
     & & Corr & $\mathbf{0.81}_{\pm 0.57}$ & $0.33_{\pm 0.46}$ & $0.48_{\pm 0.51}$ & $\underline{0.66}_{\pm 0.47}$ \\
    \midrule
    \multirow{3}{*}{ResNet50} & \multirow{3}{*}{Places365}
      & AUC & $\mathbf{0.94}_{\pm 0.08}$ & $\underline{0.89}_{\pm 0.15}$ & $0.84_{\pm 0.14}$ & $0.84_{\pm 0.18}$ \\
     & & MAD & $\mathbf{4.27}_{\pm 2.81}$ & $\underline{3.69}_{\pm 2.90}$ & $2.04_{\pm 1.70}$ & $2.74_{\pm 2.56}$ \\
     & & Corr & $\underline{0.36}_{\pm 0.56}$ & $0.24_{\pm 0.48}$ & $0.16_{\pm 0.30}$ & $\mathbf{0.57}_{\pm 0.54}$ \\
    \midrule
    \multirow{3}{*}{ViT-B/16} & \multirow{3}{*}{ImageNet}
      & AUC & $\mathbf{0.94}_{\pm 0.07}$ & $0.74_{\pm 0.20}$ & $\underline{0.79}_{\pm 0.18}$ & $0.64_{\pm 0.20}$ \\
     & & MAD & $\mathbf{2.12}_{\pm 0.54}$ & $1.00_{\pm 0.89}$ & $\underline{1.05}_{\pm 0.70}$ & $0.60_{\pm 0.84}$ \\
     & & Corr & $\mathbf{0.26}_{\pm 0.21}$ & $0.07_{\pm 0.15}$ & $\underline{0.23}_{\pm 0.21}$ & $0.08_{\pm 0.15}$ \\
     \midrule
    \multirow{3}{*}{DINOv3} & \multirow{3}{*}{LVD-1689M}
      & AUC & $\mathbf{0.94}_{\pm 0.08}$ & $\underline{0.76}_{\pm 0.22}$ & $\underline{0.76}_{\pm 0.18}$ & $0.62_{\pm 0.23}$ \\
     & & MAD & $\mathbf{2.05}_{\pm 0.68}$ & $\underline{1.12}_{\pm 1.06}$ & $0.95_{\pm 0.73}$ & $0.47_{\pm 0.92}$ \\
     & & Corr & $\mathbf{0.32}_{\pm 0.23}$ & $0.09_{\pm 0.20}$ & $\underline{0.24}_{\pm 0.24}$ & $0.11_{\pm 0.21}$ \\
    \midrule
    \multirow{3}{*}{\textbf{Overall}} & \multirow{3}{*}{-}
      & AUC & $\mathbf{0.95}_{\pm 0.08}$ & $\underline{0.84}_{\pm 0.19}$ & $0.82_{\pm 0.17}$ & $0.76_{\pm 0.22}$ \\
     & & MAD & $\mathbf{3.71}_{\pm 2.27}$ & $\underline{2.74}_{\pm 2.50}$ & $1.81_{\pm 1.61}$ & $2.06_{\pm 2.50}$ \\
     & & Corr & $\mathbf{0.47}_{\pm 0.50}$ & $0.24_{\pm 0.39}$ & $0.24_{\pm 0.38}$ & $\underline{0.41}_{\pm 0.49}$ \\
    \bottomrule
    \end{tabular}
\end{table}

\textbf{CoSy Benchmark.} \Cref{tab:comparison_table} presents our quantitative evaluation on the CoSy benchmark (\Cref{app:cosy}). Across three architectures~\footnote{DINOv3 shares the ViT-B/16 architecture; therefore, we count it as a single architecture with ViT-B/16} and three datasets, \LINE{} consistently achieves superior performance across all metrics. In \Cref{app:statistical}, we confirm that the performance gains over the baselines are statistically significant. Although \Cref{app:quantitative_comparison} shows that switching from the SDXL T2I model lowers performance on the CoSy metrics (see \Cref{sec:ablation_t2i} for further discussion), \LINE{} still outperforms the baselines. Notably, \LINE{}'s superiority extends even to lower network layers, as presented in \Cref{app:lower_evaluations}. We attribute the surprisingly low performance of DnD to its reliance on image cropping---a structural weakness similarly observed in the CoSy evaluations of the crop-based FALCON~\citep{kopf2024cosy} method. Because both methods rely heavily on crops during their search phase, they often produce misleading, uninterpretable samples for the Vision-Language Model to describe, forcing it to generate overly broad descriptions (e.g., \textit{various perched creatures}). Ultimately, these findings lead us to question whether cropping is a viable strategy for neuron analysis, even in lower layers (see \Cref{app:lower_evaluations}).

\textbf{VLM Correlation Evaluation.} A limitation of the CoSy benchmark is that neither AUC nor MAD passes all of the sanity checks of~\citet{oikarinen2025evaluating}. Moreover, since both CoSy and \LINE{} rely on T2I models, one could attribute the strong performance of \LINE{} to exploiting the evaluation setup. We therefore additionally evaluate \LINE{} with the simulation with correlation scoring~\citep{oikarinen2024linear}, which computes the Pearson correlation between neuron activations and SigLIP~\citep{zhai2023sigmoid} label-image similarities across the dataset (see implementation details in \Cref{app:siglip}). As shown in \Cref{tab:comparison_table}, \LINE{} ranks first or second on every model and first overall. While CoSy may favor more specific concepts, the global nature of the correlation evaluation indicates that the performance of \LINE{} is not an artifact of a generator-based benchmark, and that its concepts align with the full range of neuron activations rather than only the top of it.

\begin{figure}[t]
    \centering
    \includegraphics[width=\linewidth]{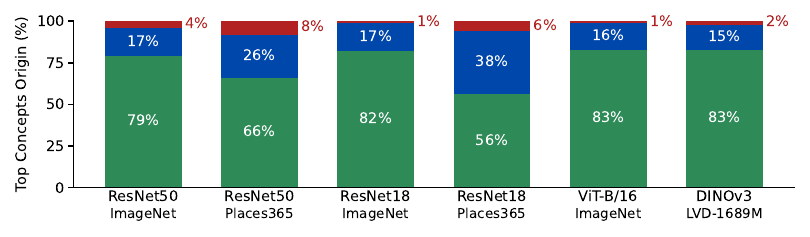}
    \caption{\textbf{Proportion of predefined vs.\ generated labels in \LINE{}.} For the models, layers and neurons evaluated in \Cref{tab:comparison_table}, we report a breakdown of the final top-scoring concepts, stacked bottom to top: concepts taken from the \textcolor{ourgreen}{\textbf{initial vocabulary}}, concepts discovered during the \textcolor{ourblue}{\textbf{iterative loop}}, and concepts proposed at the last \textcolor{ourred}{\textbf{summary step}}. Even with a rich initial vocabulary, \LINE's refinement loop discovers novel top concepts, highlighting the limitations of relying solely on a fixed vocab set.}
    \label{tab:share_of_generated_concepts}
\end{figure}

\textbf{Impact of Iterative Refinement.} To evaluate the refinement process, \Cref{tab:share_of_generated_concepts} reports the percentage of final labels proposed during the iteration process versus those selected from the initial vocabulary. The LLM in \LINE{} generates the winning label for $25\%$ of neurons on average. While the summary step yields only marginal gains, it still contributes more than a single iteration step, which justifies its presence in the pipeline. These results, alongside the results of comparing \LINE{} to open-vocabulary baselines in \Cref{tab:comparison_table}, demonstrate that zero-shot methods constrained by predefined vocabularies frequently miss higher-scoring labels that \LINE{} iteratively discovers.

\textbf{Scoreboard Analysis.}
In \Cref{app:single_loop}, we present the detailed iteration loop with the full scoreboard for neuron 1255, labeled as \textit{weightlifting}. Initial predefined labels (e.g., \textit{pool table}, \textit{dumbbell}) correctly anchor the LLM's reasoning toward workout activities. Because these initial concepts were suboptimal, \LINE{} iteratively generated novel, higher-scoring concepts, proving that single-pass reasoning is often insufficient. Ultimately, the LLM proposed \textit{table-based exercise} as a global summary. Although this summary concept scored lower than \textit{weightlifting}, it effectively captures the broader semantic abstraction of the neuron's learned features.

\subsection{Qualitative evaluation}
\label{sec:qualitative}

In \Cref{fig:main_qualitive}, we qualitatively compare the concept descriptions for randomly selected neurons from the models and layers evaluated in \Cref{sec:quantitative}. The results confirm that the concepts proposed by \LINE{} are highly consistent with the visual features of the top-activating images. While CLIP-Dissect and INVERT occasionally provide similar labels like \textit{chicken} against more precise \LINE{} \textit{hen} (see neuron 64 of DINOv3 in \Cref{fig:neuron_descriptions_dinov3}), \LINE's descriptions are notably more accurate (e.g., \textit{aquatic biome} versus \textit{snorkeling}, or \textit{coelenterate}). Potential missed concepts may arise if the T2I model fails to synthesize the target concept, which is visible for abstract concepts like \textit{weighted axis}, in \Cref{tab:iteration_visualizations}. However, because of \LINE{}'s transparency, such failures can be easily detected.

\begin{figure}[h]
    \centering
    \includegraphics[width=0.9\linewidth]{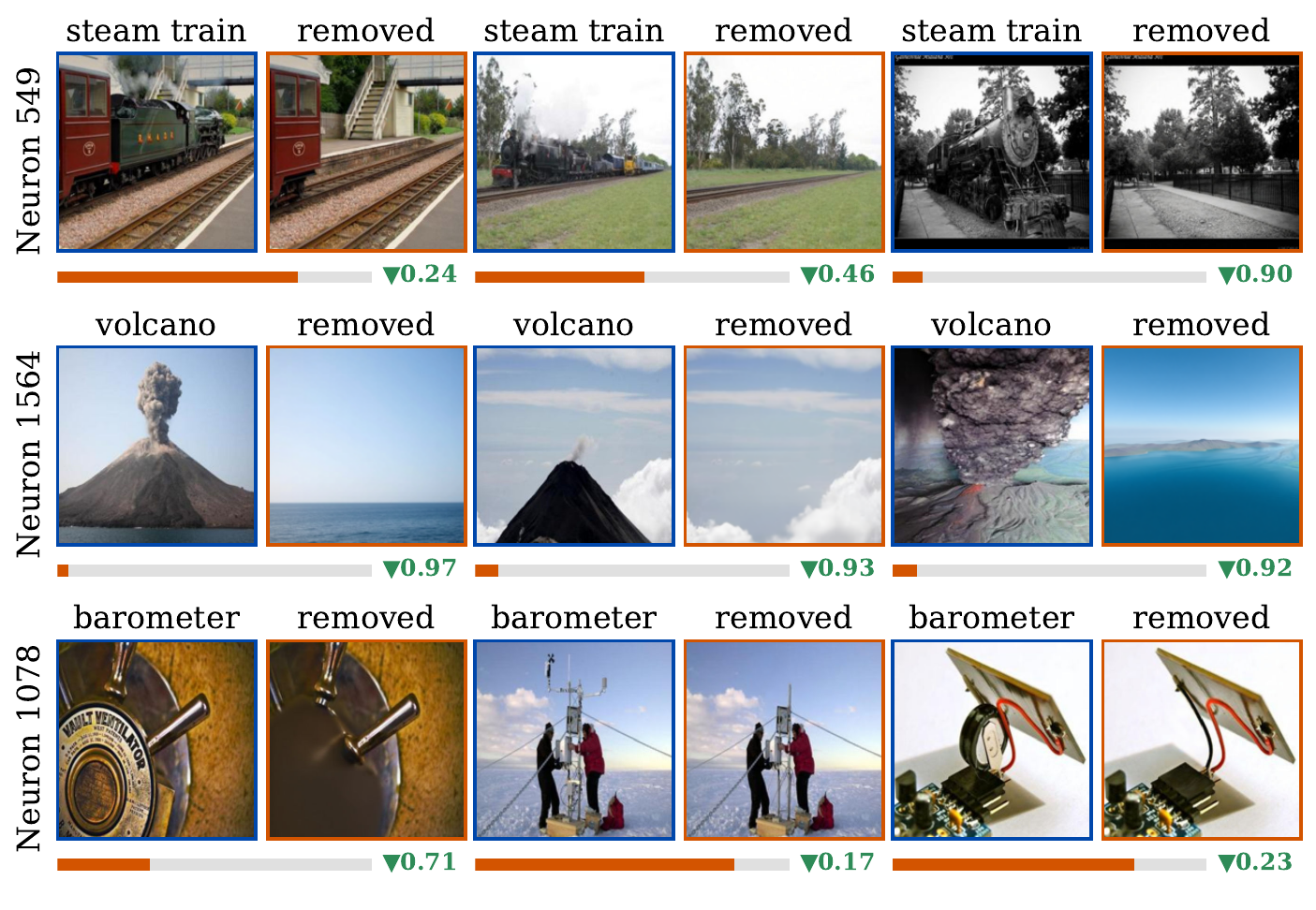}
    \caption{\textbf{Causal impact of concept ablation on neuron activation.} Using image-to-image generative models, we remove visual concepts associated with the neuron description from ResNet50 \textit{avgpool} layer identified by \LINE{}. Original (\emph{left)} and ablated (\emph{right}) images are shown side-by-side, with normalized neuron activations displayed below each pair. Original images are outlined in \textcolor{ourblue}{\textbf{blue}}, ablated versions in \textcolor{ourorange}{\textbf{orange}}, and the resulting relative drop in activation is marked in \textcolor{ourgreen}{\textbf{green}}. Extended results featuring failure cases are provided in \Cref{app:causal}.}
    \label{fig:main_causal}
\end{figure}

\textbf{Causal Analysis.}
To verify label fidelity, we perform an ablation study on Places365. Using an image-to-image (I2I) generative model, we remove only the object corresponding to the assigned neuron label while preserving the rest of the scene (details in \Cref{app:causal}). As \Cref{fig:main_causal} shows, removing the \textit{volcano} or \textit{steam train} drops activations for neurons 1564 and 549 to near zero. Similarly, altering or removing the \textit{barometer} results in reduced activation. While \Cref{fig:main_causal} confirms \LINE{} captures causal activation patterns, \Cref{app:causal} details protocol limitations. Specifically, the generative process occasionally fails to ablate the target feature or introduces image reconstruction artifacts, raising the question of whether failures stem from incorrect labeling or from added artifacts.

\subsection{Visual Explanations}
\label{sec:visual_explanations}

\begin{figure}[t]
    \centering\includegraphics[width=0.95\textwidth]{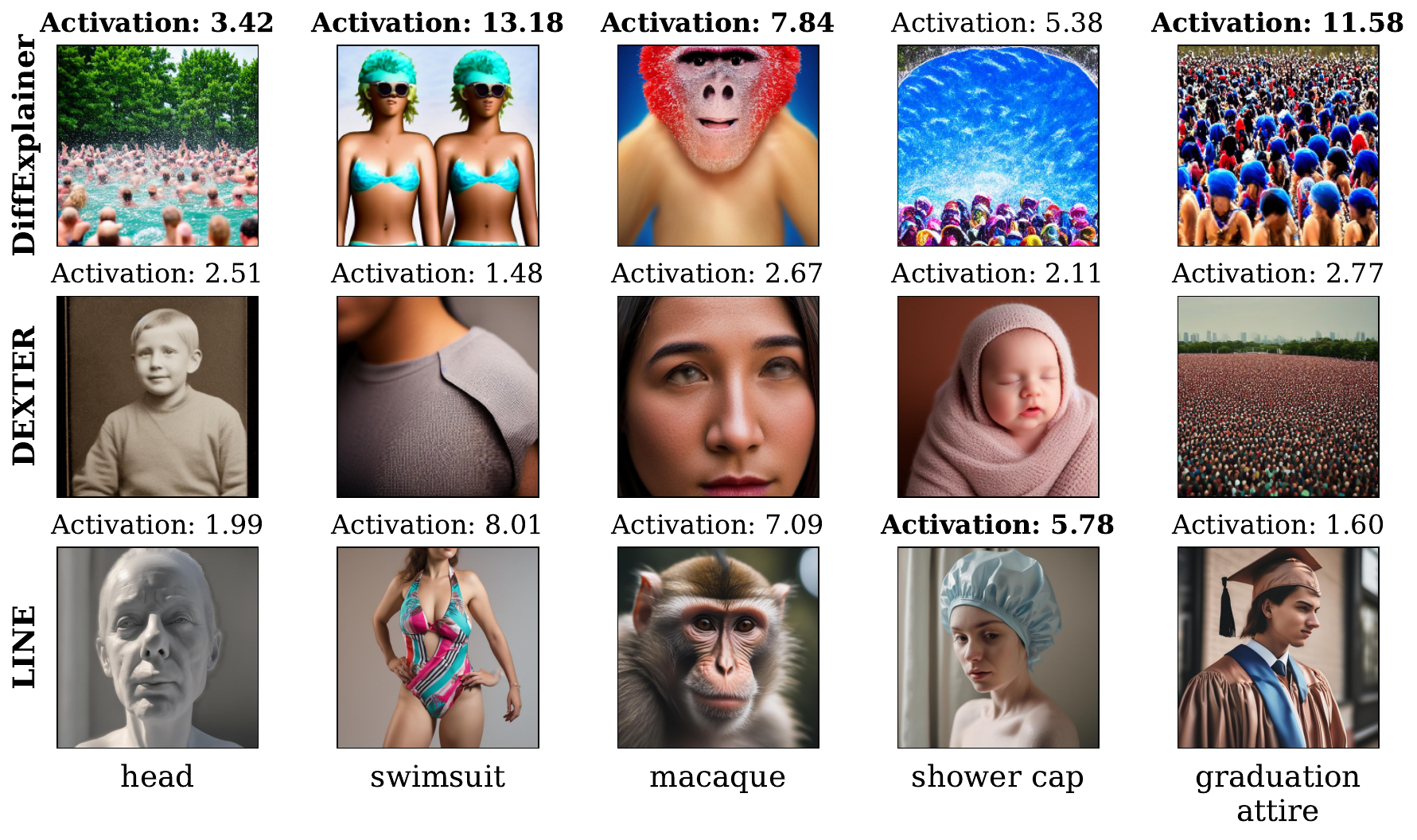}
    \caption{\textbf{Visual explanation comparison on Salient ImageNet.} We present visual explanations generated by \LINE{}, DiffExplainer, and DEXTER for the top-5 bias features of the ``Bathing Cap'' class in RobustResNet50. Neuron activation values are displayed above each visual explanation, with the corresponding \LINE{} neuron description provided at the bottom. Compared to AM methods, \LINE{} produces explanations without visual artifacts common in DiffExplainer while achieving higher activations than DEXTER. \Cref{app:extended_visual} provides extended results for core and bias features on additional classes from Salient ImageNet, alongside Grad-CAM-like maps~\citep{Selvaraju_2017_ICCV}.}
    \label{fig:am_main}
\end{figure}

By coupling iterative concept search with image synthesis, \LINE{} generates highly activating images that serve as visual explanations alongside textual descriptions. This enables direct comparison with Activation Maximization (AM) methods like DiffExplainer and DEXTER. Following the Salient ImageNet protocol~\citep{singla2022salient}, \Cref{fig:am_main} presents visual explanations for the top bias features correlated with the class in a RobustResNet50, while additional results for different biased and core features are provided in \Cref{app:extended_visual}. Although \LINE{} outperforms DEXTER on average, it often yields lower neuron activations than DiffExplainer.
This is because standard gradient-optimized AM methods often exploit non-descriptive artifacts, producing uninterpretable visual noise. While DEXTER mitigates this via a text-model proxy, \LINE{} avoids this noise entirely as it does not require gradient optimization. Compared to DEXTER, \LINE{} generates easy-to-interpret explanations that achieve higher activations (see \Cref{app:extended_visual}) while requiring only access to the neuron activation magnitude. Finally, we attribute \LINE{}’s lower activations on features like ``graduation attire'' to its initialization phase and the hard-to-describe nature of the neuron. In our setup, the scoreboard is seeded with ImageNet labels that in this case are uncorrelated with the target feature, ultimately preventing \LINE{} from successfully discovering it.

\begin{figure}[b]
    \centering
    \begin{subfigure}{0.49\textwidth}
        \centering
        \includegraphics[width=\textwidth]{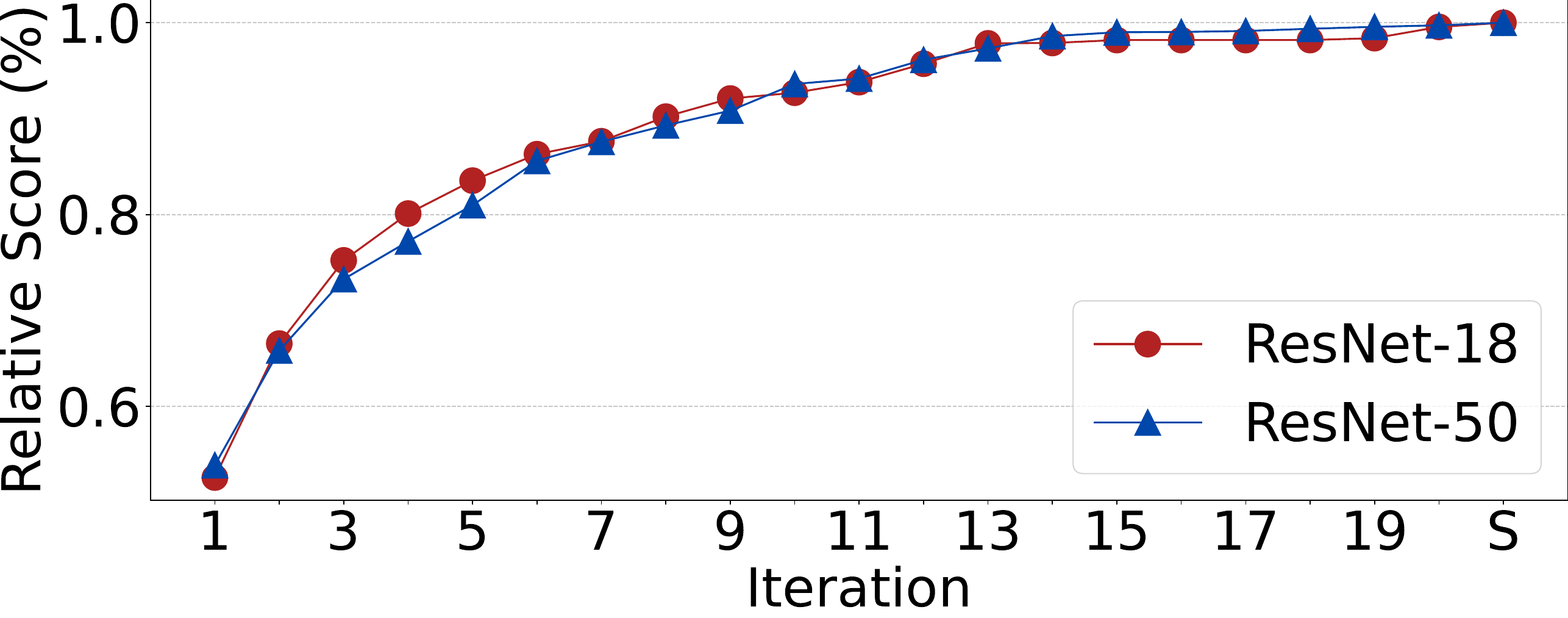}
    \end{subfigure}\hfill
    \begin{subfigure}{0.49\textwidth}
        \centering
        \includegraphics[width=\textwidth]{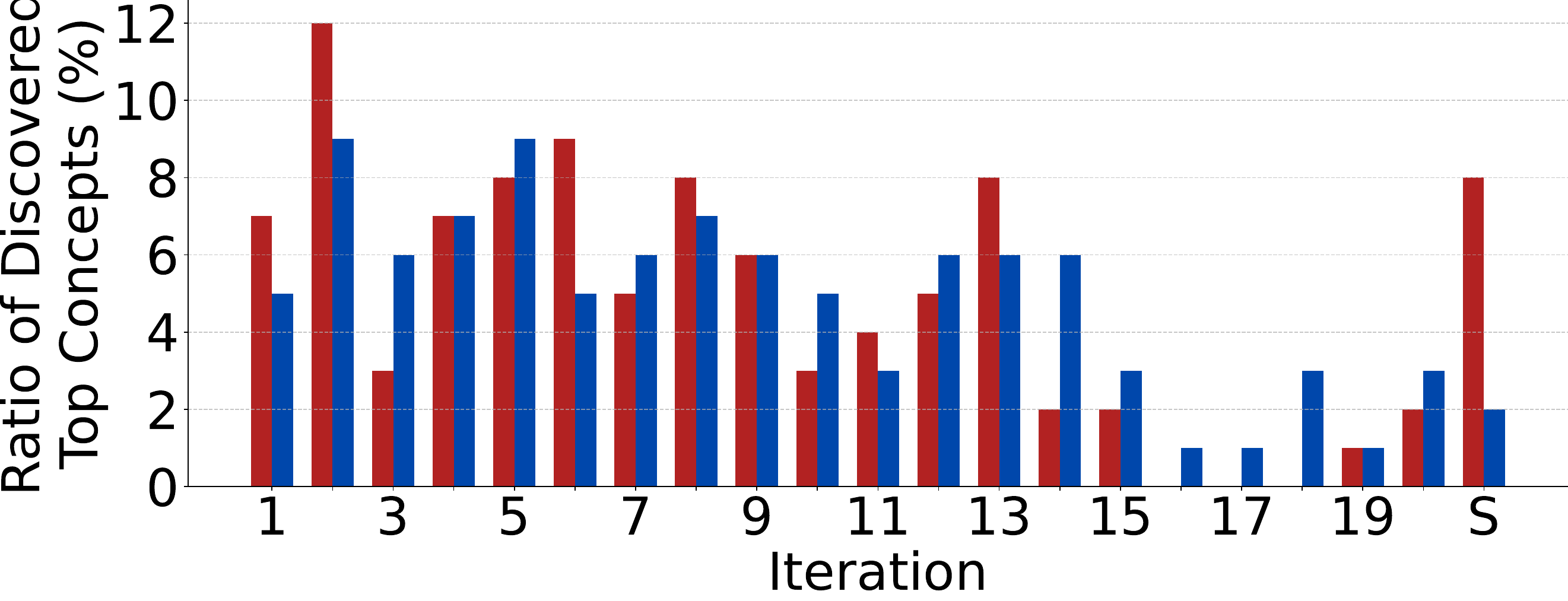}
    \end{subfigure}
    \caption{\textbf{Performance across iteration steps.} We extend the maximum number of iterations from 10 to 20 for 100 random neurons from the ResNet50 and ResNet18 (Places365) \texttt{avgpool} layers. Over successive iterations, we report the relative average best activation score (\emph{left panel}) and the discovery rate of the optimal description (\emph{right panel}) at each step. The ``S'' label denotes the final summary iteration. On average, the iteration loop consistently refines concepts, yielding rapid improvements up to steps 5-6, steady gains through step 15, and plateauing afterward.}
    \label{fig:optimization}
\end{figure}

\subsection{Ablation: Performance over Optimization Steps}
\label{sec:ablation_optimization}

To evaluate the iterative loop, we increase the number of iterations from 10 to 20. \Cref{fig:optimization} analyzes the step at which the optimal neuron label emerges for ResNet18 and ResNet50. Both models discover novel concepts beyond 10 iterations, showing rapid progress up to step 5, followed by nearly linear improvement until step 15, and marginal gains thereafter. The summary step still yields a small increase (discovering an average of 4\% of the final concepts), indicating that it remains useful even with the larger iteration budget. When comparing the frequency of the top-selected descriptions with the step at which they were discovered, we observe a similar pattern.

\subsection{Ablation: Impact of Different T2I Models}
\label{sec:ablation_t2i}

\begin{figure}[t]
    \centering
    \raisebox{0.1cm}{
    \begin{subfigure}[c]{0.4\textwidth}
        \centering
        \includegraphics[width=\textwidth, keepaspectratio]{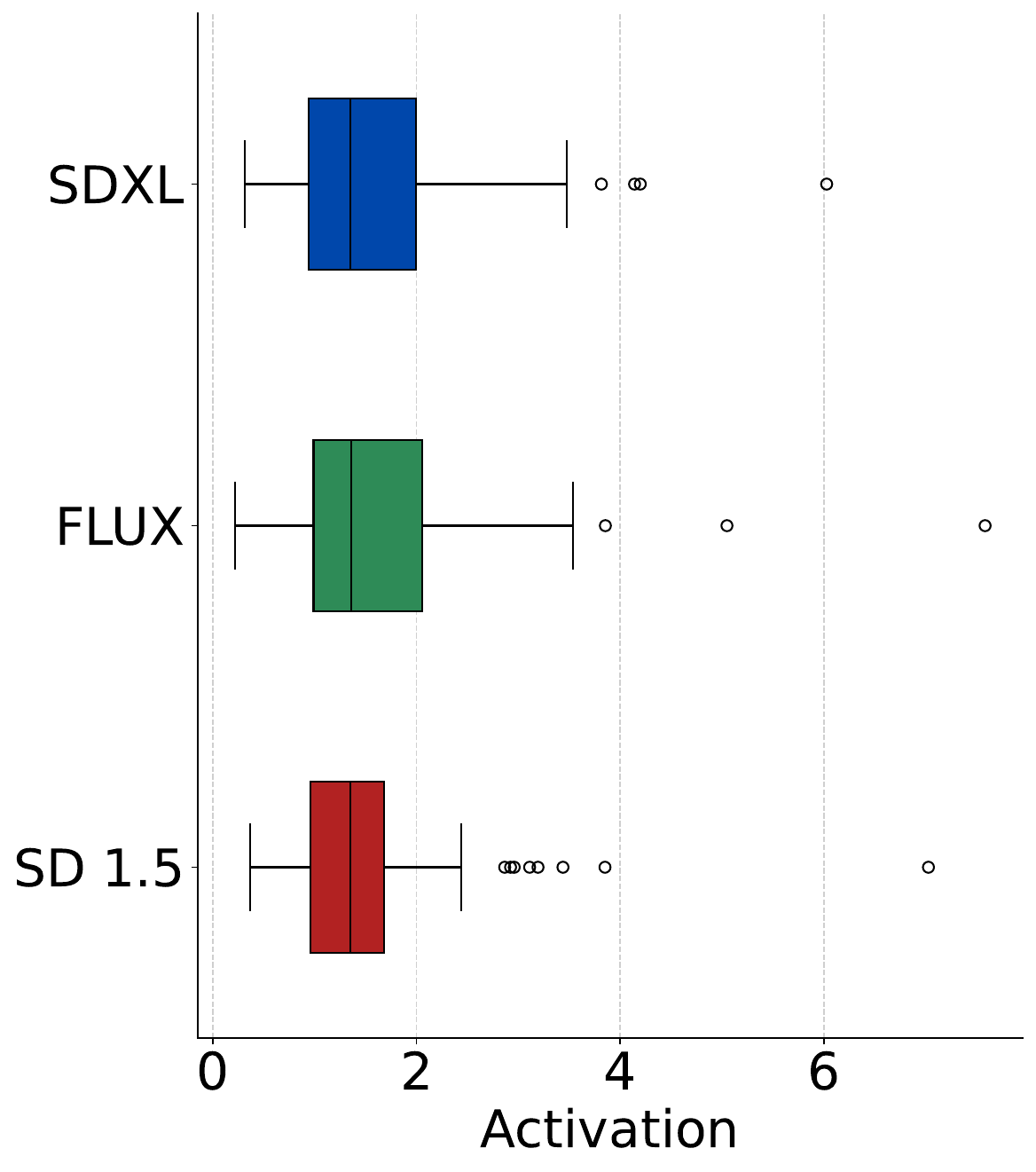}
    \end{subfigure}} \hspace{0.4cm}
    \begin{minipage}[c]{0.5\textwidth}
        \begin{subfigure}{\textwidth}
            \includegraphics[width=\textwidth]{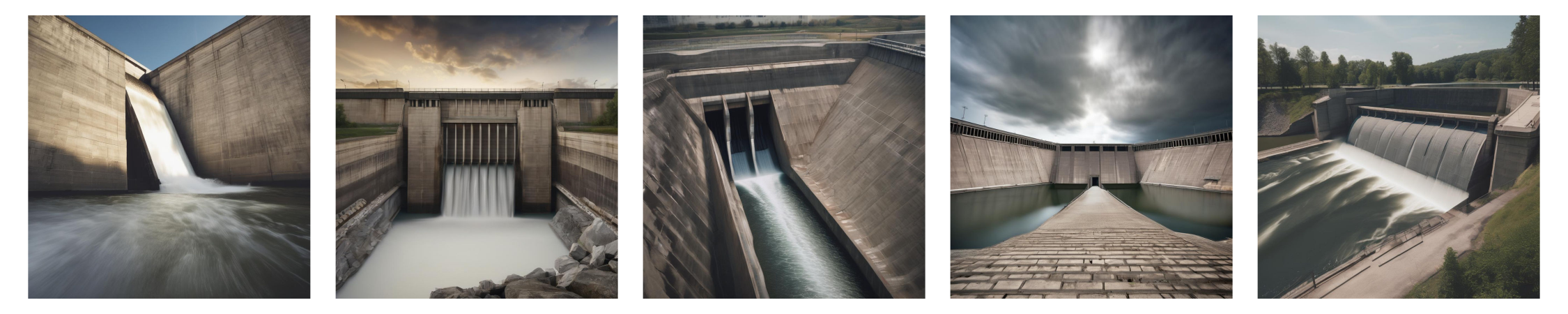}
            \vspace{-0.45cm}
            \caption{SDXL: \textit{Dam} (score: 1.65)}
        \end{subfigure}
        \begin{subfigure}{\textwidth}
            \includegraphics[width=\textwidth]{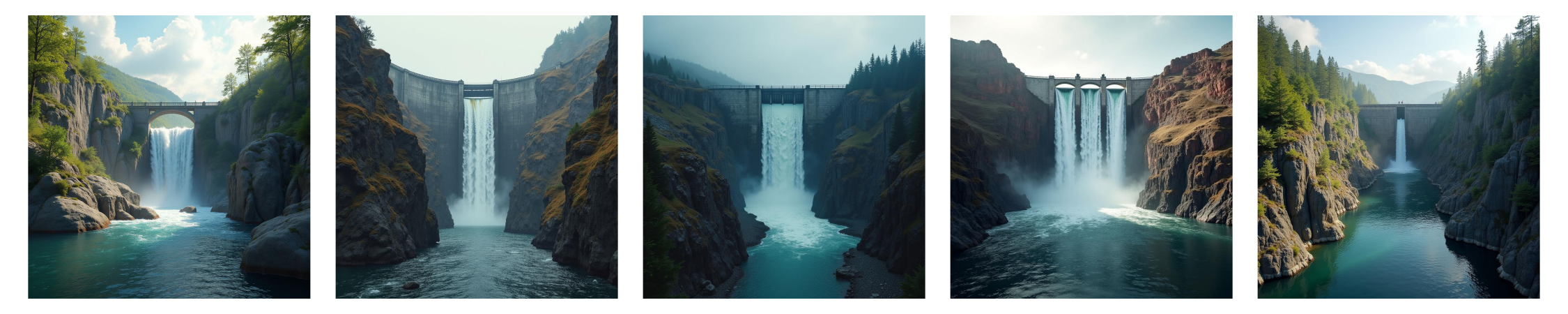}
            \vspace{-0.45cm}
            \caption{FLUX: \textit{Dam} (score: 1.11)}
        \end{subfigure}
        \begin{subfigure}{\textwidth}
            \includegraphics[width=\textwidth]{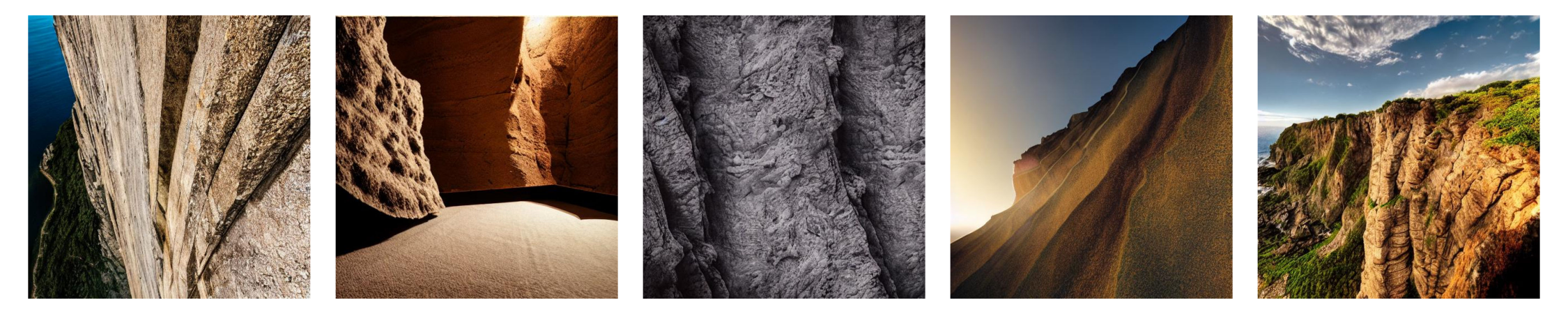}
            \vspace{-0.45cm}
            \caption{SD1.5: \textit{Cliff} (score: 1.25)}
        \end{subfigure}
        \vspace{0.01cm}
    \end{minipage}
    \caption{\textbf{Impact of text-to-image (T2I) models on \LINE\ performance.} We evaluate 100 random neurons from the \texttt{avgpool} layer of ResNet50 trained on Places365. The box plots (\textit{left}) comparing the highest concept activation scores indicate that the FLUX model produces slightly higher activations. Corresponding descriptions and visual explanations for neuron 153 (\textit{right}) illustrate the distinct generative priors of each T2I model: SD1.5 tends toward photorealistic outputs, whereas SDXL and FLUX lean into stylized aesthetics, with FLUX producing the most pronounced cinematic results.}
    \label{fig:sd_models}
\end{figure}

The text-to-image (T2I) model is a core component of \LINE. To evaluate its impact, we compare SD1.5, our SDXL baseline, and FLUX.1[dev] on ResNet50 (\Cref{fig:sd_models}). While FLUX.1[dev] concepts often have higher activation scores, the difference lacks statistical significance (\Cref{app:statistical}). We therefore adopt SDXL due to its lower memory overhead and faster inference. Qualitatively (in \Cref{fig:sd_models}), we observe that each T2I model introduces distinct generative patterns to the images, altering visual aesthetics without compromising semantic intent.
Since \LINE{} and CoSy both rely on SDXL, the CoSy results could in principle depend on this shared generator. In \Cref{app:quantitative_comparison} we therefore compare AUC when we swap T2I with FLUX in \LINE{}. On CoSy, scores drop, confirming that the benchmark is sensitive to the generator, yet \LINE{} with FLUX still remains above the baselines. On the VLM evaluation, which uses no generated images, the two variants perform on par, indicating that the choice of T2I model matters only when the evaluation uses a matching generator.

\section{Conclusion}
\label{sec:conclusion}
As deep learning models become more prevalent in high-stakes applications, it is pivotal to understand their decision-making. To this end, we present \LINE{}, the first training-free, gray-box iterative pipeline for automated neuron labeling in vision models. \LINE{} achieves state-of-the-art performance via a transparent discovery process that yields textual and visual explanations. Moreover, its outputs form structured, model-specific findings that could be aggregated in catalogs of model findings such as Modelpedia \citep{bernat2026modelpediacatalogmodelfindings}, supporting meta-scientific analysis of how vision models represent concepts. \LINE's modular structure provides a highly scalable solution, enabling seamless integration of future GenAI models.

\textbf{Limitations and Future Work.}
We identify three main limitations of \LINE{} that provide natural opportunities for future research. First, for polysemantic neurons \LINE{} may optimize the score by exploring only a single direction, which may not generalize to the whole activation space, as discussed in \Cref{app:single_loop}.
Therefore, further improvement can be achieved by promoting more diverse exploration. In \Cref{app:multi_comparison}, we present how simply utilizing \LINE{}'s scoreboard for multi-label description can boost its performance.
Second, the inference can become a computational burden when explaining a high number of neurons.
For lowering the computational demand, we recommend reducing the number of inference steps or replacing the T2I model, responsible for 88\% of the compute time (see \Cref{app:computationl}), with a distilled or smaller model.
Finally, \LINE{} requires LLMs and T2I models to comprehend target concepts, creating a bottleneck in specialized domains (e.g., medical and satellite imaging) that might lack high-fidelity image and text generators.

\textbf{Broader impact.}
Ultimately, we envision \LINE{} as a foundational step toward more transparent and accountable vision models, enabling safer deployment in high-stakes domains.
We further elaborate on the potential positive and negative societal impacts of this work in \Cref{app:broader_impact}.

\clearpage

\section*{Acknowledgements}

Work on this project is financially supported by the Foundation for Polish Science (FNP) grant ‘Centre for Credible AI’ No. \texttt{FENG.02.01-IP.05-0058/24}. The authors declare no competing interests.

The computational resources for this work were provided by the Laboratory of Bioinformatics and Computational Genomics and the High Performance Computing Center of the Faculty of Mathematics and Information Science, Warsaw University of Technology.

We also thank the anonymous reviewers and the AC for their useful comments.

\bibliographystyle{plainnat}
\bibliography{main}

\newpage
\appendix
\crefalias{section}{appendix}
\crefalias{subsection}{subappendix}
\crefalias{subsubsection}{subsubappendix}

\section*{Appendix for ``\LINE{}: LLM-based Iterative Neuron Explanations for Vision Models''}

\startcontents[sections]
{\hypersetup{linkcolor={blue!50!black}}
\printcontents[sections]{l}{1}{\setcounter{tocdepth}{2}}}

\clearpage
\section{Notation}
\label{app:notation}

\begin{table}[ht]
    \centering
    \caption{\textbf{Main notations used in this work.}}
    \vspace{2pt}
    \label{tab:line_notation}
    \renewcommand{\arraystretch}{1.2}
    \begin{tabularx}{\linewidth}{c X}
    \toprule
    \textbf{Notation\hspace{20pt}} & \textbf{Meaning}\\
    \midrule
    \multicolumn{2}{c}{\textit{Models \& representations}} \\
    \midrule
    $f$ & Target vision model\\
    $\mathcal{X}$ & Input image domain\\
    $\mathcal{A}$ & Activation space of a specific model layer\\
    $p$ & Spatial global pooling function\\
    $D$ & Dimensionality of the pooled activation vector\\
    $E$ & Explanation function assigning semantic descriptions\\
    $\mathcal{T}$ & Set of semantic textual descriptions\\
    \midrule
    \multicolumn{2}{c}{\textit{Neuron labeling \& evaluation}} \\
    \midrule
    $t$ & Concept description\\
    $n$ & Specific neuron identifier being analyzed\\
    $\mathcal{H}$ & \LINE{} scoreboard\\
    $A_t$ & Activation vector for a proposed concept $t$\\
    $\psi$ & Scoring function\\
    $\psi_{avg}$ & Scoring function based on mean activation over $A$\\
    $s_t$ & Scalar score produced by the scoring function for concept $t$\\
    \midrule
    \multicolumn{2}{c}{\textit{Initialization phase}} \\
    \midrule
    $K$ & Number of classes used for initialization ($1000$)\\
    $M$ & Number of images sampled per class ($50$)\\
    $A_{init}$ & Initial activation matrix for analyzed neuron, $A_{init} \in \mathbb{R}^{K \times M}$\\
    $s_{init}$ & Initial class-wise score vector, $\mathbf{s}_{init} \in \mathbb{R}^{K}$\\
    \midrule
    \multicolumn{2}{c}{\textit{\LINE{} framework \& iterations}} \\
    \midrule
    $N$ & Number of refinement iterations (excl. summary iteration) \\
    $\mathcal{P}_t$ & Set of synthetic images generated by the T2I model for the concept $t$\\
    $S$ & Final summary iteration index ($N+1$)\\
    \bottomrule
    \end{tabularx}
\end{table}

\clearpage
\section{Neuron Labeling Methods Comparison}
\label{app:method_comparison}

A comprehensive comparison between prior neuron labeling approaches and our proposed method (\LINE{}) is provided below. To properly contextualize these advancements, we categorize the methods across several key characteristics, as summarized in \Cref{tab:method_comparison}.

\begin{table}[h]
    \centering
    \footnotesize
    \caption{\textbf{Comparison of neuron labeling methods.} We contrast prior approaches with \LINE{} across several key characteristics: explanation output format, targeted neuron representation, optimization metric, and external method dependencies. We also denote whether each method is training-free and architecture-agnostic. Notably, \LINE{} leverages LLMs and T2I models to generate rich, free-form explanations without imposing architectural constraints or requiring additional training pipelines. $^*$~indicates the dependency is only required during specific phases of the method.}
    \label{tab:method_comparison}
    \vspace{0.5em}
    \setlength{\tabcolsep}{2pt}
    \resizebox{\linewidth}{!}{%
    \begin{tabular}{l c c c c c c}
    \toprule
    \textbf{Method} & \textbf{Explanation Format} & \textbf{Neuron Type} & \textbf{Target} & \textbf{Method Dependency} & \textbf{Training Free} & \textbf{Model Agnostic} \\
    \midrule
    \begin{tabular}{@{}l@{}}Network Dissection \\ \citep{bau2017CVPR}\end{tabular} & fixed-label & conv. & IoU & seg. masks & \textcolor{ourgreen}{\ding{51}} & \textcolor{ourgreen}{\ding{51}} \\
    \midrule
    \begin{tabular}{@{}l@{}}MILAN \\ \citep{hernandez2022natural}\end{tabular} & free-form & conv. & WPMI & \begin{tabular}{@{}l@{}}labeled corpus*, \\  trained LSTM\end{tabular} & \textcolor{ourred}{\ding{55}} & \textcolor{ourgreen}{\ding{51}} \\
    \midrule
    \begin{tabular}{@{}l@{}}CLIP-Dissect \\ \citep{oikarinen2023clip}\end{tabular} & open-vocabulary & scalar & SoftWPMI & CLIP & \textcolor{ourgreen}{\ding{51}} & \textcolor{ourgreen}{\ding{51}} \\
    \midrule
    \begin{tabular}{@{}l@{}}FALCON \\ \citep{kalibhat2023identifying}\end{tabular} & open-vocabulary & pre-det. & avg. CLIP score & CLIP & \textcolor{ourgreen}{\ding{51}} & \textcolor{ourred}{\ding{55}} \\
    \midrule
    \begin{tabular}{@{}l@{}}INVERT \\ \citep{bykov2024labeling}\end{tabular} & compositional & scalar & AUC & multi-labeled data & \textcolor{ourgreen}{\ding{51}} & \textcolor{ourgreen}{\ding{51}} \\
    \midrule
    \begin{tabular}{@{}l@{}}Describe-and-Dissect (DnD) \\ \citep{bai2025interpreting}\end{tabular} & free-form & scalar & avg. rank & I2T, LLM, T2I &  \textcolor{ourgreen}{\ding{51}} & \textcolor{ourgreen}{\ding{51}} \\
    \midrule
    \begin{tabular}{@{}l@{}}\LINE{} \\ (Ours)\end{tabular} & free-form & scalar & avg. activation & LLM, T2I & \textcolor{ourgreen}{\ding{51}} & \textcolor{ourgreen}{\ding{51}} \\
    \bottomrule
    \end{tabular}%
}
\end{table}

\textbf{Explanation Format:} The output format determines the expressive capacity of the explanation. Early approaches rely on \textit{fixed-label} sets, which restrict the vocabulary to predefined concepts. While \textit{open-vocabulary} methods (e.g., CLIP-Dissect) expand this range using vision-language models (VLMs), they still require the user to provide a candidate concept vocabulary. \textit{Compositional} approaches further increase complexity by building logical formulas from base concepts. Finally, \textit{free-form} methods, including \LINE{}, leverage the generative power of LLMs to produce unconstrained textual descriptions capable of capturing highly nuanced or abstract neuron behaviors.

\textbf{Neuron Type:} The unit of analysis is often dictated by the model architecture. Early methods (e.g., NetDissect) primarily analyze spatial \textit{convolutional} feature maps. To maintain relevance for modern architectures like Vision Transformers (ViTs), recent methods target specific \textit{scalar} activations within a layer. Notably, while most methods are architecture-agnostic, some (e.g., FALCON) remain restricted to \textit{predetermined} structures required by the tools like Grad-CAM~\citep{Selvaraju_2017_ICCV}.

\textbf{Optimization Target:} To map neurons to semantic concepts, methods optimize specific alignment metrics. Spatial grounding methods typically use \textit{Intersection over Union (IoU)} to measure overlap with segmentation masks. Other statistical approaches utilize continuous metrics such as \textit{Weighted Pointwise Mutual Information (WPMI)} or \textit{Area Under the Curve (AUC)}. Generative and language-guided methods, such as \LINE{} and DnD, utilize \textit{average activation} or \textit{average rank} to measure the correlation between the neuron's peak firing patterns and the synthesized concept.

\textbf{Method Dependencies and Constraints:} Requirement profiles vary significantly across methods. Fixed-label approaches historically necessitated expensive, dense \textit{segmentation masks} or \textit{multi-labeled corpora}, while MILAN requires a labeled corpus for training its LSTM-based captioner. In contrast, our framework leverages off-the-shelf large language models (LLMs) and text-to-image (T2I) models, ensuring it is strictly \textit{training-free} and \textit{architecture-agnostic}. This modularity allows for future scaling, as each component (LLM or T2I) can be upgraded in parallel with state-of-the-art advancements.

\subsection{CoSy Benchmark Methodology}
\label{app:cosy}

\begin{figure}[h]
    \centering\includegraphics[width=1.0\textwidth]{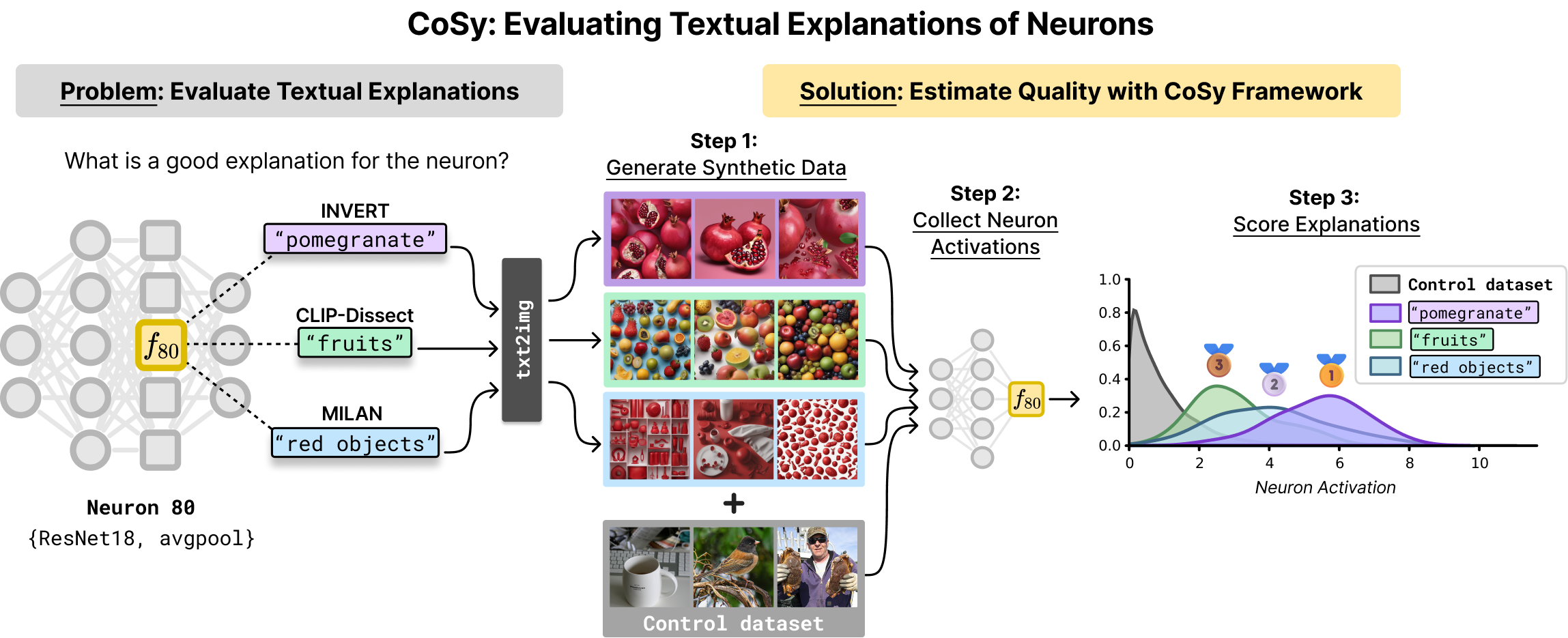}
    \caption{\textbf{CoSy evaluation framework.} A schematic illustration of the CoSy framework for Neuron 80 in ResNet18’s \texttt{avgpool} layer. The figure is sourced from the original paper~\citep{kopf2024cosy}.}
    \label{fig:cosy}
\end{figure}

The CoSy benchmark~\citep{kopf2024cosy} evaluates the quality of open-vocabulary textual explanations for vision model neurons. Given the difficulty of finding natural datasets that perfectly isolate arbitrary concepts, CoSy leverages T2I generative models to synthesize validation data, similarly to how \LINE{} generates concept images. The assumptions of CoSy are that if a concept label $t$ accurately explains a neuron $n$, the neuron should activate significantly more on generated images representing $t$ than on a baseline control (ImageNet~\citep{ILSVRC15} or Places365~\citep{zhou2017places} validation) set of random natural images.

The evaluation process, which is illustrated in \Cref{fig:cosy}, follows three steps:
\begin{enumerate}
    \item \textbf{Generate Synthetic Data.} Given a proposed concept label $t$ (e.g., \textit{polka dots}), a generative T2I model is used to synthesize a collection of images, denoted in our work as $\mathcal{P}_t$.
    \item \textbf{Collect Neuron Activations.} Both the synthetic images $\mathcal{P}_t$ and a control dataset of natural images $\mathcal{X}_{\text{control}} \subset \mathcal{X}$ are passed through the target vision network $f$ and the activations are extracted from the specific neuron $n$. This yields two sets of scalar activations:
    \begin{itemize}
        \item $A_{t}$: Activations from the synthetic concept images in $\mathcal{P}_t$.
        \item $A_{\text{control}}$: Activations from the natural control images in $\mathcal{X}_{\text{control}}$.
    \end{itemize}
    \item \textbf{Score Explanations.} A scoring function $\psi(A_{\text{control}}, A_t)$ is used to quantify the difference between the two activation distributions. A higher score indicates that the concept $t$ is a better match for the neuron $n$.
\end{enumerate}

The benchmark evaluates these explanations using two complementary scoring functions that capture different aspects of neuron behavior: \\
\noindent \textbf{Area Under the ROC Curve (AUC):} A non-parametric metric evaluating the neuron's ability to act as a binary classifier. It represents the probability that a synthetic concept image activates the neuron more strongly than a random control image:
\begin{equation}
    \psi_{\text{AUC}}(A_{\text{control}}, A_t) = \frac{1}{|A_{\text{control}}| \cdot |A_t|} \sum_{a \in A_{\text{control}}} \sum_{b \in A_t} \mathbf{1}[a < b]
\end{equation}
where $\mathbf{1}[\cdot]$ is the indicator function. AUC is robust to outliers. \\

\textbf{Mean Activation Difference (MAD):} A parametric metric measuring the magnitude of the activation shift. It calculates the difference in mean activations between the synthetic and control images, normalized by the control's standard deviation:
\begin{equation}
    \psi_{\text{MAD}}(A_{\text{control}}, A_t) = \frac{\mu(A_t) - \mu(A_{\text{control}})}{\sigma(A_{\text{control}})}
\end{equation}
where $\mu$ and $\sigma$ denote the mean and standard deviation. MAD quantifies the absolute extent to which concept $t$ maximizes the neuron's output.

\subsection{Simulation Evaluation Methodology}
\label{app:siglip}

\citet{oikarinen2024linear}~propose evaluating the neuron explanations by measuring how well a description can simulate the true activations. The simulation evaluation pipeline proceeds as follows:
\begin{enumerate}
    \item Generate a human understandable explanation $t$ for the neurons using an \textit{Explainer} (in our case \LINE{}, INVERT, CLIP-Dissect, or DnD), where $t$ is the final neuron description \textit{Explainer} returns.
    \item Use a \textit{Simulator} to predict the neuron activations $A_t(X)$ on the evaluation set of images $X$, given access only to the text explanation $t$.
    \item Score how well the simulated activations $\mathrm{sim}(X, t)$ fit the actual neuron activations $A_t(X)$, using the Pearson correlation $\mathrm{Corr}(A_t(X), \mathrm{sim}(X, t))$.
\end{enumerate}
Following the original paper, we use SigLIP-SO400M-14-384~\citep{zhai2023sigmoid} as the \textit{Simulator} and the corresponding dataset (ImageNet~\citep{ILSVRC15} or Places365~\citep{zhou2017places} validation) as the probe $X$.

\subsection{Baseline Implementation Details}
\label{app:baseline_implementation}

We use the following publicly available source code:
\begin{itemize}
    \item CoSy~\citep{kopf2024cosy}: \\ \url{https://github.com/lkopf/cosy} (License unknown)
    \item INVERT~\citep{bykov2024labeling}: \\ \url{https://github.com/lapalap/invert} (License unknown)
    \item CLIP-Dissect~\citep{oikarinen2023clip}: \\ \url{https://github.com/Trustworthy-ML-Lab/CLIP-dissect} (License unknown)
    \item DnD~\citep{bai2025interpreting}: \\ \url{https://github.com/trustworthy-ml-lab/describe-and-dissect} \\ (License unknown)
\end{itemize}

When evaluating the INVERT and CLIP-Dissect baselines, we observed that CoSy augmented the methods' default vocabularies with ImageNet and Places365 class labels, thereby boosting performance. To ensure fairness and standardized evaluation, we restrict both methods to their original default vocabularies, omitting the dataset-specific labels added by CoSy. In the case of INVERT, the vocabulary is the same as the one used by \LINE{} (Imagenet-1k class names) unlike CLIP-Dissect which uses 20,000 most common English words. INVERT uses parameter $L$ to specify number of concepts to combine. We use $L=1$ as suggested in both the INVERT and CoSy papers.  We evaluate DnD using \texttt{gpt-3.5-turbo}, which is still more powerful than the one used by \LINE{}\footnote{https://llm-stats.com/models/compare/gpt-3.5-turbo-0125-vs-llama-3.1-8b-instruct}, to match the original paper. Additionally, during our evaluation, we identified a bug in DnD's attention cropping implementation for ViTs. However, given DnD's performance on CNNs, we do not expect that resolving this bug would bridge the performance gap with \LINE{}.

\clearpage
\section{\LINE{} Implementation Details}
\subsection{\LINE{} Algorithm}
\label{app:pseudocode}

In addition to the detailed description in \Cref{sec:line} and the overview schema in \Cref{fig:schema}, we include the complete pseudocode for our iterative concept proposal and scoring algorithm in \Cref{alg:pipeline}. This outlines the exact step-by-step process \LINE{} uses to initialize the search space, generate and evaluate potential concepts, and extract the best concept description.

\begin{algorithm}
    \caption{\LINE{} Pipeline}
    \label{alg:pipeline}
    \SetAlgoNoLine
    \DontPrintSemicolon

    \SetKwFunction{FSetup}{initialize\_scoreboard}
    \SetKwFunction{FGenCon}{propose\_concept}
    \SetKwFunction{FGenSynImg}{generate\_images}
    \SetKwFunction{FColAct}{extract\_activations}
    \SetKwFunction{FScCon}{score\_concept}
    \SetKwFunction{FUpd}{update\_scoreboard}
    \SetKwFunction{FRunSumIter}{run\_summary\_step}
    \SetKwFunction{FGetBest}{get\_best\_concept}
    \SetKwBlock{Step}{}{end}
    \SetKwInput{KwIn}{Input}
    \SetKwInput{KwOut}{Output}

    \KwIn{Number of iterations $N$, target neuron $n$}
    \KwOut{Best textual concept $t_\text{best}$}
    \BlankLine
    $\mathcal{H} \gets$ \FSetup{} \tcp*{Initialize scoreboard}

    \For{$i \in \{1,\dots,N\}$}
    {
        $t \gets$ \FGenCon{$\mathcal{H}$} \tcp*{LLM proposes a new concept $t$}
        $\mathcal{P}_t \gets$ \FGenSynImg{$t$} \tcp*{T2I model generates image set $\mathcal{P}_t$}
        $A_t \gets$ \FColAct{$n$, $\mathcal{P}_t$} \tcp*{Collect neuron $n$ activation}
        $s_t \gets$ \FScCon{$A_t$} \tcp*{Calculate concept score $s_t$}
        $\mathcal{H} \gets$ \FUpd{$t$, $s_t$} \tcp*{Update scoreboard with new score}
    }

    $t_\text{summary}, s_\text{summary} \gets$ \FRunSumIter{$\mathcal{H}$} \tcp*{Final iteration}
    $\mathcal{H} \gets$ \FUpd{$t_\text{summary}$, $s_\text{summary}$} \tcp*{Add the final iteration score}
    $t_\text{best} \gets$ \FGetBest{$\mathcal{H}$} \tcp*{Retrieve highest-scoring concept}
    \BlankLine
    \KwRet{$t_\text{best}$}
\end{algorithm}

\subsection{Prompts}
\label{app:prompts}

The complete set of prompt templates used for both the T2I and LLM components of the \LINE{} pipeline is detailed below.

\subsubsection{T2I Prompt}

To generate diverse images for a given \texttt{concept}, we use the following template:

\begin{tcolorbox}[
  breakable,
  enhanced,
  colback=white,
  colframe=black,
  boxrule=0.5pt,
  title=\textbf{Concept Image Synthesis},
  fonttitle=\bfseries
]
    A realistic photo of a \{concept\}, \{angle\}, \{lighting\}
\end{tcolorbox}

The environmental modifiers are uniformly sampled from the following sets of values:
\begin{itemize}
    \item \texttt{angle}: extreme close-up, wide angle shot, aerial view, low angle;
    \item \texttt{lighting}: cinematic lighting, natural sunlight, studio lighting.
\end{itemize}
To ensure reproducibility, the \LINE{} pipeline is configured to always use fixed generation seeds for the same concepts.

\subsubsection{LLM Prompts}

The LLM in \LINE{} is guided by two distinct prompt templates: one utilized during the main refinement loop across all $N$ iterations, and another exclusively for the final summary step (iteration $S$). To maximize the quality of the generated concepts, we employ standard reasoning techniques such as Chain-of-Thought (CoT) and few-shot prompting. This compels the model to articulate its rationale before predicting the final concept string, guided by in-context examples of the desired reasoning format.

\textbf{Main Loop Prompt:} In the main loop, the current scoreboard $\mathcal{H}$ is provided to the LLM as \texttt{concept\_list}, while all the concepts proposed by the LLM in the previous iterations are supplied in \texttt{generation\_history}. This explicit constraint prevents the LLM from falling into a loop of continuously proposing suboptimal solutions.

\begin{tcolorbox}[
  breakable,
  enhanced,
  colback=white,
  colframe=black,
  boxrule=0.5pt,
  title=\textbf{Main Loop: Iterative Concept Generation},
  fonttitle=\bfseries
]
You are an assistant that analyzes lists of concepts with scores and produces one new concept based on them.

\vspace{4mm}
\noindent \textbf{You MUST follow this output format:}

\noindent \texttt{<thinking>}\\
(Short reasoning. List 3 potential concepts, check them against the forbidden list, and select the best one. Max 3 sentences.)\\
\texttt{</thinking>}\\
\texttt{<answer>}\\
FINAL\_CONCEPT\\
\texttt{</answer>}

\vspace{4mm}
\noindent \textbf{Rules:}
\begin{itemize}
    \item The \texttt{<thinking>} section must be extremely concise (max 8 sentences).
    \item The \texttt{<answer>} section contains only the final generated concept and nothing else.
    \item You are not allowed to generate concepts present in the forbidden list.
    \item You must generate exactly one new concept.
    \item Keep the concept short (1--3 words).
    \item It must relate to the highest-scored concepts.
    \item It must generalize, unify, or creatively combine relevant concepts.
    \item It must not be identical to any forbidden or previously used concept (even if capitalization differs).
    \item You can be guided by the shape of objects.
    \item Do not comment outside the required tags.
    \item The \texttt{<thinking>} section must be short.
\end{itemize}

\vspace{4mm}
\noindent \textbf{Examples:}

\begin{quote}
List of concepts and their scores:
\begin{itemize}
    \item soccer ball: 0.93,
    \item ladybug: 0.90,
    \item dice: 0.87,
    \item dalmatian: 0.84,
    \item mushroom cap: 0.79,
    \item tree leaf: 0.41,
    \item hammer: 0.22,
    \item envelope: 0.18
\end{itemize}
You are not allowed to generate these concepts:\\
animal, insect, toy, tool

\vspace{4mm}
\noindent \texttt{<thinking>}\\
The highest-scoring concepts share a common visual pattern: they all feature black spots on a contrasting background.\\
\texttt{</thinking>}\\
\texttt{<answer>}\\
polka dots\\
\texttt{</answer>}
\end{quote}

\begin{quote}
List of concepts and their scores:
\begin{itemize}
    \item rain: 0.93,
    \item wet: 0.89,
    \item storm: 0.86,
    \item blowing wind: 0.83,
    \item handle: 0.77,
    \item sandwich: 0.25,
    \item rocky mountain: 0.22,
    \item television: 0.14
\end{itemize}
You are not allowed to generate these concepts:\\
coat, water drops, storm, blowing wind

\vspace{4mm}
\noindent \texttt{<thinking>}\\
The high scores for rain, wet, storm, and handle suggest an object used for protection against weather. "Umbrella" fits perfectly as it has a handle and protects from rain. It is not on the forbidden list.\\
\texttt{</thinking>}\\
\texttt{<answer>}\\
umbrella\\
\texttt{</answer>}
\end{quote}

\begin{quote}
List of concepts and their scores:
\begin{itemize}
    \item wooden door: 0.95,
    \item computer screen: 0.92,
    \item old book: 0.89,
    \item windows: 0.87,
    \item dark chocolate: 0.83,
    \item travel suitcase: 0.46,
    \item rug: 0.29,
    \item yellow flower: 0.17
\end{itemize}
You are not allowed to generate these concepts:\\
door, book, window, screen

\vspace{4mm}
\noindent \texttt{<thinking>}\\
The highest-scoring objects all share a clear rectangular outline, defined by straight edges and right angles.\\
\texttt{</thinking>}\\
\texttt{<answer>}\\
rectangle\\
\texttt{</answer>}
\end{quote}

\begin{quote}
List of concepts and their scores:
\begin{itemize}
    \item pomegranate: 0.96,
    \item strawberry: 0.96,
    \item cherry: 0.84,
    \item fruit: 0.77,
    \item seeds: 0.19,
    \item buckeye: 0.46,
    \item aril: 0.55,
    \item jewel: 0.38,
    \item Bursting capsule: 0.392
\end{itemize}
You are not allowed to generate these concepts:\\
fruit, seeds, apple, buckeye

\vspace{4mm}
\noindent \texttt{<thinking>}\\
Pomegranate, strawberry, and cherry are all red fruits. Since "fruit" is forbidden, I need a more specific descriptor. "Red fruit" generalizes these top items accurately and is allowed.\\
\texttt{</thinking>}\\
\texttt{<answer>}\\
red fruit\\
\texttt{</answer>}
\end{quote}

\vspace{4mm}
\noindent \textbf{Task:}

Analyze the new list and produce a concept.\\

\noindent List of concepts and their scores:\\
\{concept\_list\}\\
You are not allowed to generate these concepts:\\
\{generation\_history\} \\

Produce your answer in the required format. Keep thinking very short.
\end{tcolorbox}

\vspace{4mm}
\noindent \textbf{Summary Prompt:}
In the final iteration ($S$), the three highest-scoring concepts from the final scoreboard $\mathcal{H}$ are passed to the LLM as \texttt{concept\_list}. The final prompt is designed to generalize and unify these top-tier concepts into a single high-level concept.

\begin{tcolorbox}[
  breakable,
  enhanced,
  colback=white,
  colframe=black,
  boxrule=0.5pt,
  title=\textbf{Summary Iteration: Concept Summarization},
  fonttitle=\bfseries
]
You are drawing a final conclusion about what a neuron in a computer vision model has learned to detect.
You are given the top-scoring concepts from an iterative search — these are the concepts whose images activated the neuron most strongly.
The feature you are looking for is present in the high-scoring concepts and absent (or weak) in the lower-scoring ones.

\vspace{4mm}
\noindent \textbf{You MUST follow this output format:}

\noindent \texttt{<thinking>}\\
(Reason contrastively: what do the top-scoring concepts share that lower-scoring ones lack? Select the label that most precisely names this distinguishing feature. Max 5 sentences.)\\
\texttt{</thinking>}

\noindent \texttt{<answer>}\\
SUMMARY\_CONCEPT\\
\texttt{</answer>}

\vspace{4mm}
\noindent \textbf{Rules:}
\begin{itemize}
    \item The \texttt{<thinking>} section must be extremely concise (max 5 sentences).
    \item The \texttt{<answer>} section contains only the final concept and nothing else.
    \item The concept must name the distinguishing feature: what the top-scorers have that the lower-scorers lack.
    \item Prefer a label that is specific and falsifiable over one that is broad or vague.
    \item Keep the concept very short (1--3 words).
    \item Do not score the new concept.
    \item Output exactly one concept.
\end{itemize}

\vspace{4mm}
\noindent \textbf{Examples:}

\begin{quote}
List of concepts and their scores:
\begin{itemize}
    \item lava: 0.94
    \item eruption: 0.91
    \item Ash: 0.78
\end{itemize}

\vspace{4mm}
\noindent \texttt{<thinking>}\\
Lava, eruption, and ash are all components or products of a volcanic event — this is what makes them score high together. "Volcano" is the most precise label for this distinguishing feature.\\
\texttt{</thinking>}\\
\texttt{<answer>}\\
volcano\\
\texttt{</answer>}
\end{quote}

\begin{quote}
List of concepts and their scores:
\begin{itemize}
    \item grass: 0.93
    \item crocodile: 0.86
    \item bushes: 0.83
    \item beach: 0.22
\end{itemize}

\vspace{4mm}
\noindent \texttt{<thinking>}\\
Grass, crocodiles, and bushes all have prominent green coloring or texture; beach scores low and lacks this. "Green texture" is the most specific falsifiable label for what the high-scorers share.\\
\texttt{</thinking>}\\
\texttt{<answer>}\\
green texture\\
\texttt{</answer>}
\end{quote}

\begin{quote}
List of concepts and their scores:
\begin{itemize}
    \item banana: 0.96
    \item lemon: 0.90
    \item pineapple: 0.84
\end{itemize}

\vspace{4mm}
\noindent \texttt{<thinking>}\\
Banana, lemon, and pineapple are all yellow fruits — both their color and category distinguish them from non-yellow or non-fruit items. "Yellow fruit" most precisely names this distinguishing feature.\\
\texttt{</thinking>}\\
\texttt{<answer>}\\
yellow fruit\\
\texttt{</answer>}
\end{quote}

\vspace{4mm}
\noindent \textbf{Task:}

Identify the distinguishing feature of the top-scoring concepts — what they share that the lower-scoring ones lack.\\

\noindent List of concepts and their scores:\\
\{concept\_list\}\\

Produce your answer in the required format. Keep thinking very short.
\end{tcolorbox}

\subsection{Detailed Iteration Process}
\label{app:single_loop}

In this section, we provide a complete, step-by-step trace of the \LINE{} framework's refinement process. This trace includes the initial scoreboard (\Cref{tab:start_scorboard}), the concept proposed at each iteration, its corresponding score, the synthesized images (visual explanations), and the LLM's reasoning. These results supplement the final scoreboard presented in \Cref{tab:history_vocab} and trace the behavior of neuron 1255 (ultimately labeled as \textit{weightlifting}) from the ResNet50 Places365 \texttt{avgpool}.

\begin{table}[ht]
    \centering
    \caption{\textbf{Initial Scoreboard for neuron 1255 from ResNet50 \texttt{avgpool}.} We provide the starting scoreboard $\mathcal{H}$ used at the beginning of the \LINE{} refinement process. The scoreboard contains concepts closely related to the final concept (\textit{weightlifting}) like \textit{dumbbell}, alongside lower-scoring, diverse concepts that facilitate broader exploration. We \textbf{bold} the best score and \underline{underline} the second best.}
    \label{tab:start_scorboard}
    \vspace{0.5em}
    \begin{tabular}{l c}
    \toprule
    \textbf{Concept} & \textbf{Score} \\
    \midrule
    pool table & \textbf{0.88}  \\
    dumbbell & \underline{0.72} \\
    barbell & 0.62 \\
    parallel bars & 0.35 \\
    leonberg & 0.27 \\
    christmas stocking & 0.21 \\
    beer glass & 0.19 \\
    pier & 0.17 \\
    leaf beetle & 0.11 \\
    chocolate sauce & 0.06 \\
    \bottomrule
    \end{tabular}%
\end{table}

Following the initialization, \Cref{tab:iteration_process} details the inner refinement loop conducted at each step. \LINE{} first identifies \textit{pool table} as the primary semantic path and proposes \textit{flat surface} as the corresponding concept. It then merges gym-related concepts with the \textit{flat surface}, ultimately yielding the top concept \textit{weightlifting}. However, this trace highlights two limitations: (1) the LLM occasionally stalls refinement by proposing duplicate concepts (e.g., \textit{dumbbell} or \textit{barbell}), and (2) it fixates on the highest-scoring path (weightlifting), neglecting secondary features that could also capture the neuron's inherent polysemanticity. Finally, \Cref{tab:iteration_visualizations} presents the full set of synthesized images per step. We found that for these highly abstract concepts, modifying visual characteristics (e.g., lighting, camera angle) often overshadows the core prompt during T2I generation.

\begin{table}[ht]
    \centering
    \small
    \caption{\textbf{Concept scoreboard for the \textit{weightlifting} neuron.} The table presents the final scoreboard for neuron 1255 from ResNet50 \texttt{avgpool}, tracking the evolution from the initial vocabulary to the final summary. The \textit{Novel} row indicates whether the proposed concept was newly generated by the LLM during the iterative process (\textcolor{ourgreen}{\ding{51}}) or selected from the initialization vocabulary (\textcolor{ourred}{\ding{55}}). Step ``S'' denotes the final summary step. We \textbf{bold} the best score and \underline{underline} the second best.}
    \label{tab:history_vocab}
    \vspace{0.5em}
    \renewcommand{\tabularxcolumn}[1]{m{#1}}
    \begin{tabularx}{\linewidth}{l C C C C C C C}
    \toprule
    \textbf{Label} & pool table & dumbbell & barbell & parallel bars & weightlifting & weight bar & table-based exercise \\
    \midrule
    \textbf{Novel} & \textcolor{ourred}{\ding{55}} & \textcolor{ourred}{\ding{55}} & \textcolor{ourred}{\ding{55}} & \textcolor{ourred}{\ding{55}} & \textcolor{ourgreen}{\ding{51}} & \textcolor{ourgreen}{\ding{51}} & \textcolor{ourgreen}{\ding{51}} \\
    \textbf{Step}  & 0 & 0 & 0 & 0 & 3 & 10 & S \\
    \textbf{Score} & \underline{0.88} & 0.72 & 0.62 & 0.35 & \textbf{1.46} & 0.64 & 0.66 \\
    \bottomrule
    \end{tabularx}%
\end{table}

\begin{table}[ht]
    \centering
    \small
    \caption{\textbf{Iteration process of \LINE{}.} We provide the outputs from the LLM, the T2I, and the scoring function for each iterative step, from which the final scoreboard in \Cref{tab:history_vocab} is built. The best and second-best scores are highlighted in \textbf{bold} and \underline{underlined}, respectively.}
    \label{tab:iteration_process}
    \vspace{0.5em}
    \begin{tabularx}{\linewidth}{c l c X p{1.7cm}}
    \toprule
    \textbf{Step} & \textbf{Concept} & \textbf{Score} & \textbf{Reasoning} & \textbf{T2I Image} \\
    \midrule
    1 & parallel bars & 0.35 & Top concepts all relate to exercise or sports equipment, with a focus on parallel bars. & \includegraphics[width=\linewidth, valign=t]{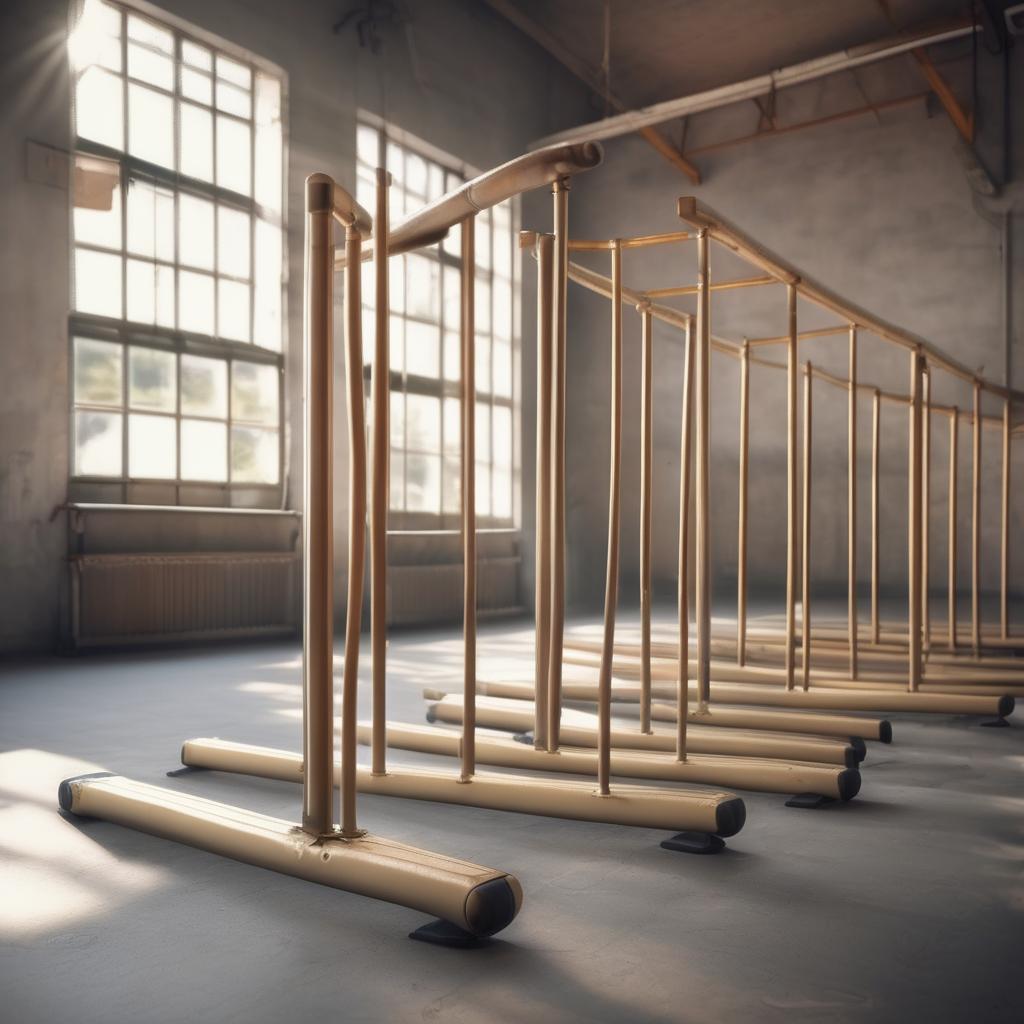} \\
    \midrule
    2 & flat surface & 0.11 & Top concepts relate to objects with flat surfaces and straight edges, often used for exercise or recreation. & \includegraphics[width=\linewidth, valign=t]{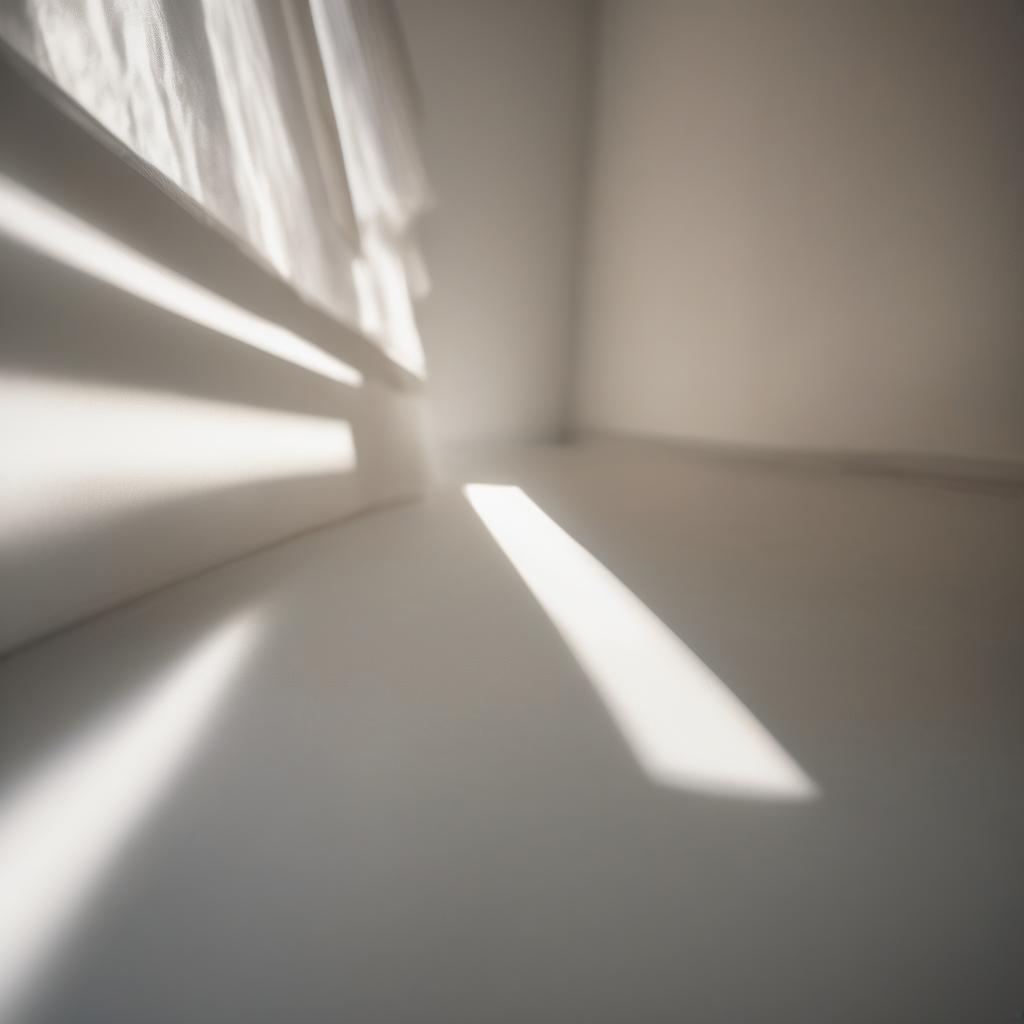} \\
    \midrule
    3 & weightlifting & \textbf{1.46} & High scores for pool table, dumbbell, and barbell suggest exercise equipment used on a flat surface; ``flat surface'' is forbidden, and their common thread is the use of weights. & \includegraphics[width=\linewidth, valign=t]{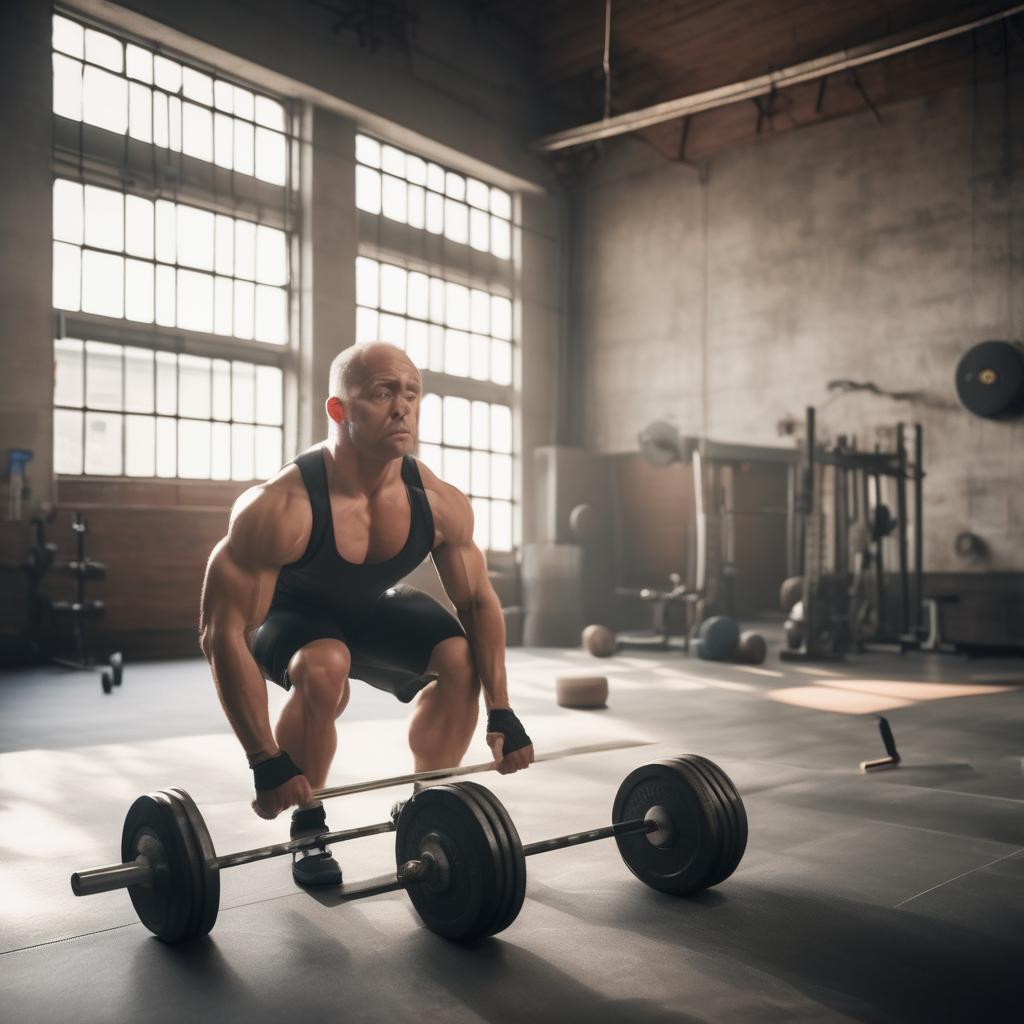} \\
    \midrule
    4 & bar & 0.10 & Top concepts are heavy objects used for exercise or strength training, often with a long handle or bar. & \includegraphics[width=\linewidth, valign=t]{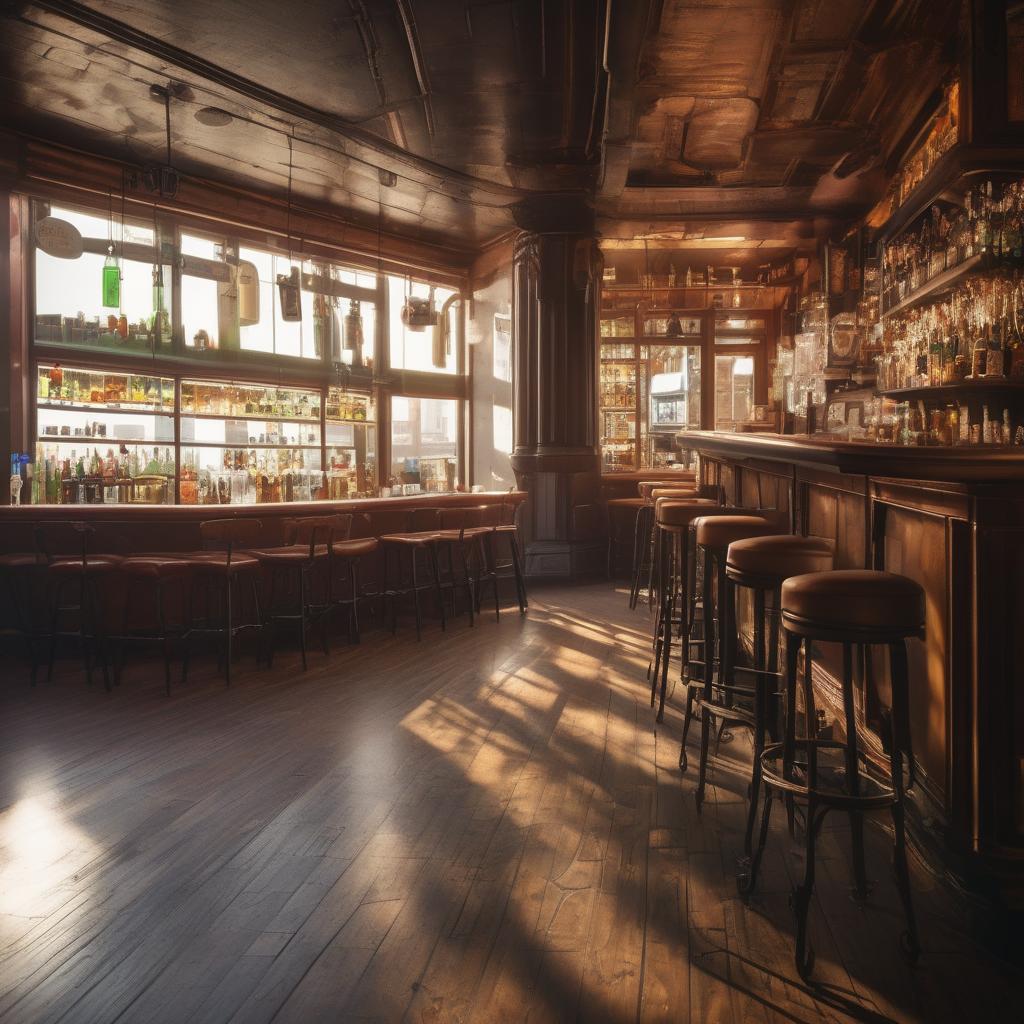} \\
    \midrule
    5 & weighted axis & 0.16 & Top concepts relate to strength training; dumbbell and barbell both feature a central axis with weights attached. & \includegraphics[width=\linewidth, valign=t]{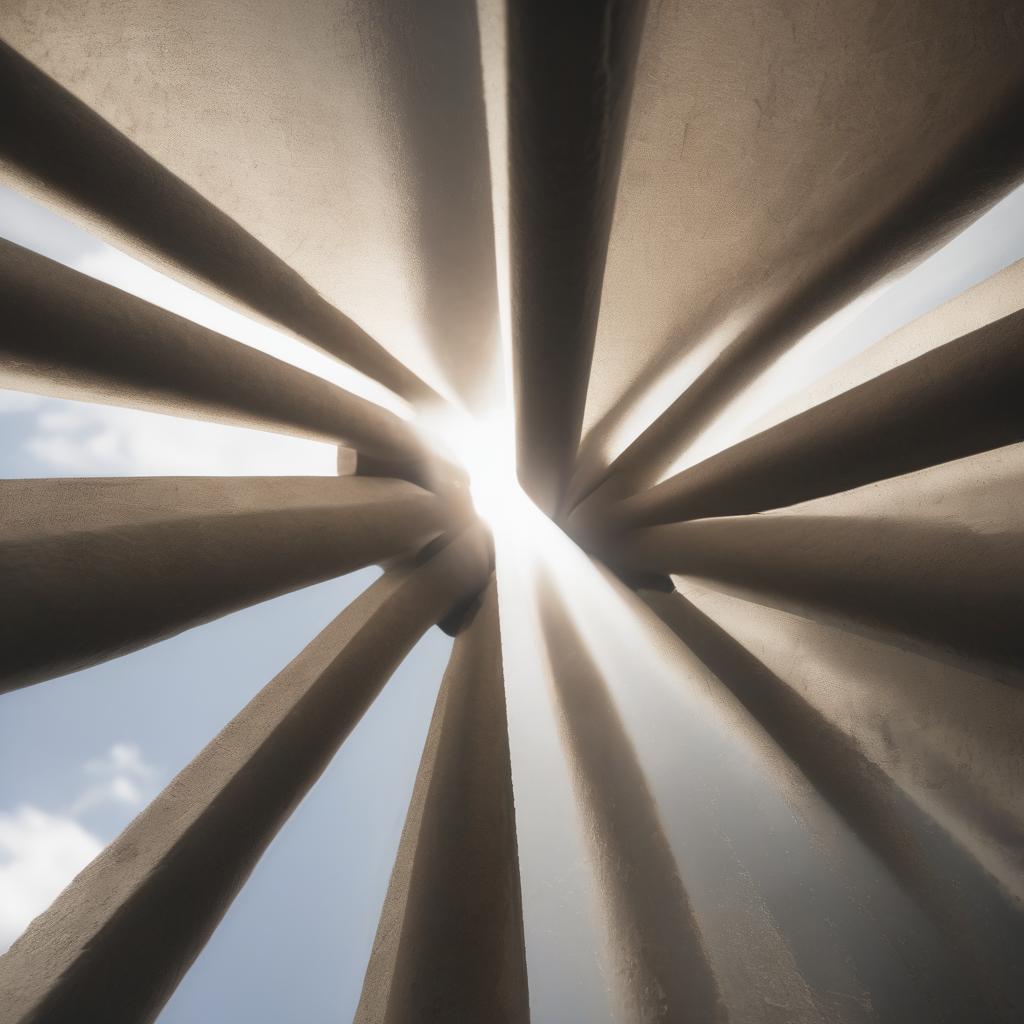} \\
    \midrule
    6 & iron & 0.23 & Dumbbell and barbell indicate heavy strength-training objects; iron is their common material. & \includegraphics[width=\linewidth, valign=t]{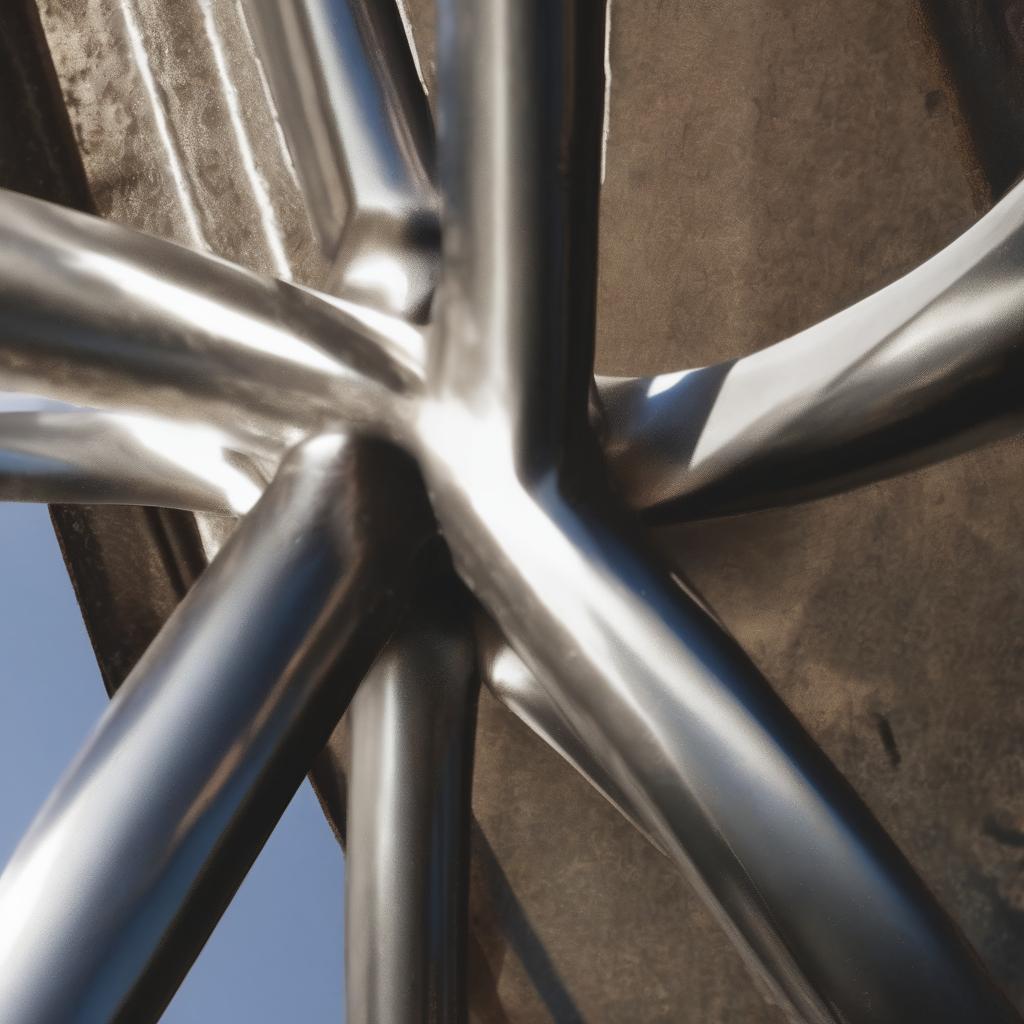} \\
    \midrule
    7 & heavy bar & 0.18 & Top concepts relate to strength training or heavy objects, often with long, straight shapes. & \includegraphics[width=\linewidth, valign=t]{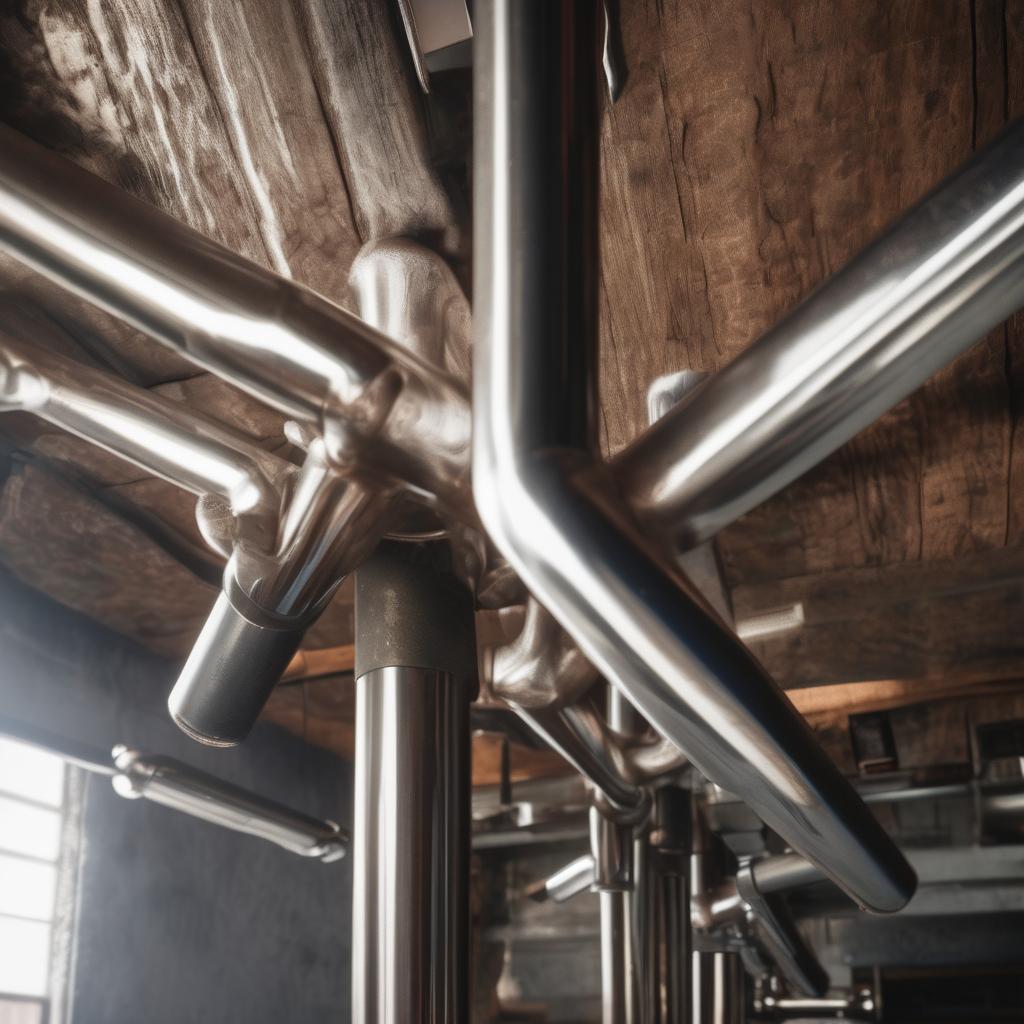} \\
    \midrule
    8 & dumbbell & \underline{0.72} & Top concepts relate to strength training equipment; dumbbell is a common piece of weightlifting equipment. & \includegraphics[width=\linewidth, valign=t]{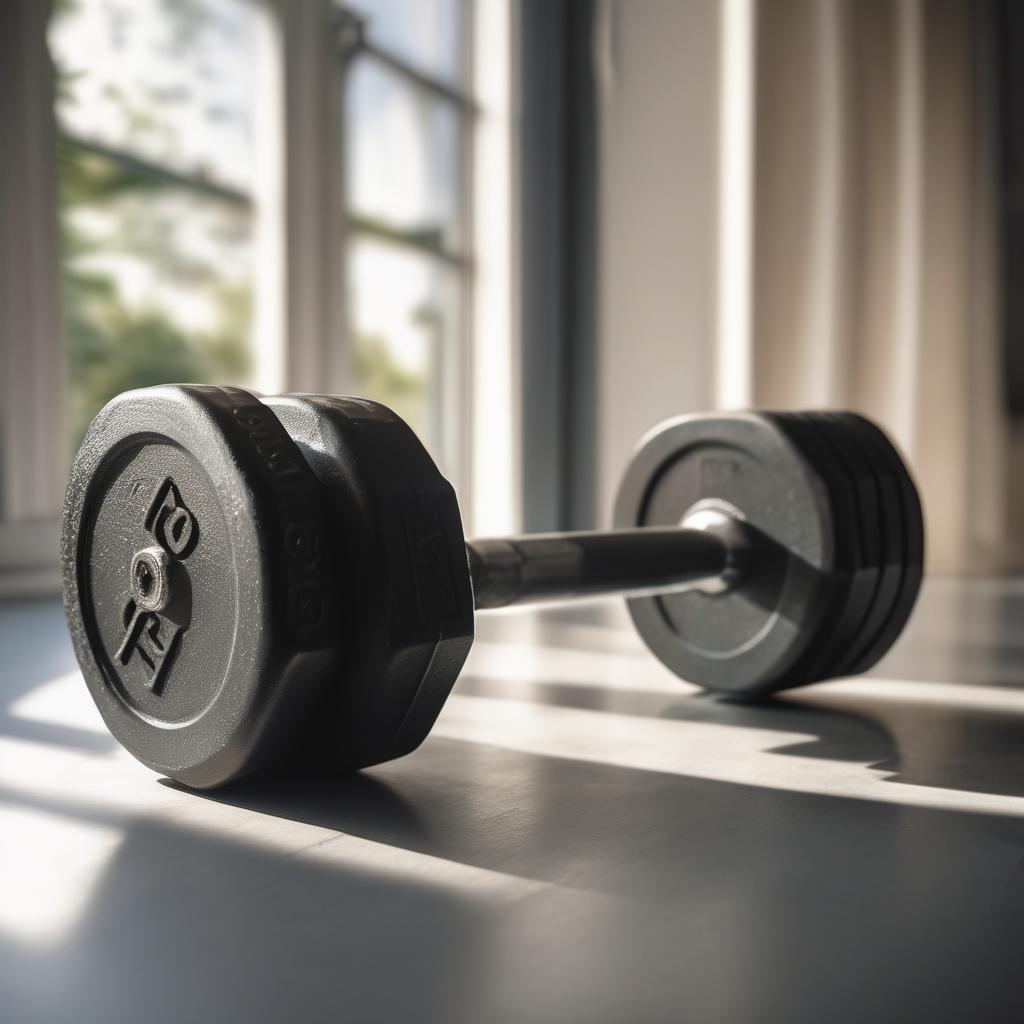} \\
    \midrule
    9 & barbell & 0.62 & Top concepts relate to weights and strength training; the objects are long and thin, often with a handle. & \includegraphics[width=\linewidth, valign=t]{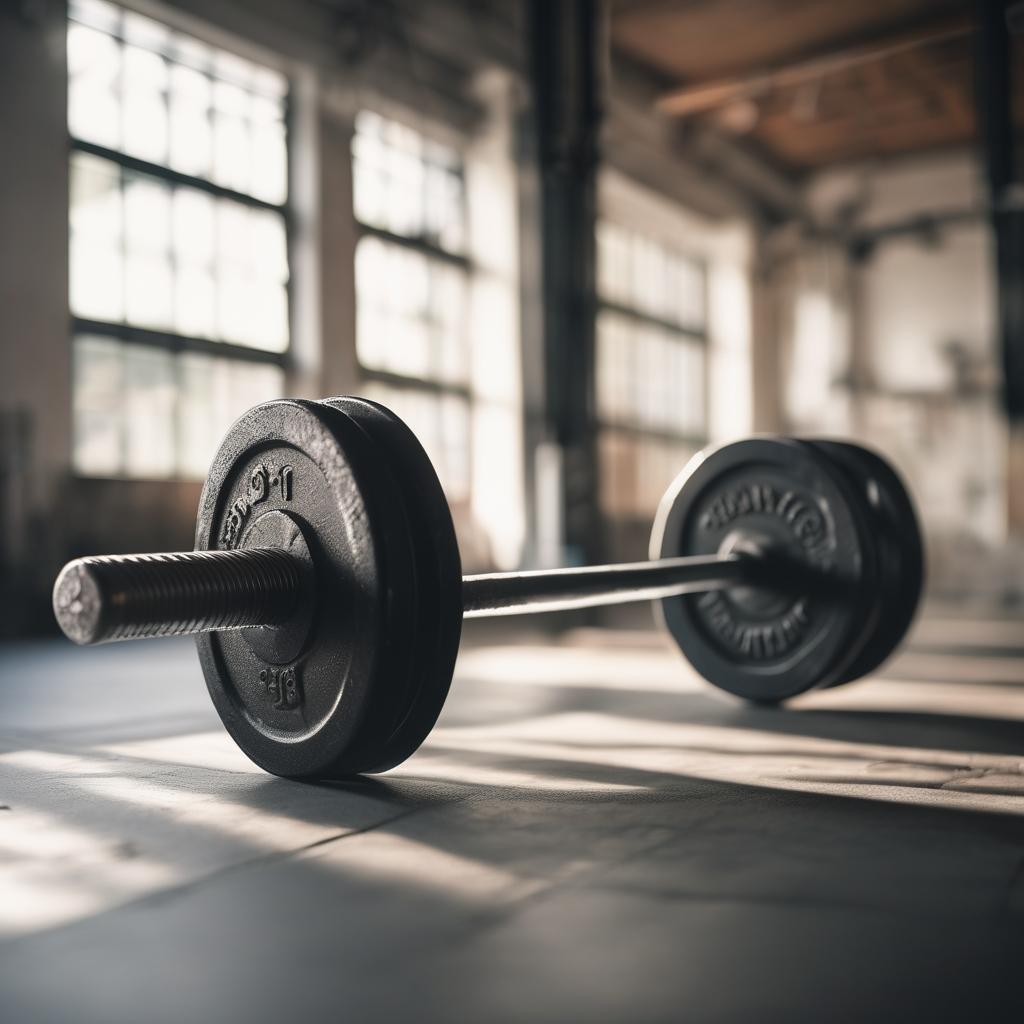} \\
    \midrule
    10 & weight bar & 0.64 & Weightlifting and dumbbell, the highest-scoring concepts, involve lifting heavy objects, commonly with a long, straight object. & \includegraphics[width=\linewidth, valign=t]{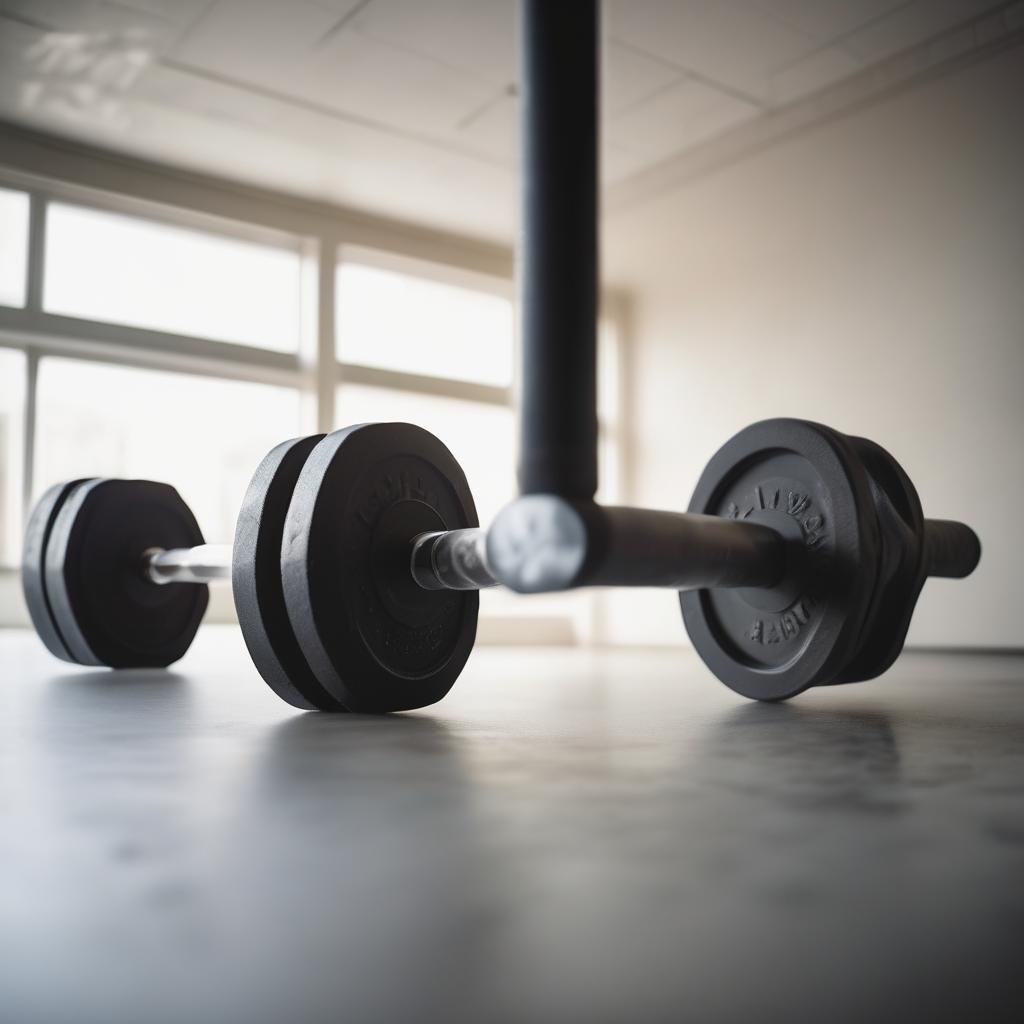} \\
    \midrule
    S & table-based exercise & 0.66 & Weightlifting and pool table both involve physical activity with a table; dumbbell scores low and lacks this. ``Table-based exercise'' most precisely names this distinguishing feature. & \includegraphics[width=\linewidth, valign=t]{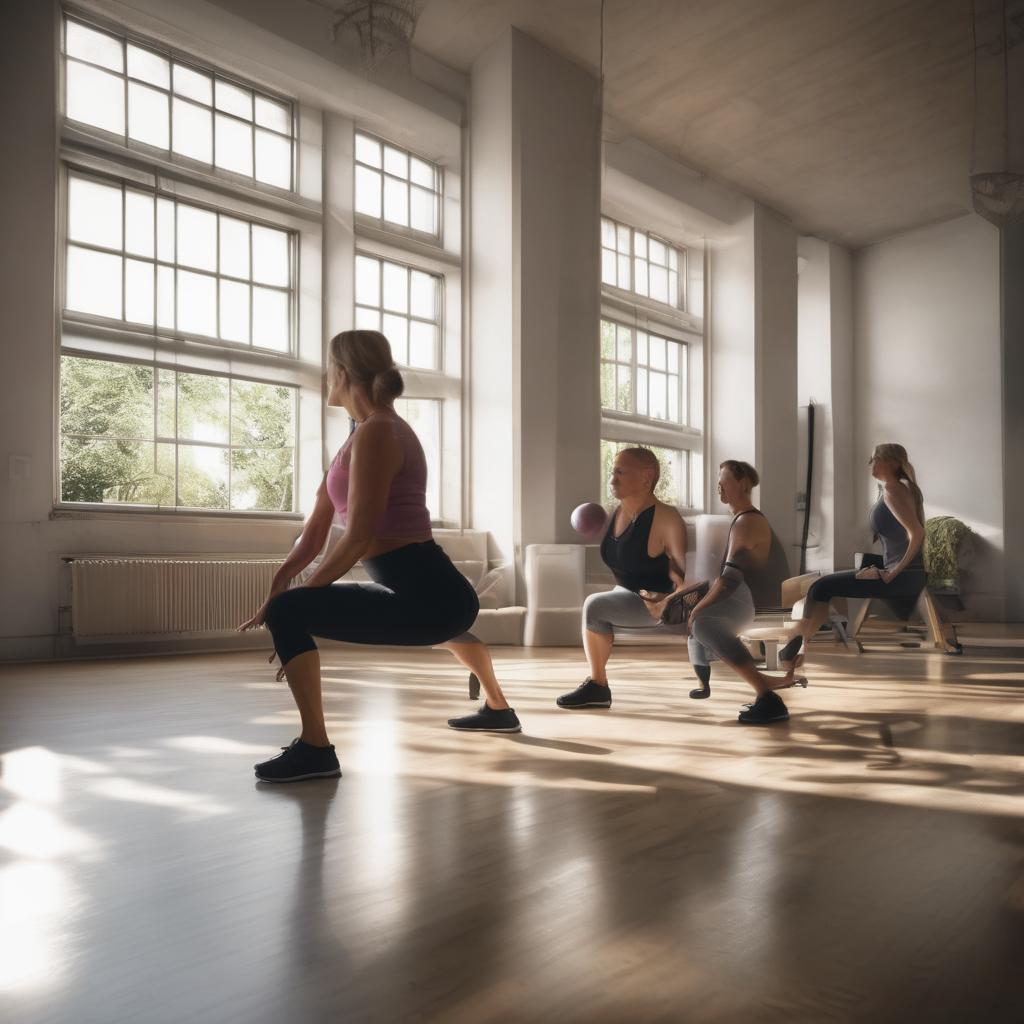} \\
    \bottomrule
    \end{tabularx}
\end{table}

\begin{table}[ht]
    \centering
    \small
    \caption{\textbf{Iteration process of \LINE{} (synthesized images).} Expanding on \Cref{tab:iteration_process}, we present the full set of images generated by the T2I model for each step.}
    \label{tab:iteration_visualizations}
    \vspace{0.5em}
    \begin{tabularx}{\linewidth}{c p{3cm} C C C C C}
    \toprule
    \textbf{Step} & \textbf{Concept} & \multicolumn{5}{c}{\textbf{Synthesized Images}} \\
    \midrule
    1 & parallel bars &
    \includegraphics[width=0.86\linewidth, valign=t]{figures/appendix/neuron_1255_new/iteration_1_parallel_bars/image_1.jpeg} &
    \includegraphics[width=0.86\linewidth, valign=t]{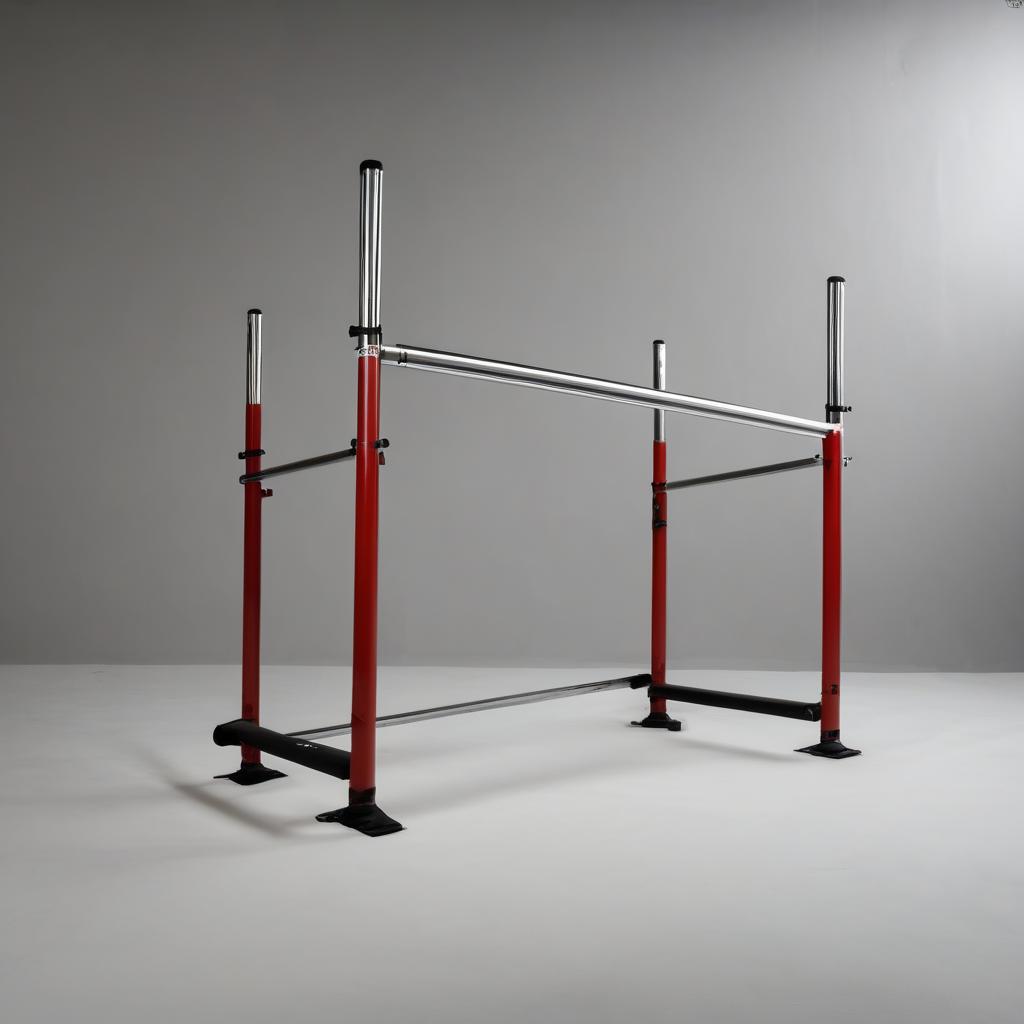} &
    \includegraphics[width=0.86\linewidth, valign=t]{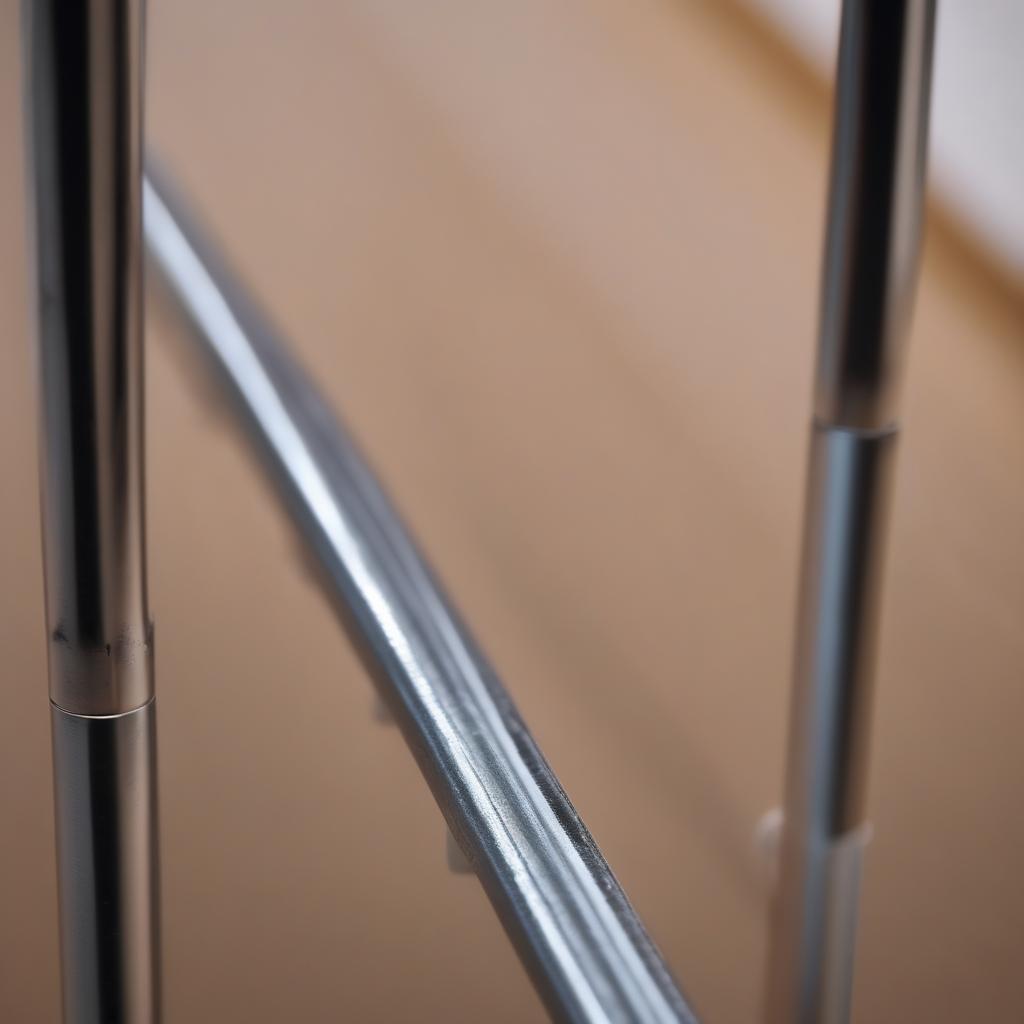} &
    \includegraphics[width=0.86\linewidth, valign=t]{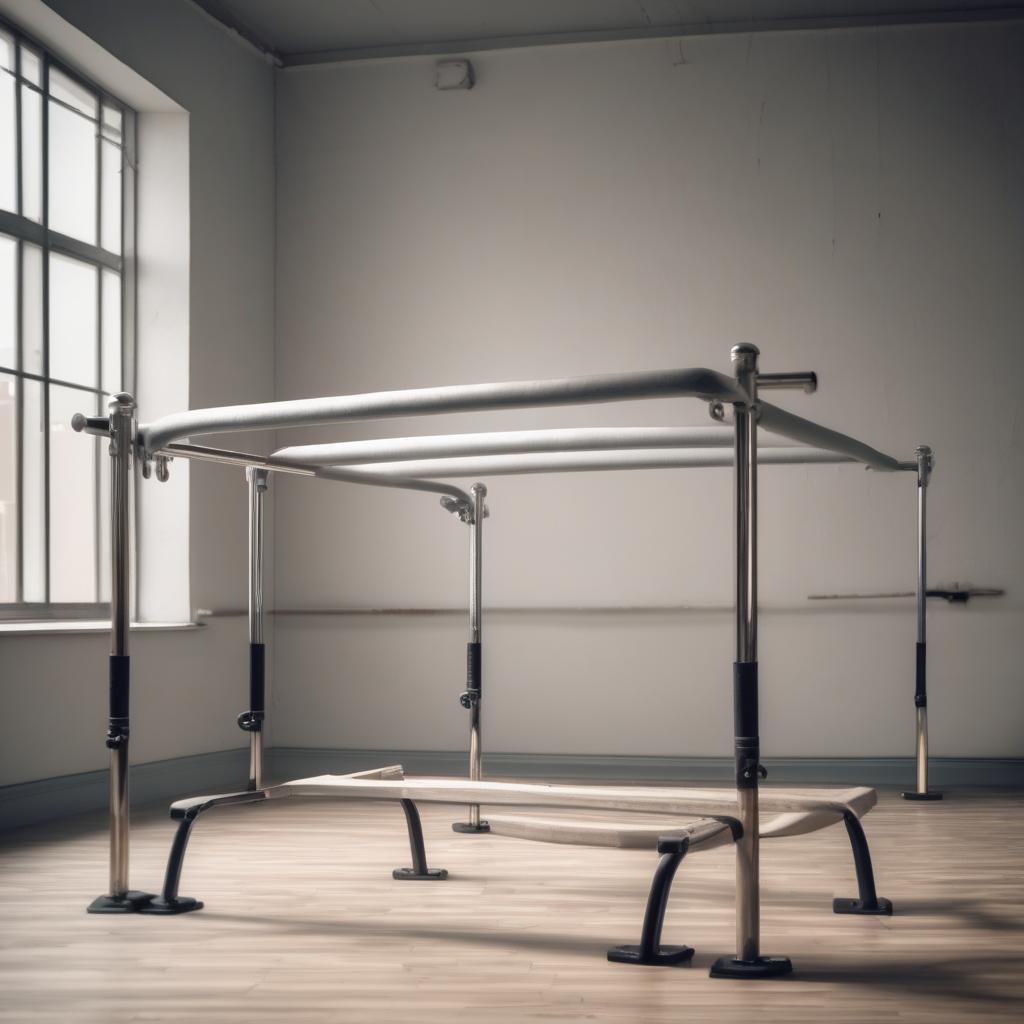} &
    \includegraphics[width=0.86\linewidth, valign=t]{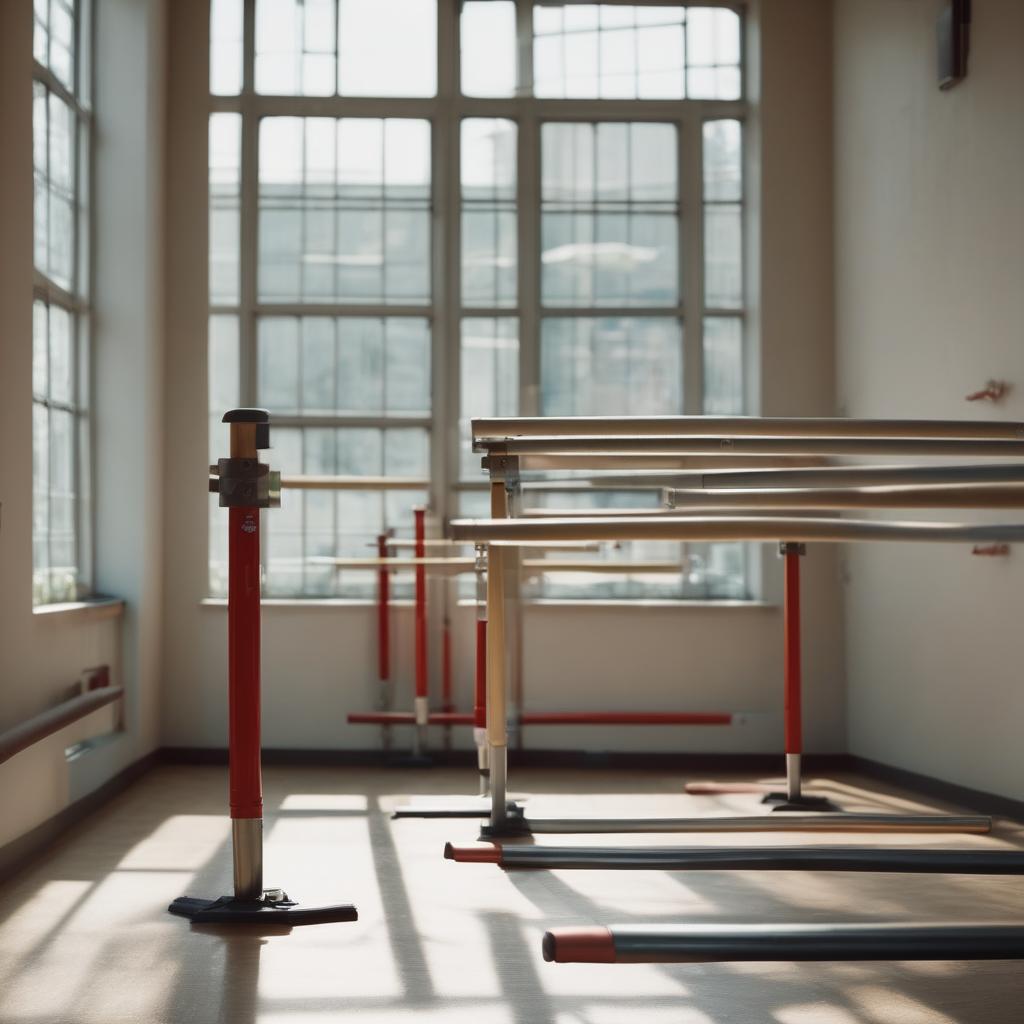} \\
    \midrule
    2 & flat surface &
    \includegraphics[width=0.86\linewidth, valign=t]{figures/appendix/neuron_1255_new/iteration_2_flat_surface/image_1.jpeg} &
    \includegraphics[width=0.86\linewidth, valign=t]{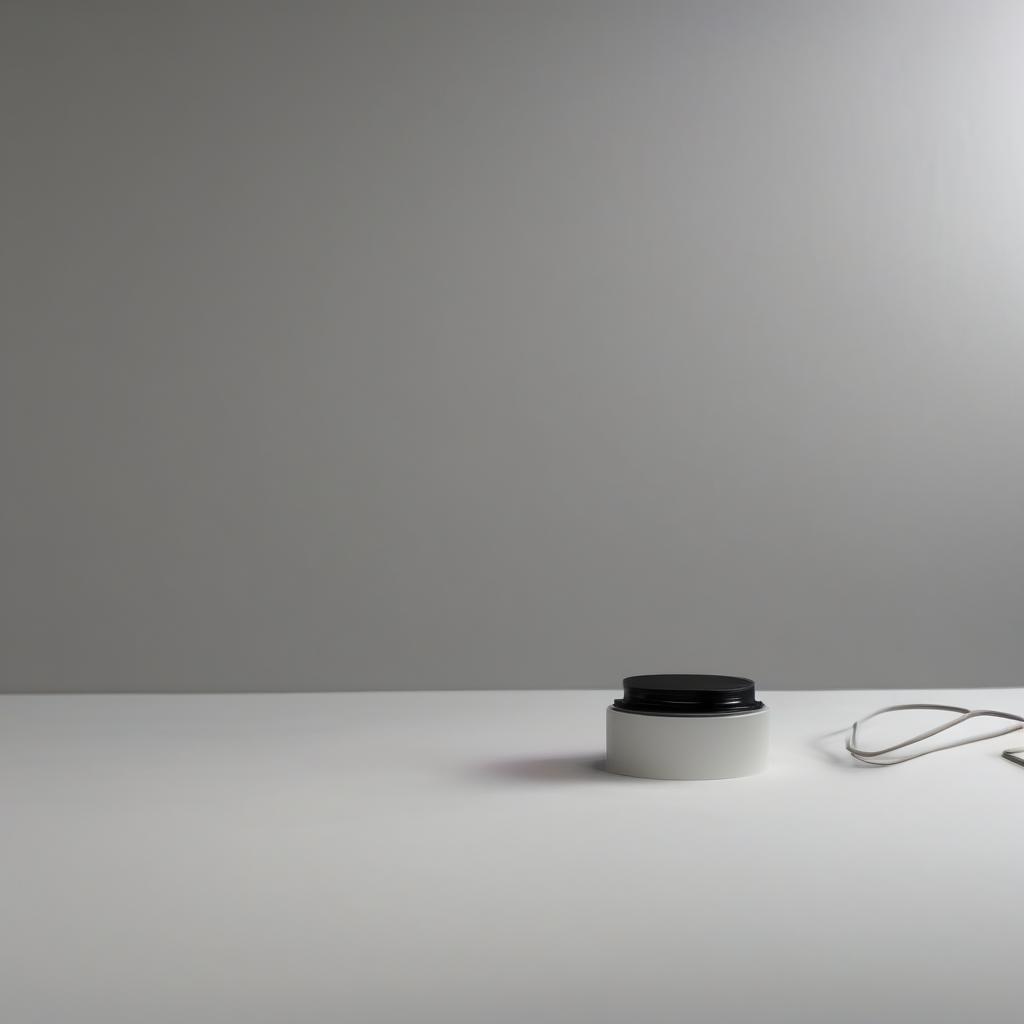} &
    \includegraphics[width=0.86\linewidth, valign=t]{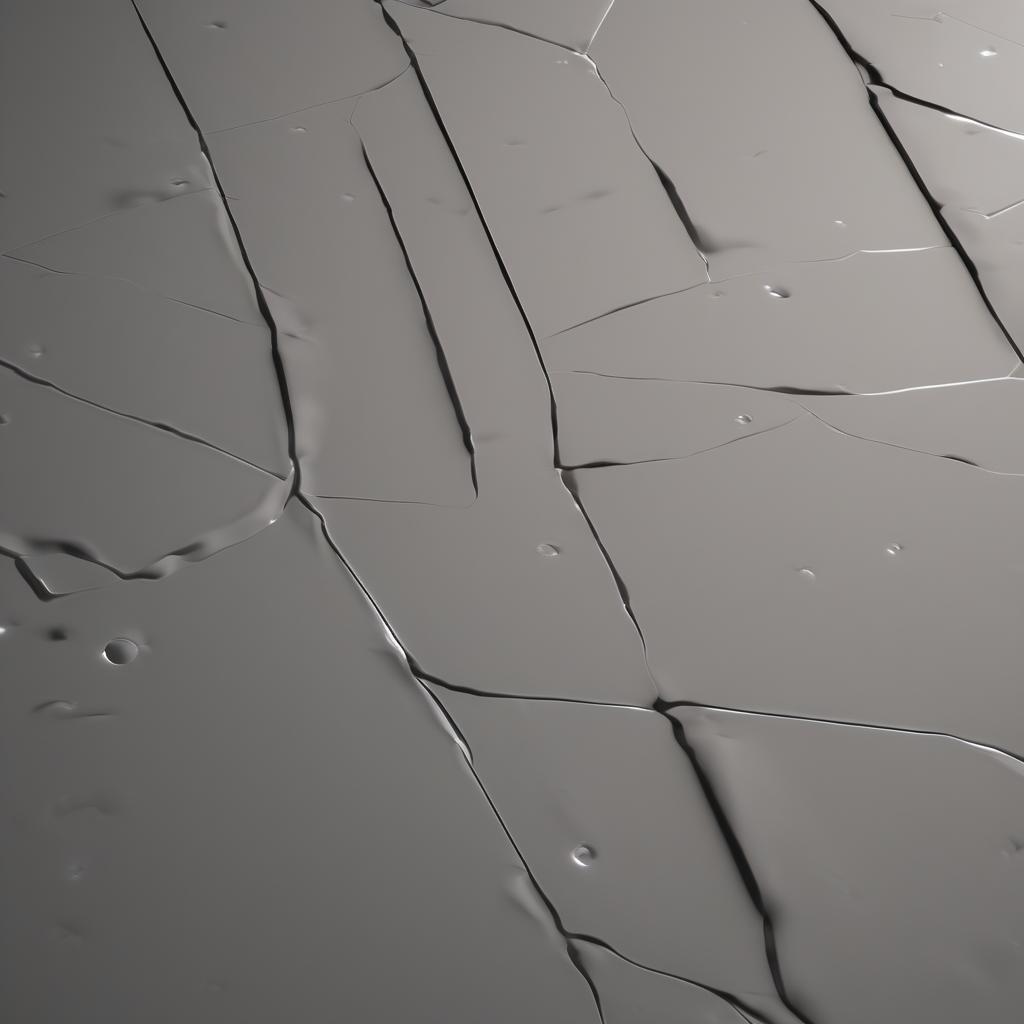} &
    \includegraphics[width=0.86\linewidth, valign=t]{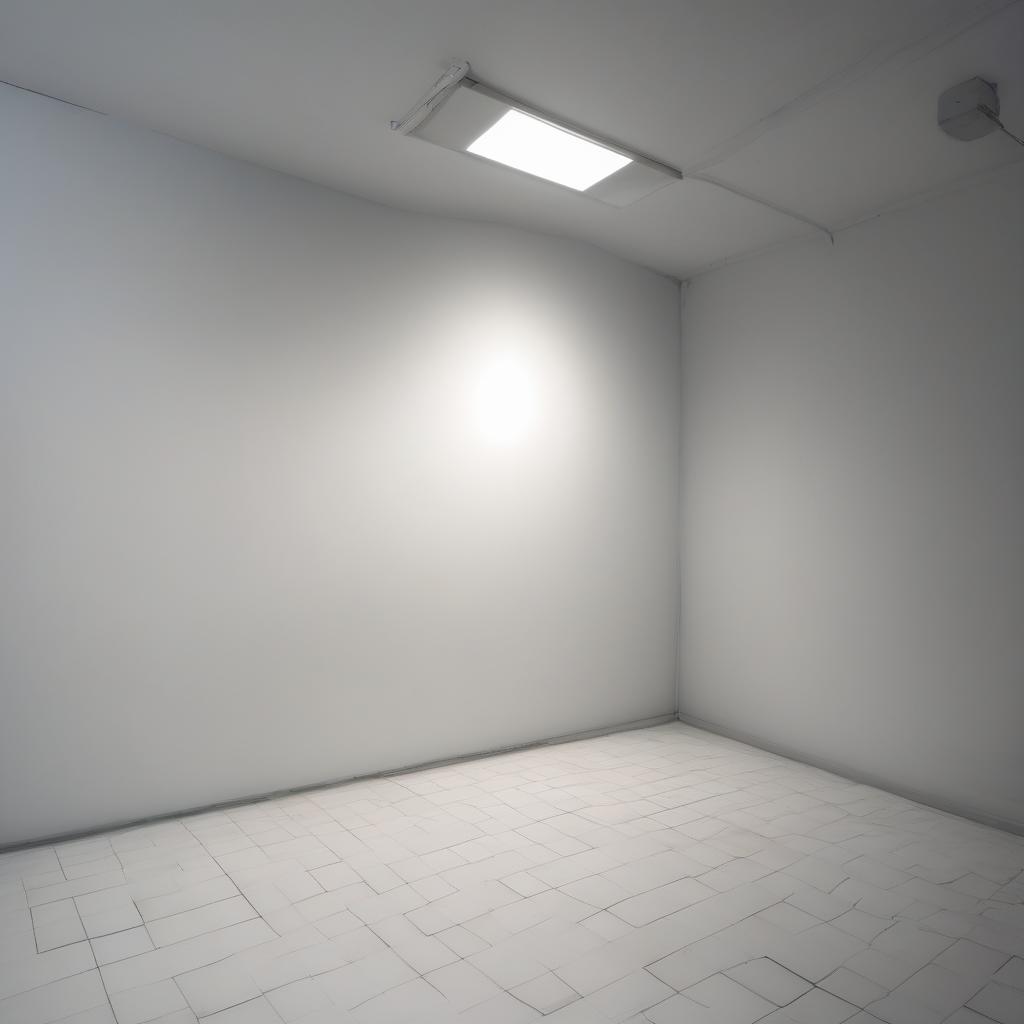} &
    \includegraphics[width=0.86\linewidth, valign=t]{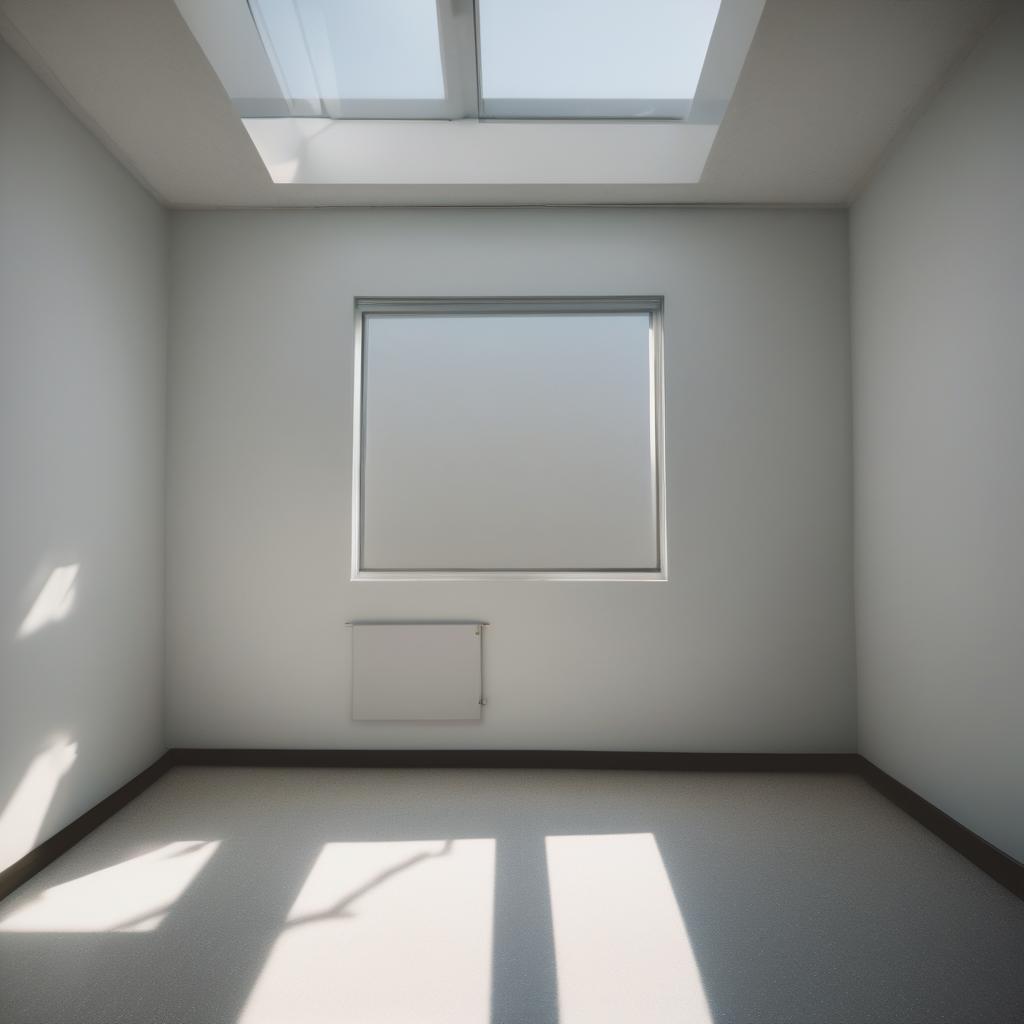} \\
    \midrule
    3 & weightlifting &
    \includegraphics[width=0.86\linewidth, valign=t]{figures/appendix/neuron_1255_new/iteration_3_weightlifting/image_1.jpeg} &
    \includegraphics[width=0.86\linewidth, valign=t]{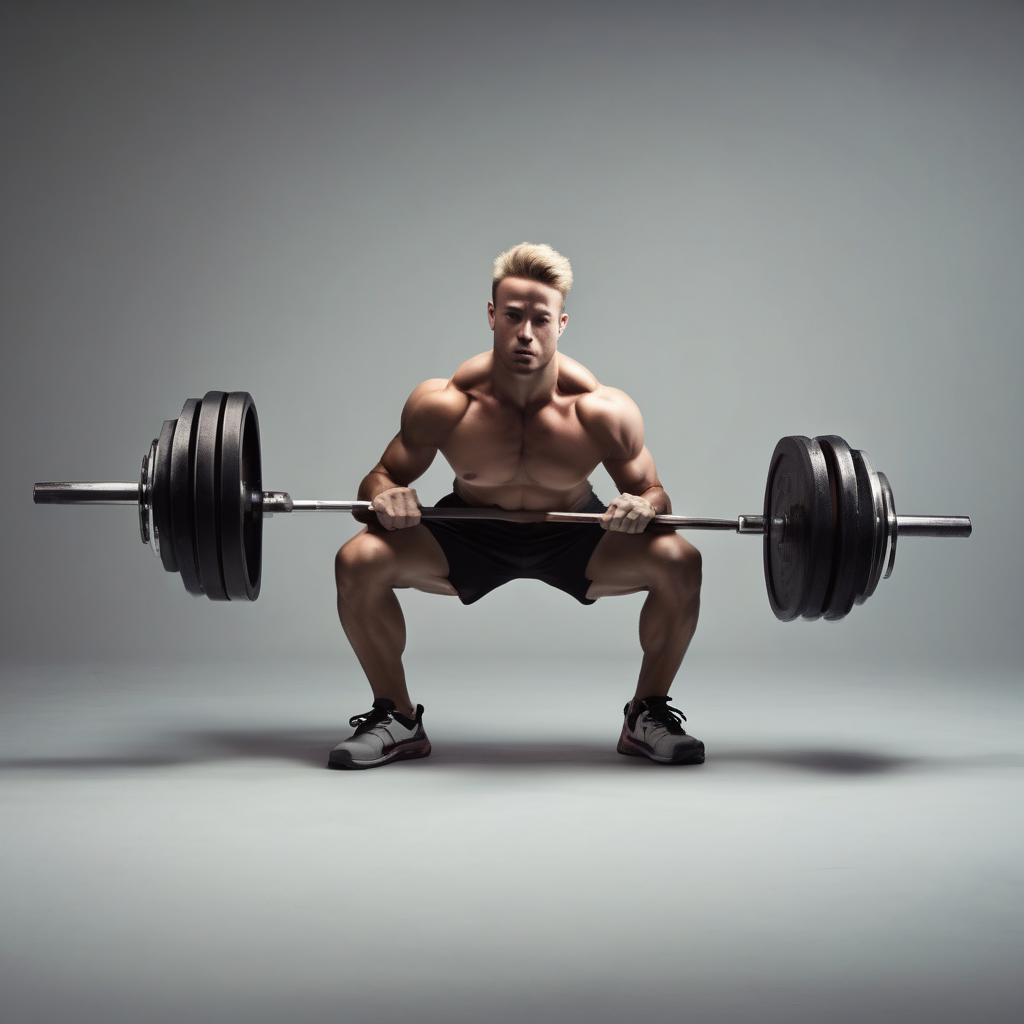} &
    \includegraphics[width=0.86\linewidth, valign=t]{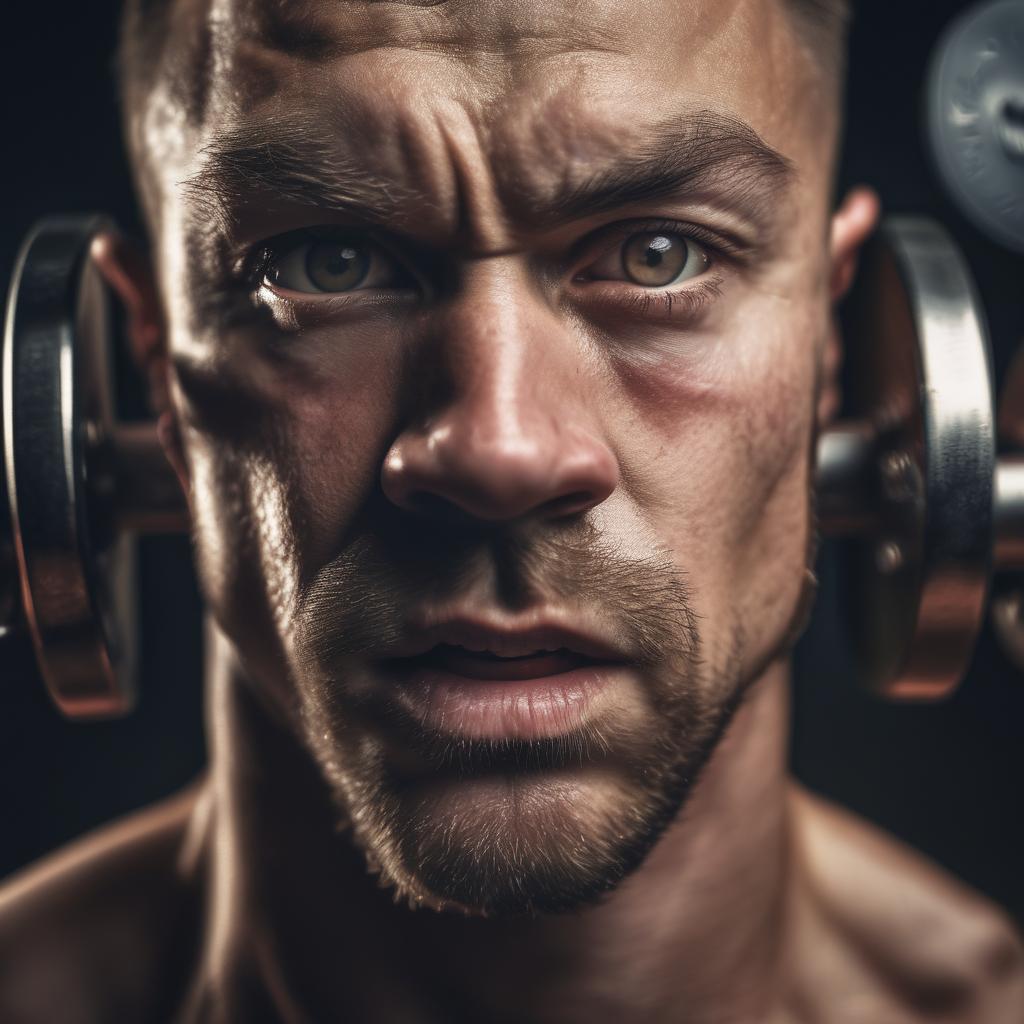} &
    \includegraphics[width=0.86\linewidth, valign=t]{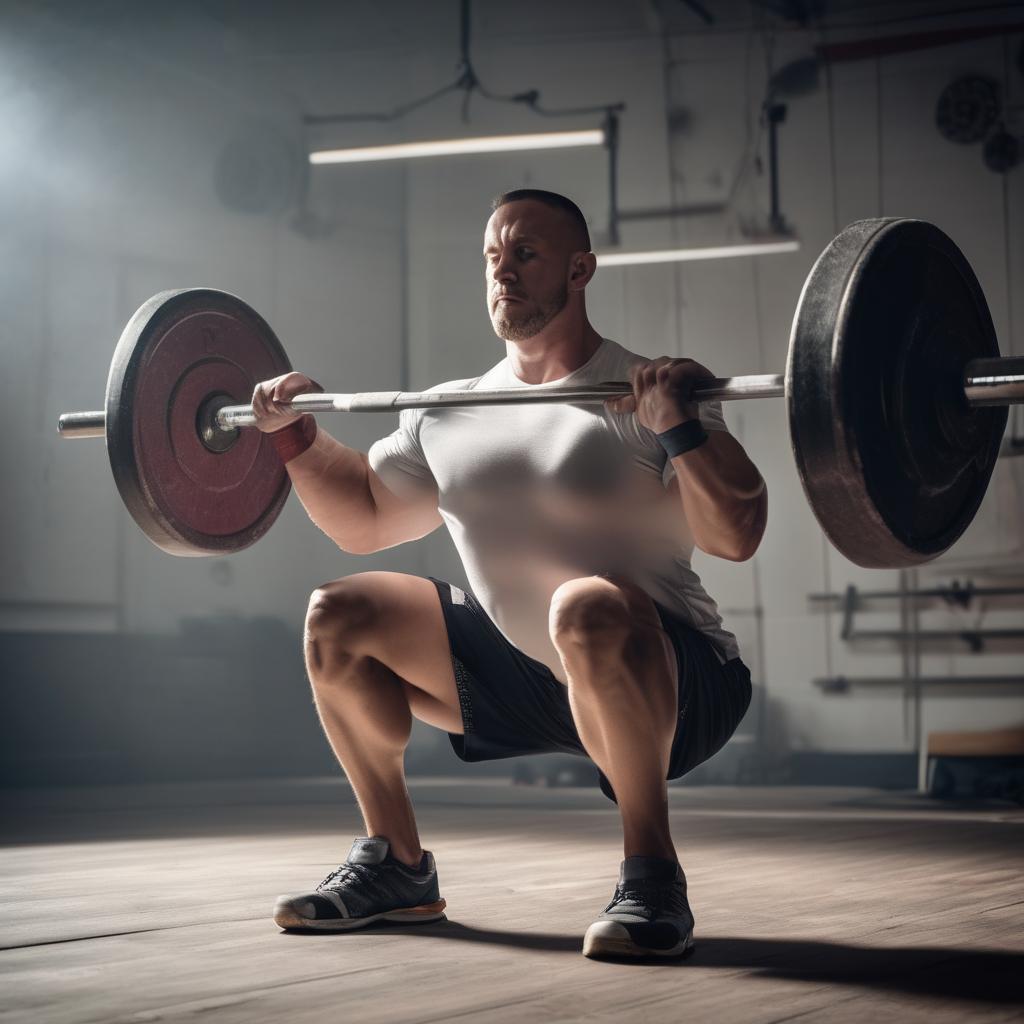} &
    \includegraphics[width=0.86\linewidth, valign=t]{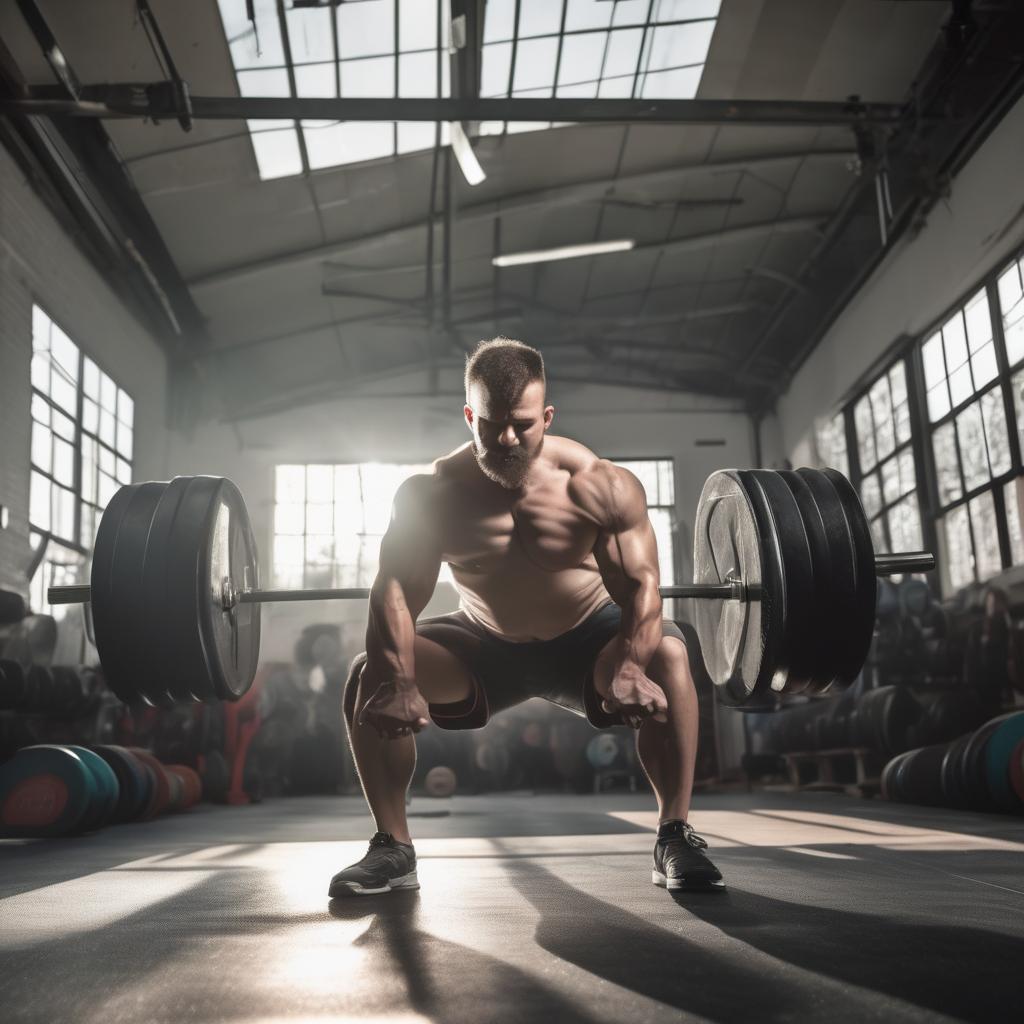} \\
    \midrule
    4 & bar &
    \includegraphics[width=0.86\linewidth, valign=t]{figures/appendix/neuron_1255_new/iteration_4_bar/image_1.jpeg} &
    \includegraphics[width=0.86\linewidth, valign=t]{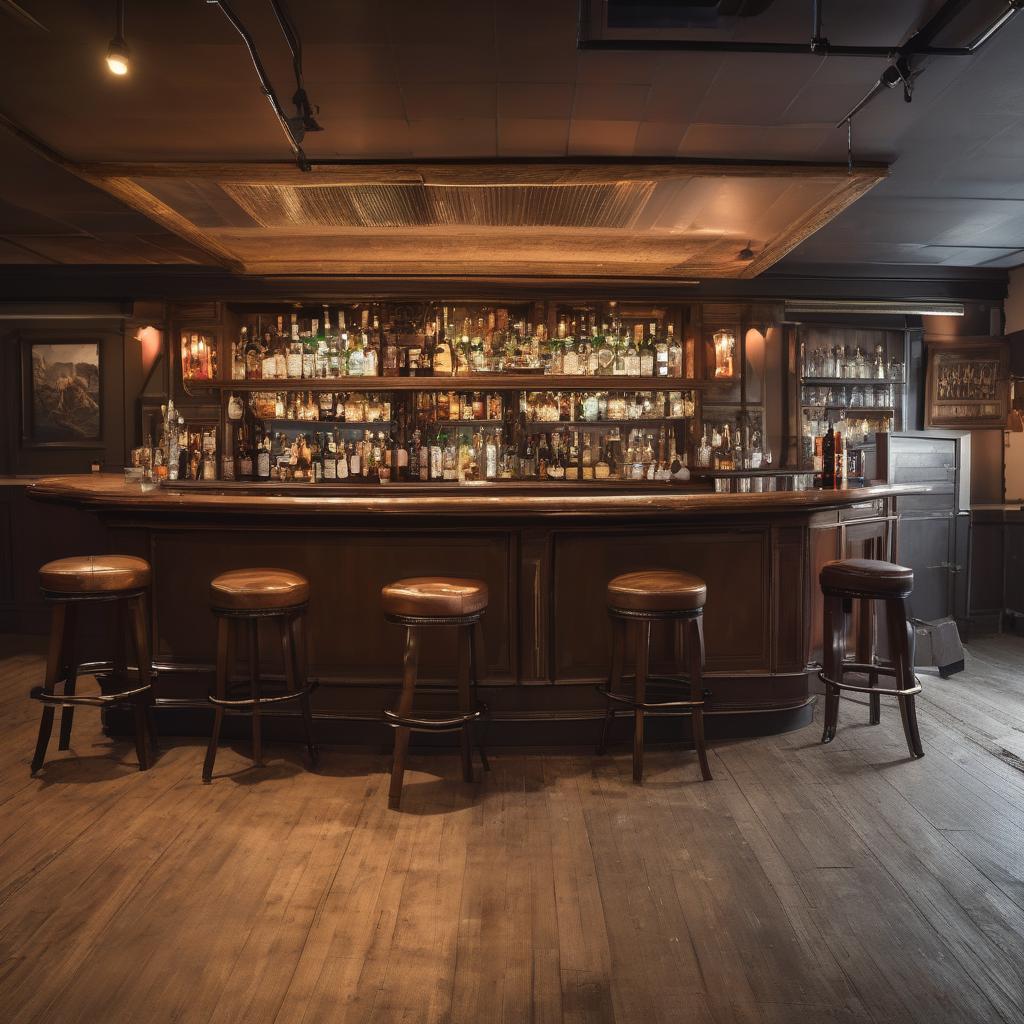} &
    \includegraphics[width=0.86\linewidth, valign=t]{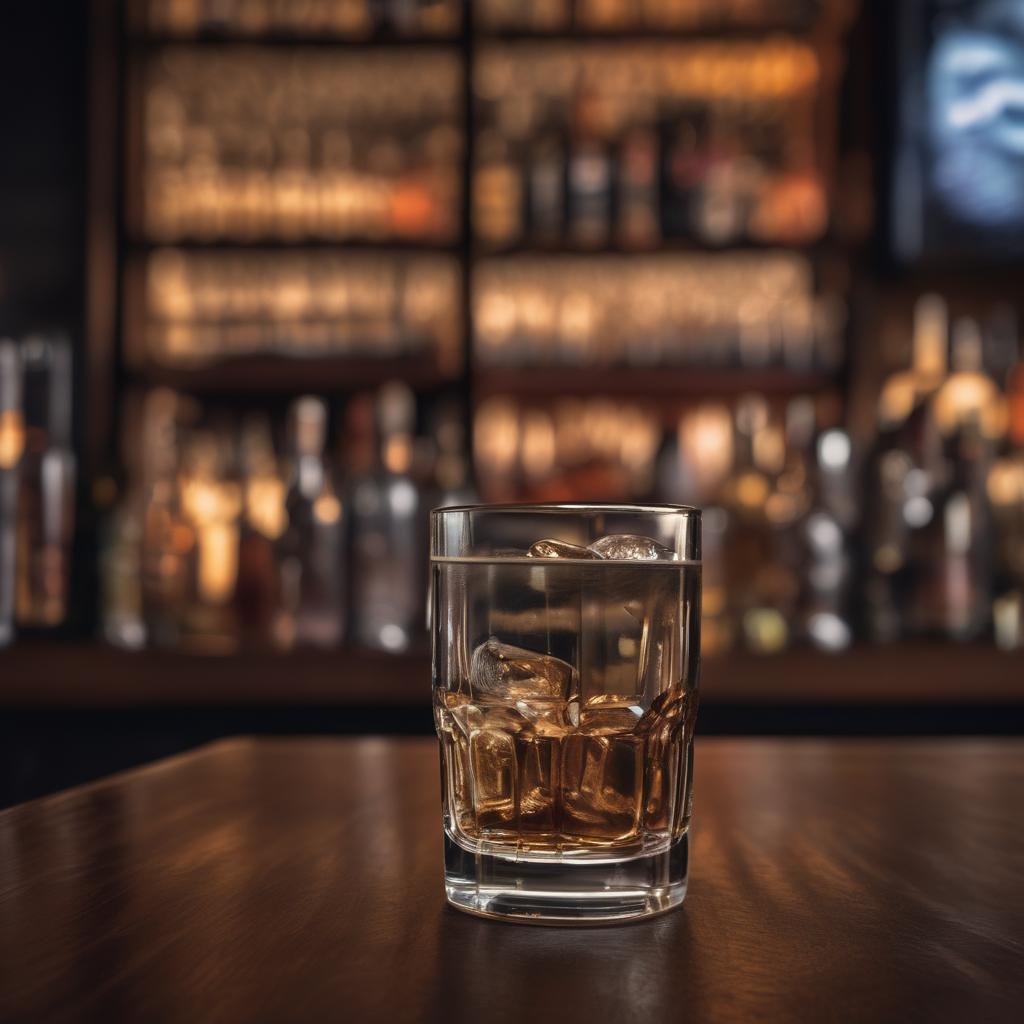} &
    \includegraphics[width=0.86\linewidth, valign=t]{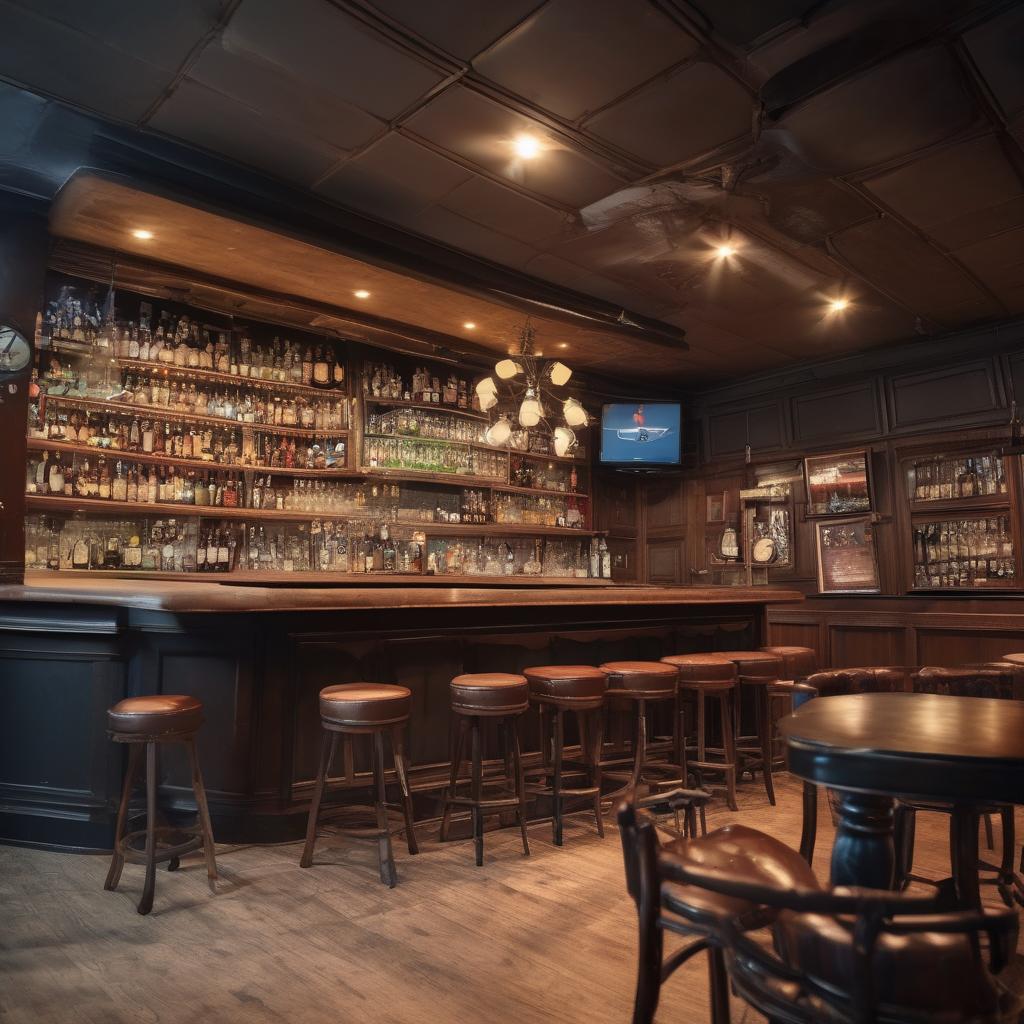} &
    \includegraphics[width=0.86\linewidth, valign=t]{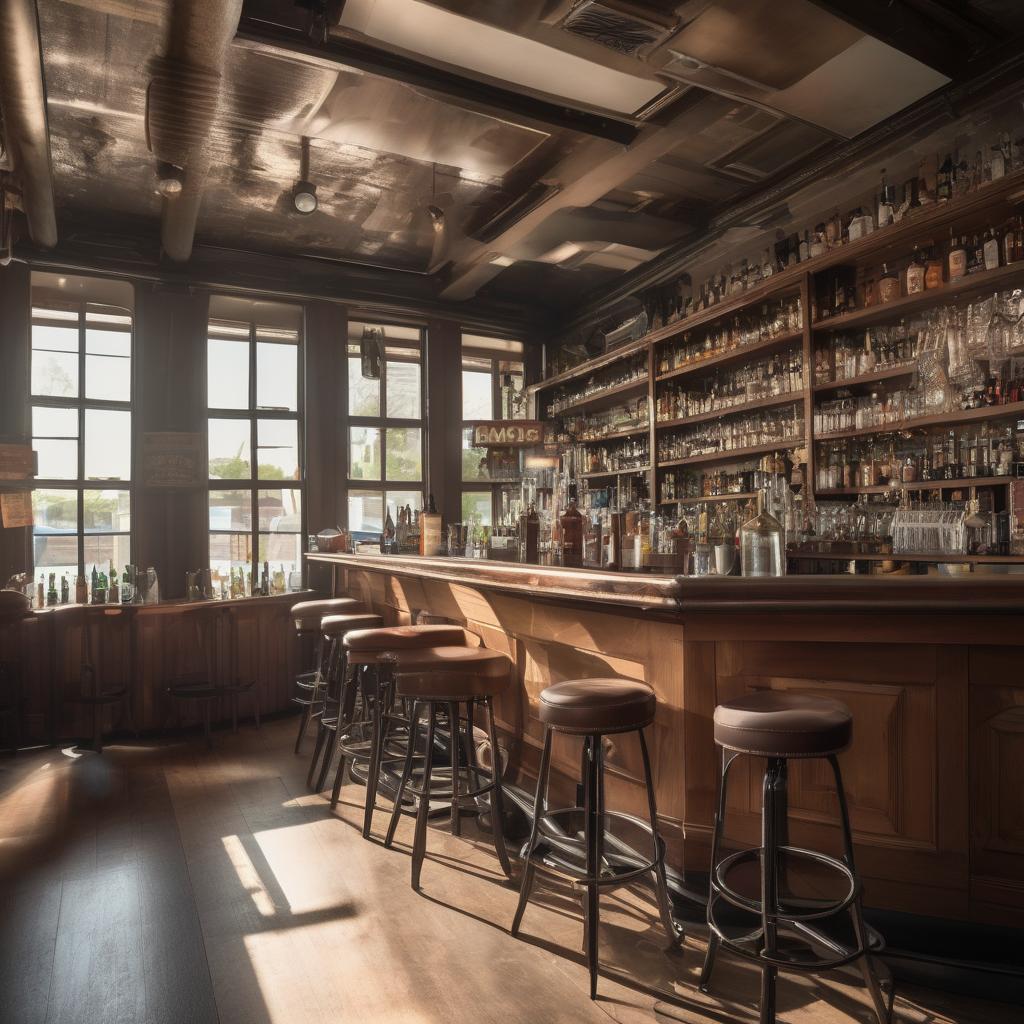} \\
    \midrule
    5 & weighted axis &
    \includegraphics[width=0.86\linewidth, valign=t]{figures/appendix/neuron_1255_new/iteration_5_weighted_axis/image_1.jpeg} &
    \includegraphics[width=0.86\linewidth, valign=t]{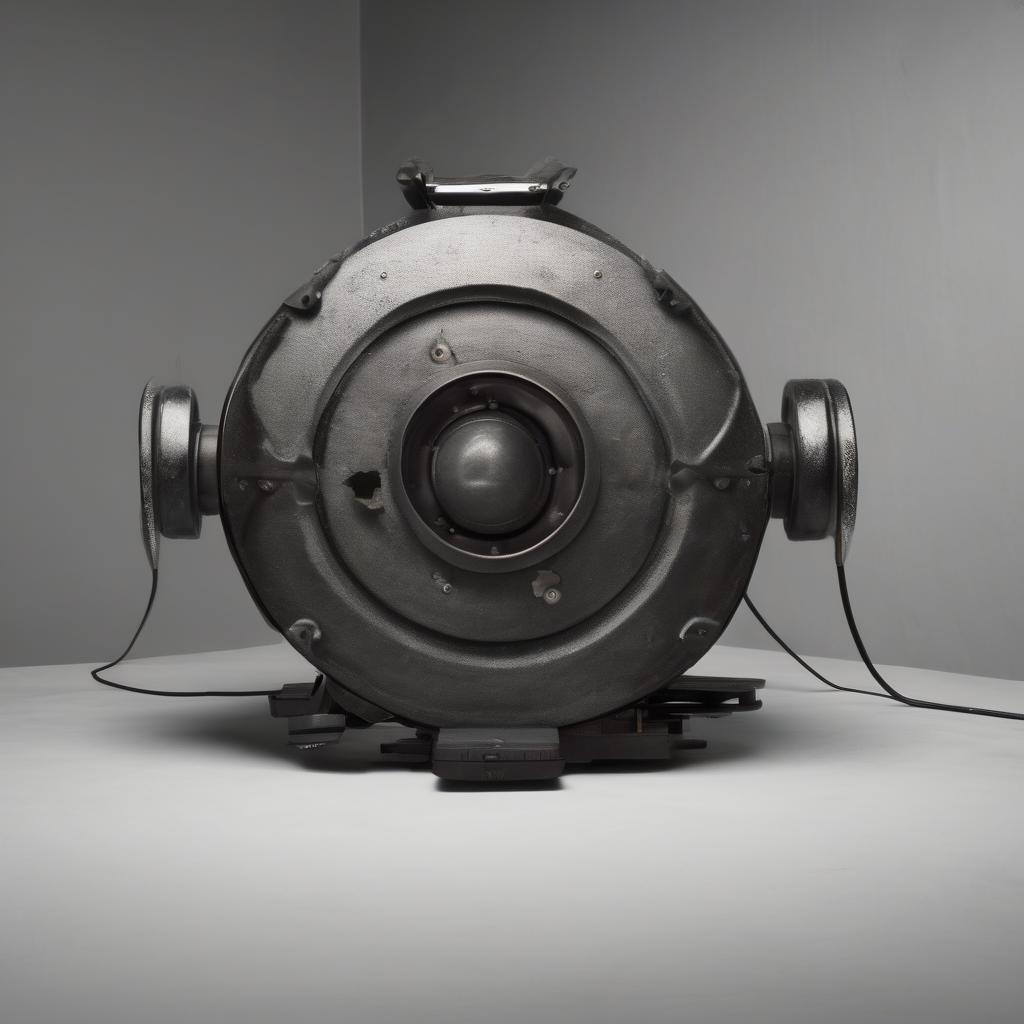} &
    \includegraphics[width=0.86\linewidth, valign=t]{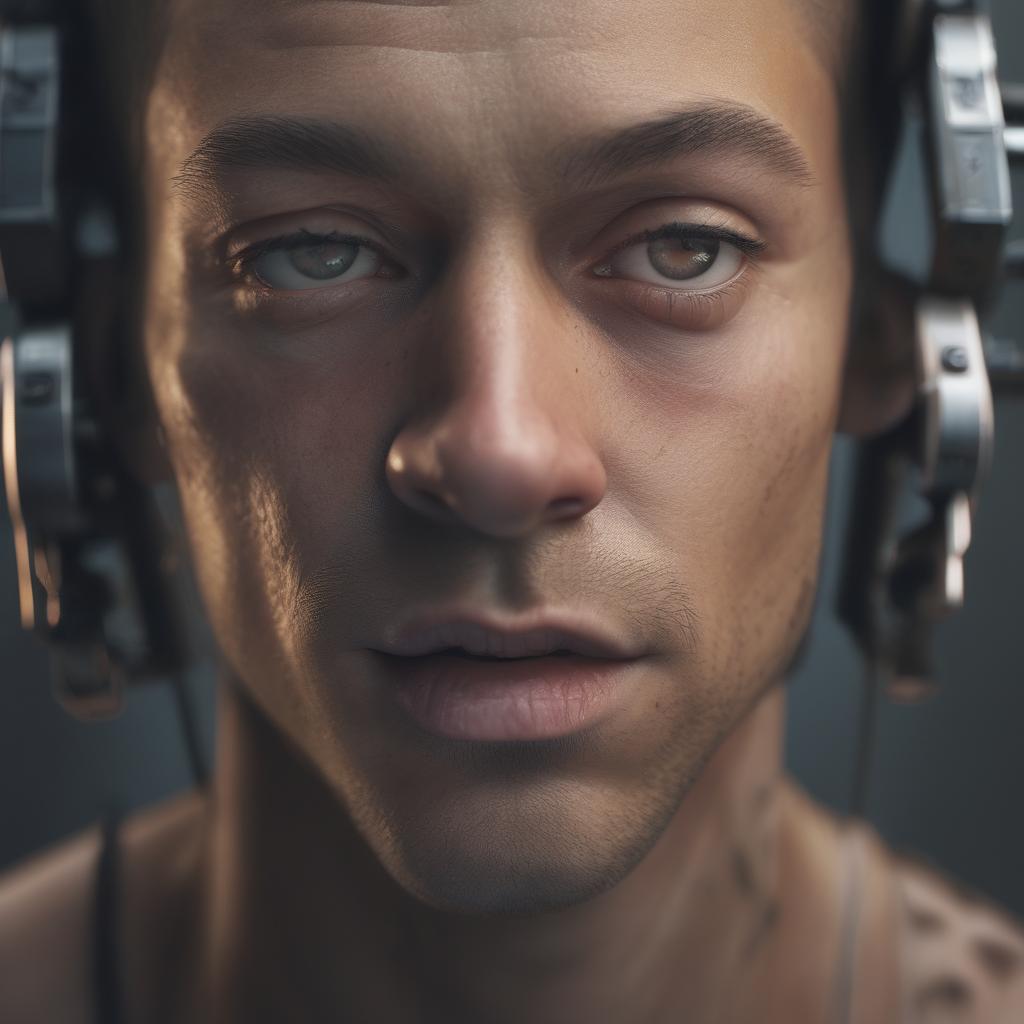} &
    \includegraphics[width=0.86\linewidth, valign=t]{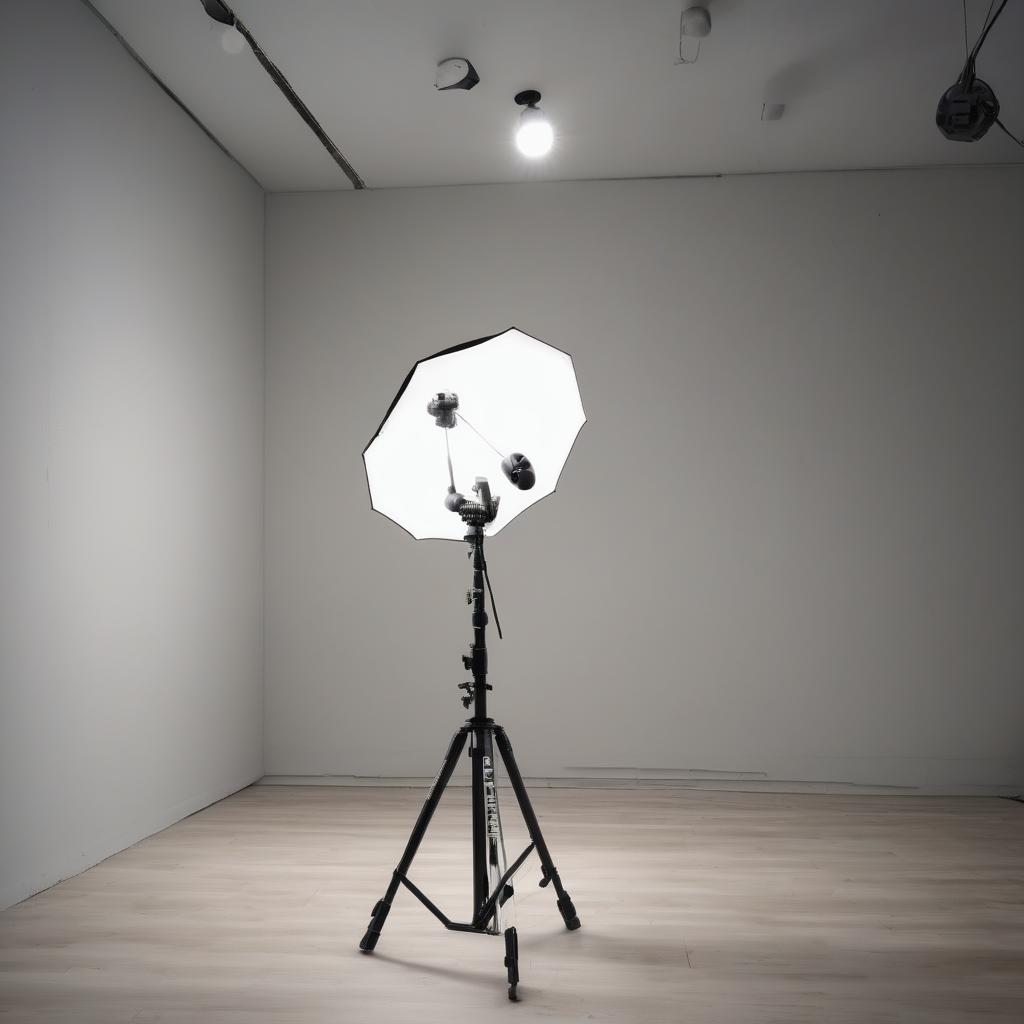} &
    \includegraphics[width=0.86\linewidth, valign=t]{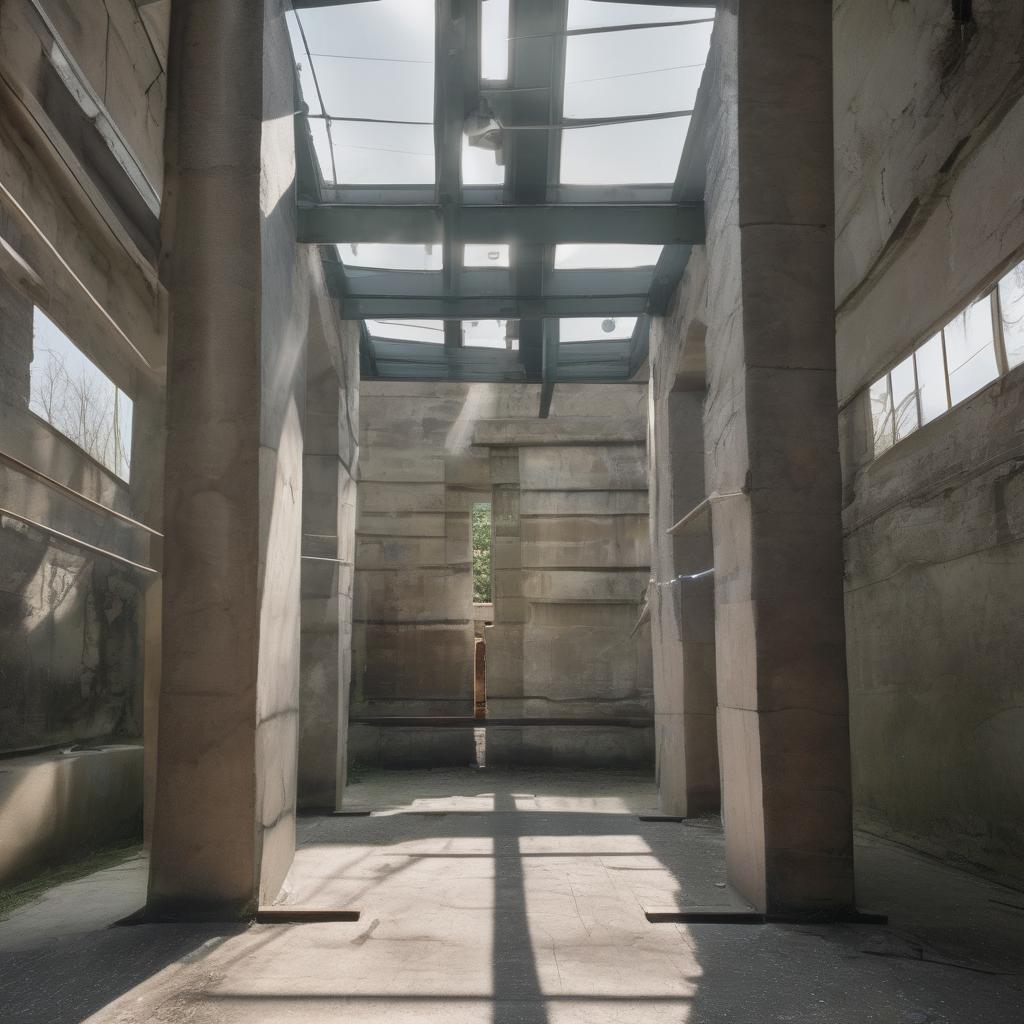} \\
    \midrule
    6 & iron &
    \includegraphics[width=0.86\linewidth, valign=t]{figures/appendix/neuron_1255_new/iteration_6_iron/image_1.jpeg} &
    \includegraphics[width=0.86\linewidth, valign=t]{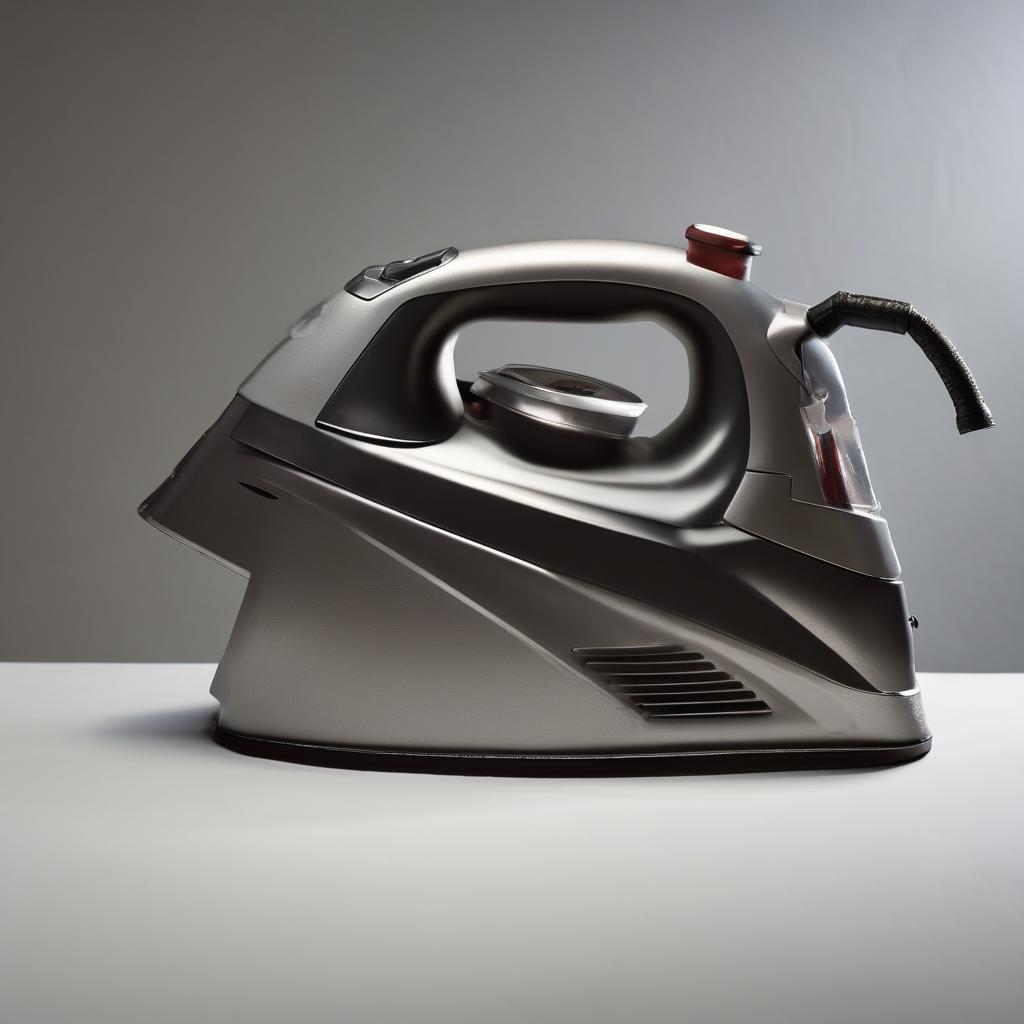} &
    \includegraphics[width=0.86\linewidth, valign=t]{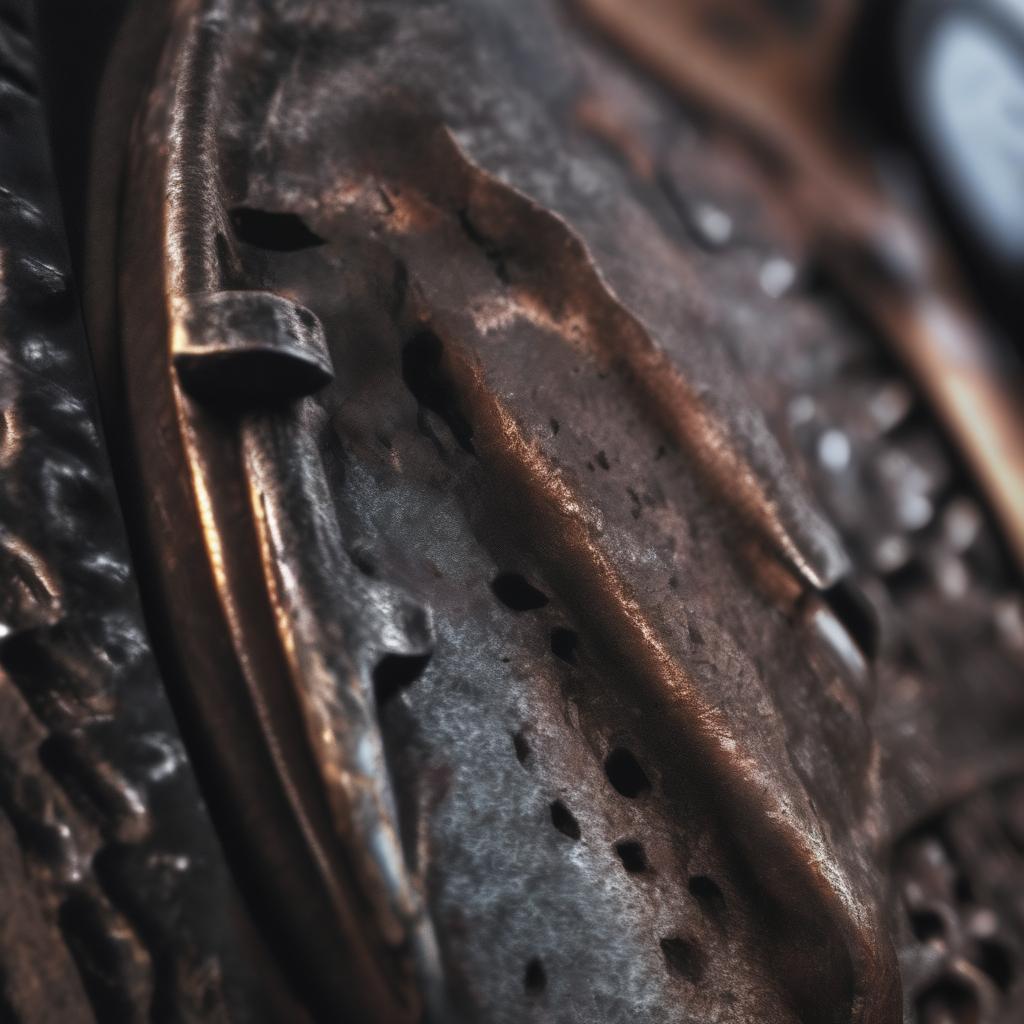} &
    \includegraphics[width=0.86\linewidth, valign=t]{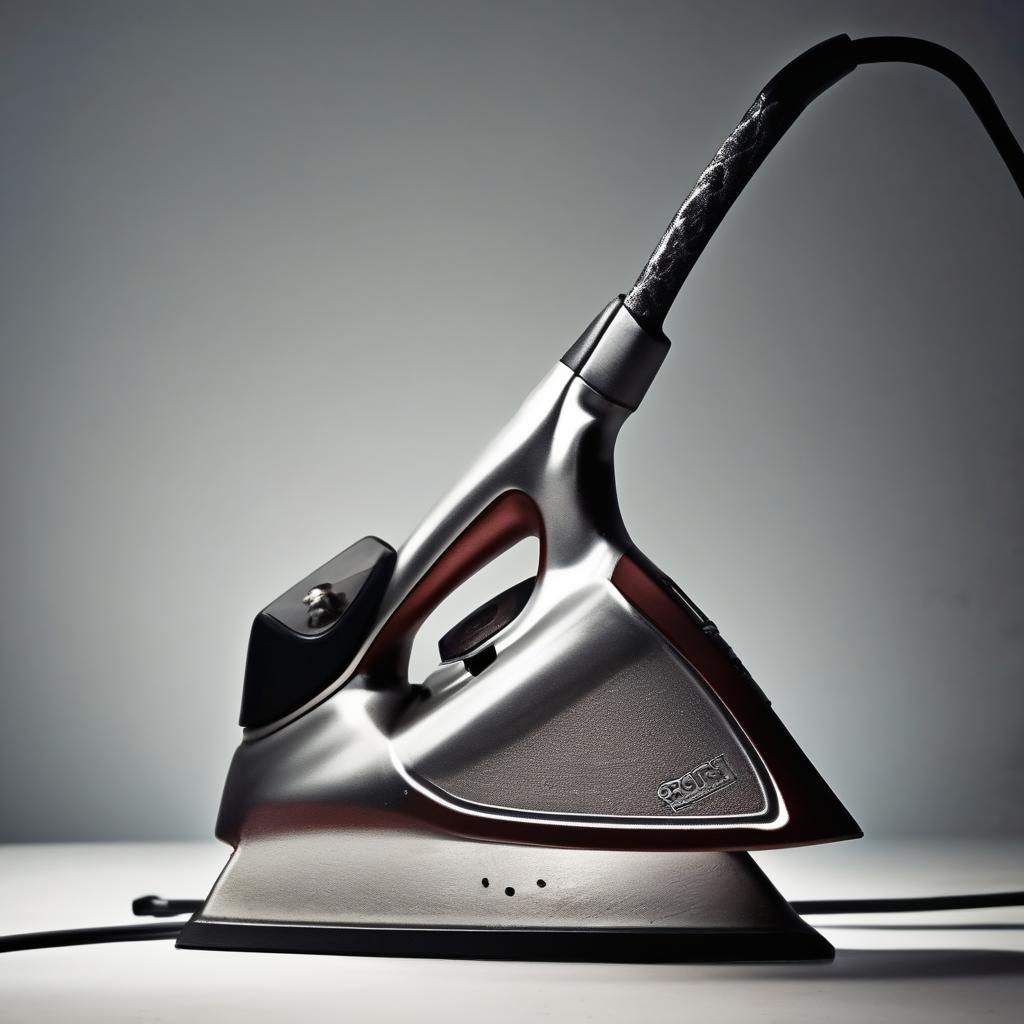} &
    \includegraphics[width=0.86\linewidth, valign=t]{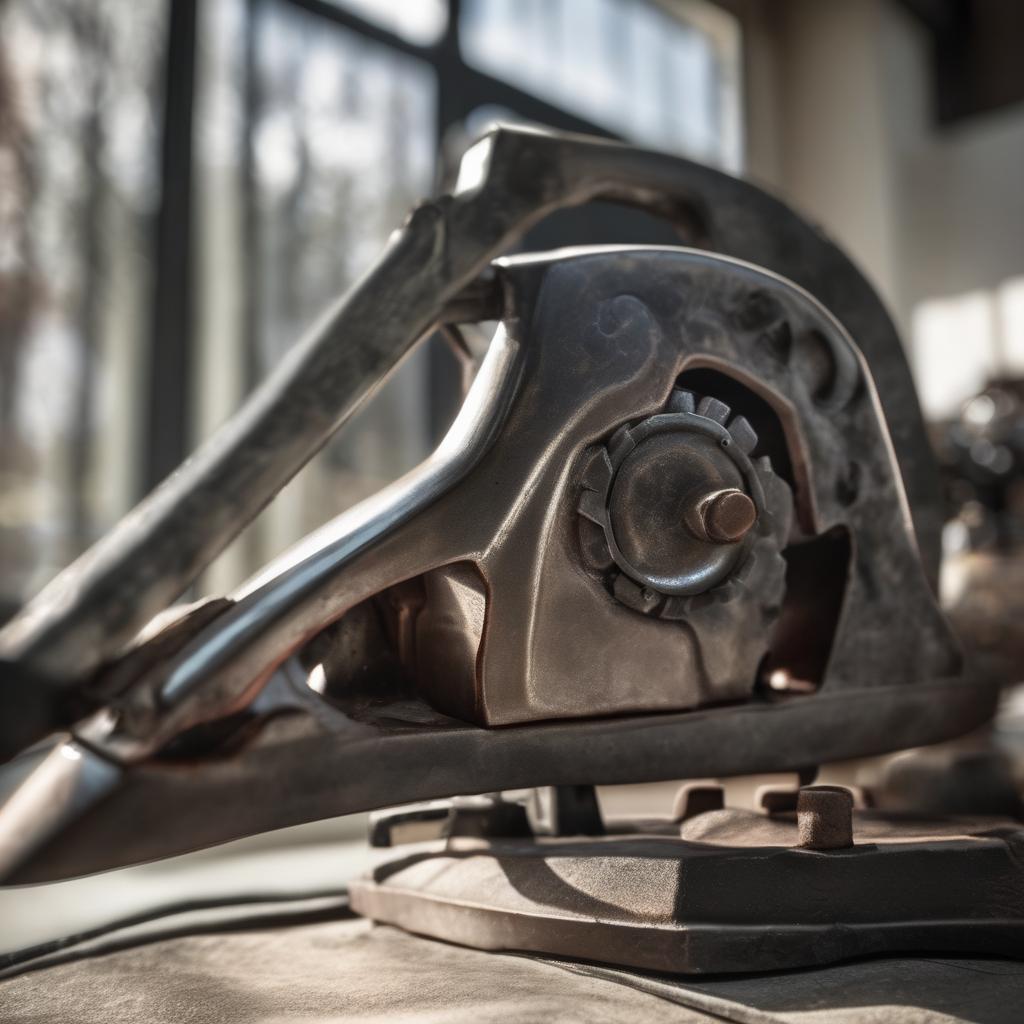} \\
    \midrule
    7 & heavy bar &
    \includegraphics[width=0.86\linewidth, valign=t]{figures/appendix/neuron_1255_new/iteration_7_heavy_bar/image_1.jpeg} &
    \includegraphics[width=0.86\linewidth, valign=t]{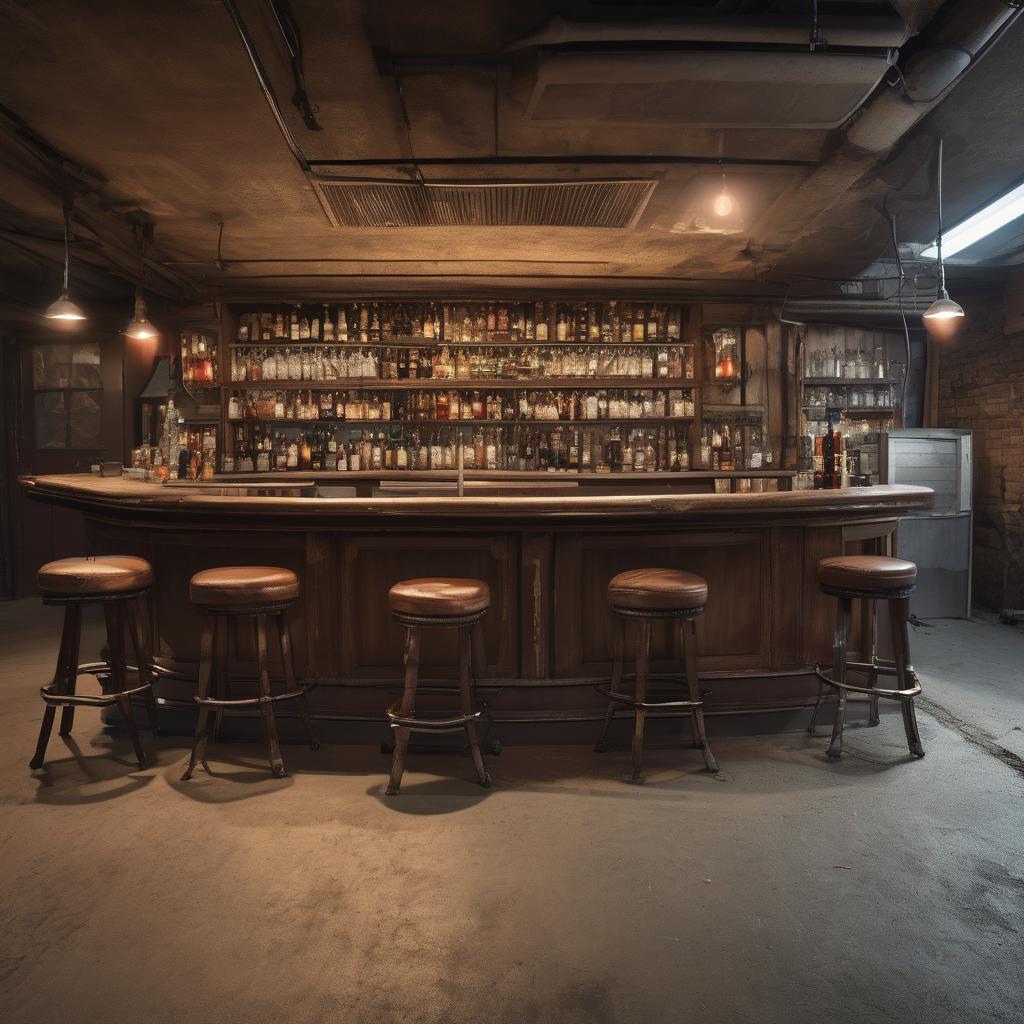} &
    \includegraphics[width=0.86\linewidth, valign=t]{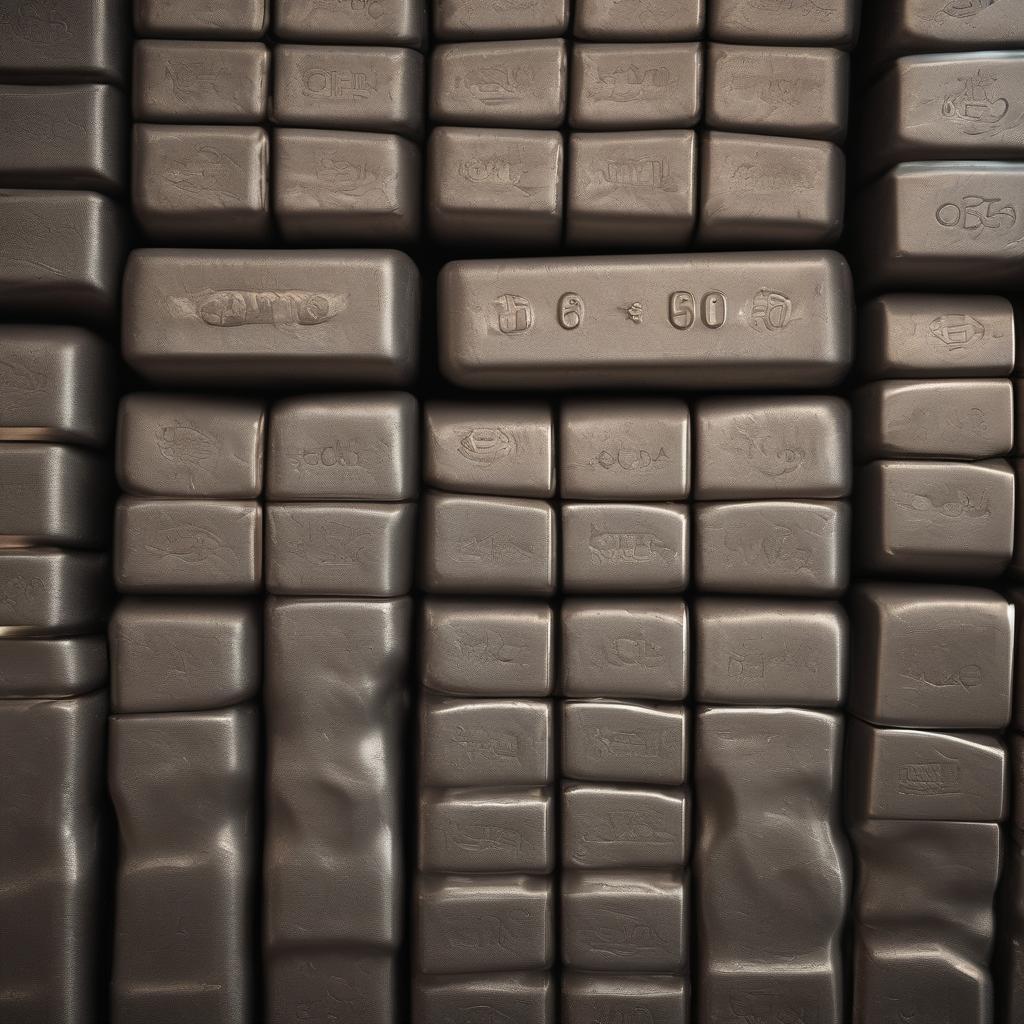} &
    \includegraphics[width=0.86\linewidth, valign=t]{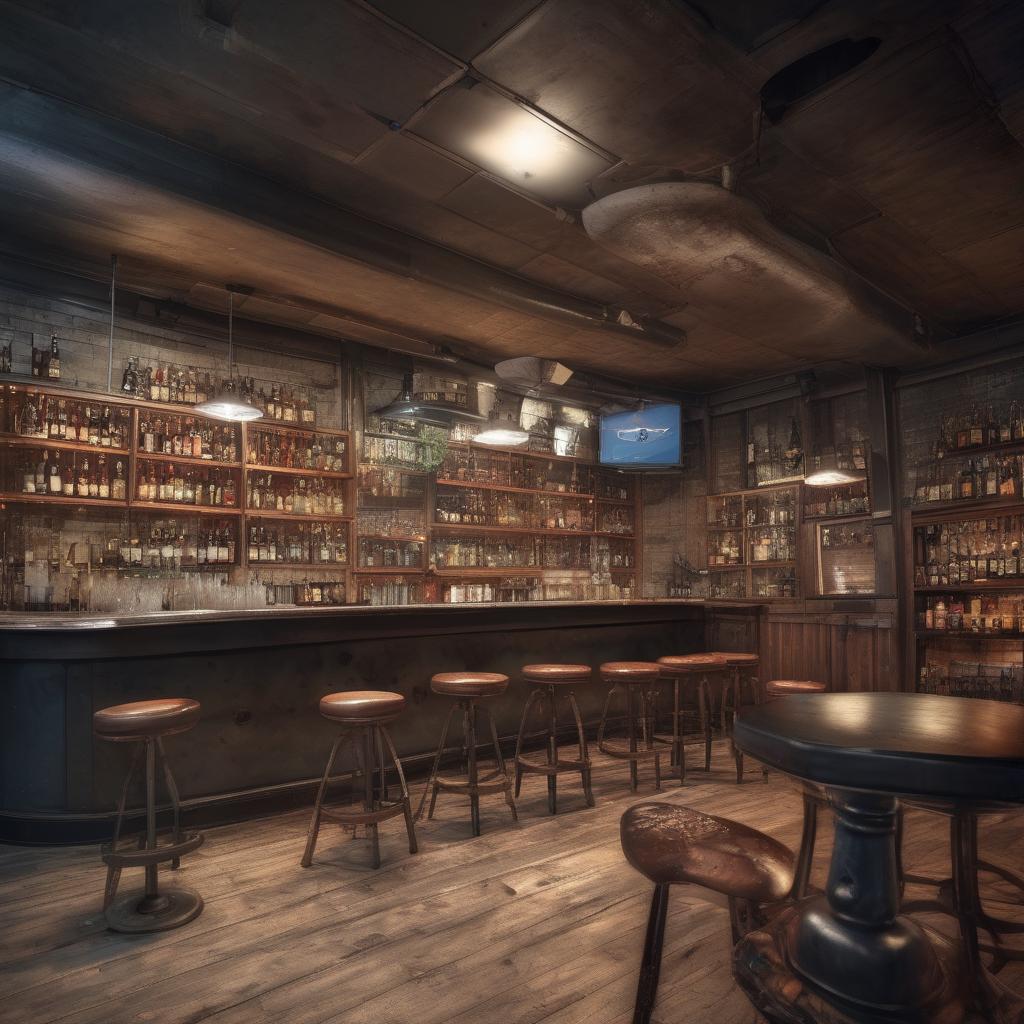} &
    \includegraphics[width=0.86\linewidth, valign=t]{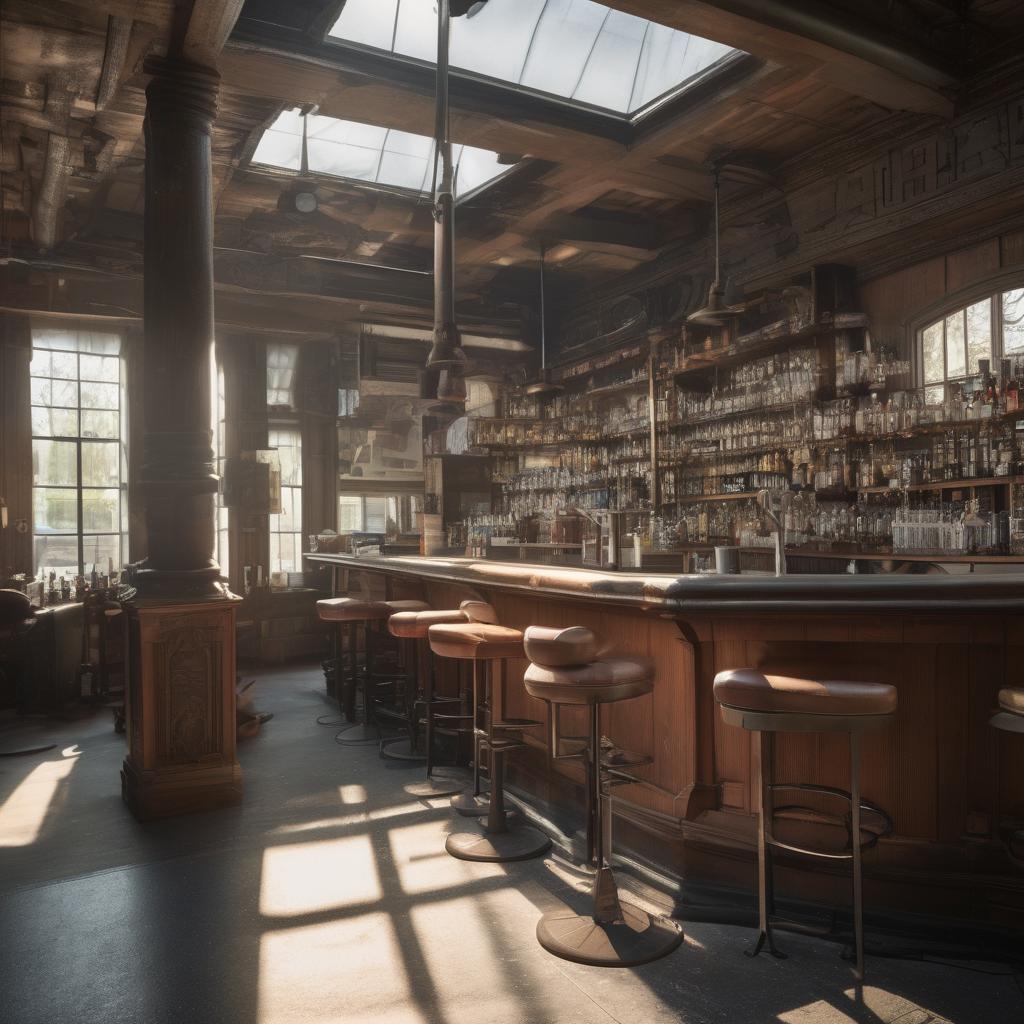} \\
    \midrule
    8 & dumbbell &
    \includegraphics[width=0.86\linewidth, valign=t]{figures/appendix/neuron_1255_new/iteration_8_dumbbell/image_1.jpeg} &
    \includegraphics[width=0.86\linewidth, valign=t]{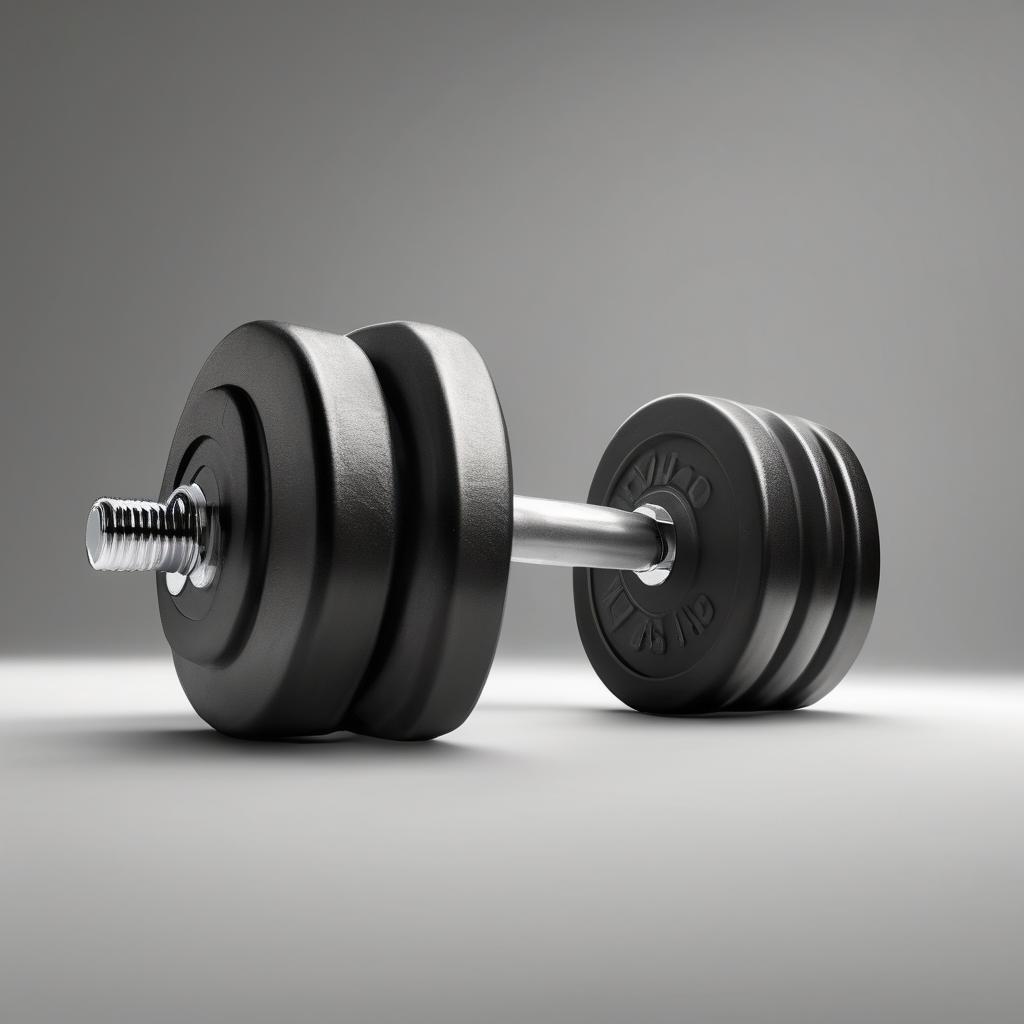} &
    \includegraphics[width=0.86\linewidth, valign=t]{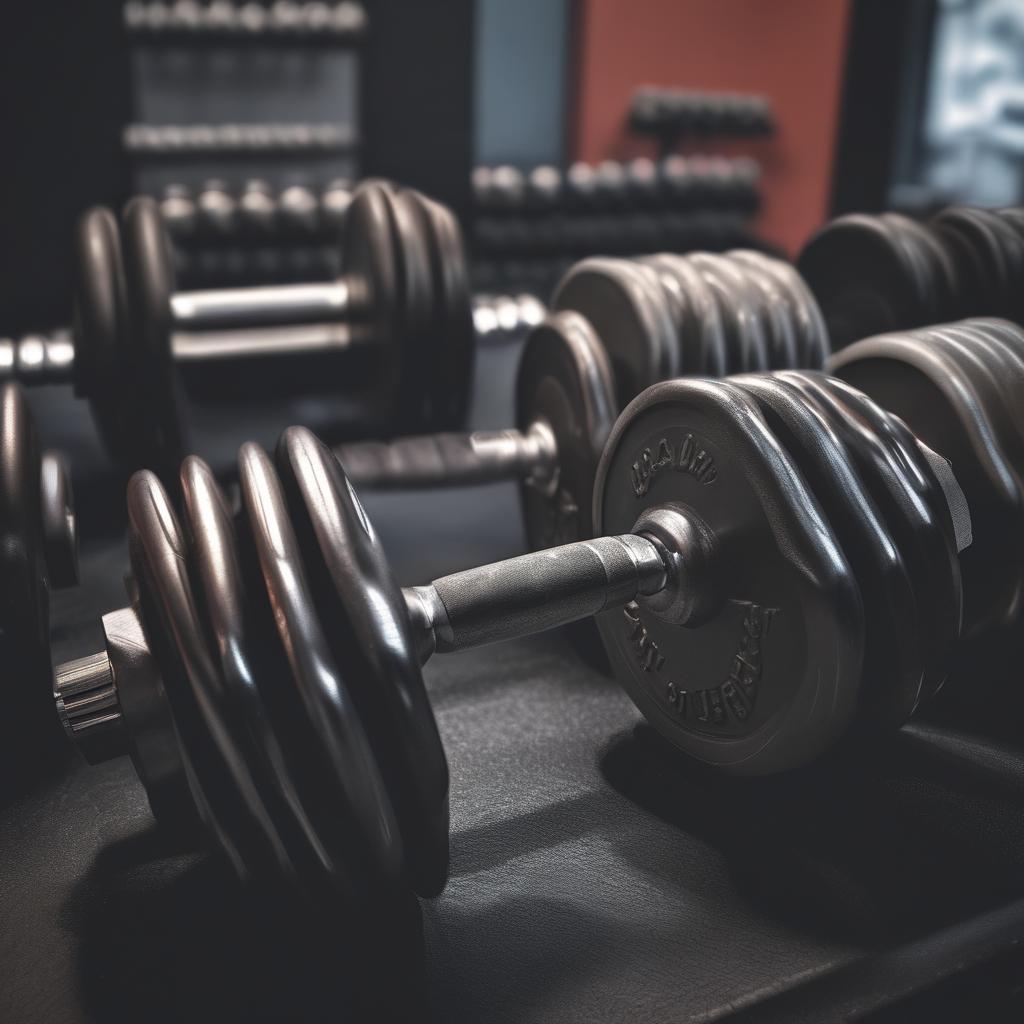} &
    \includegraphics[width=0.86\linewidth, valign=t]{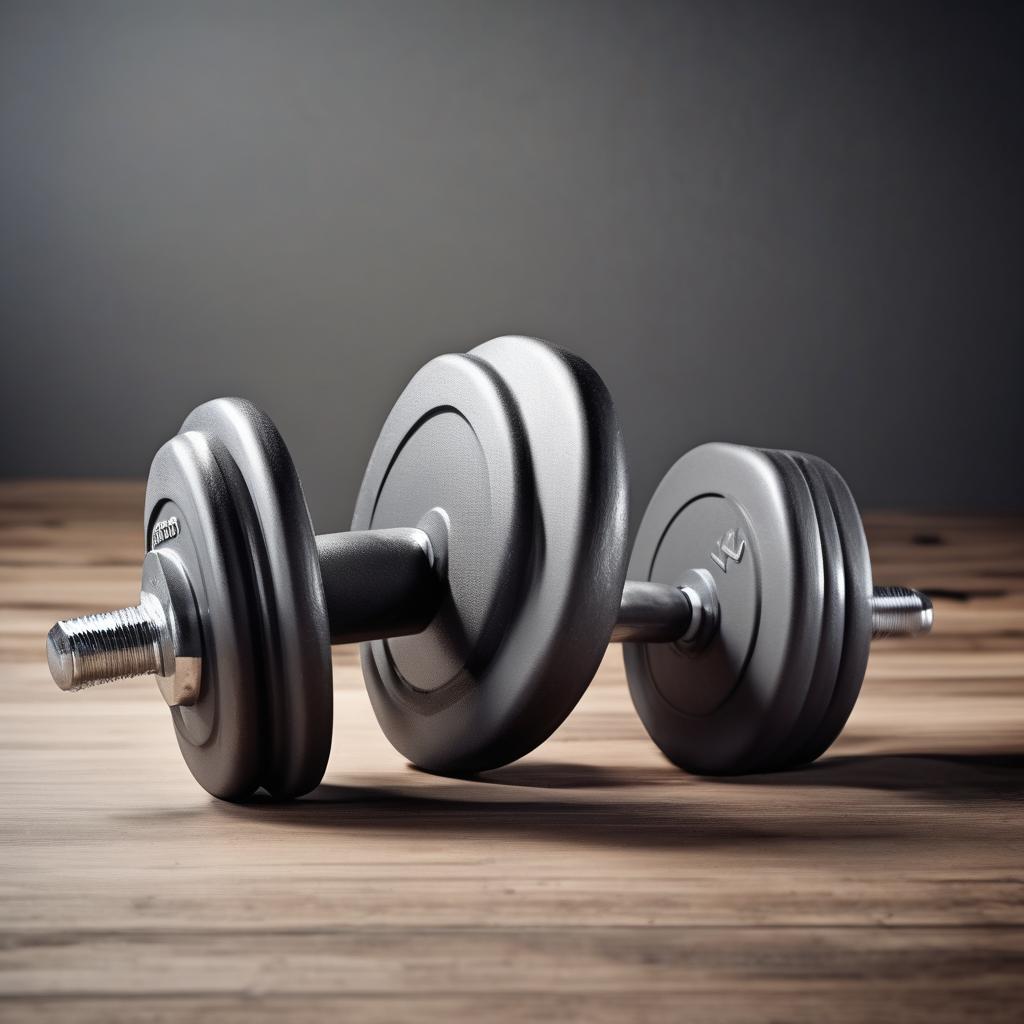} &
    \includegraphics[width=0.86\linewidth, valign=t]{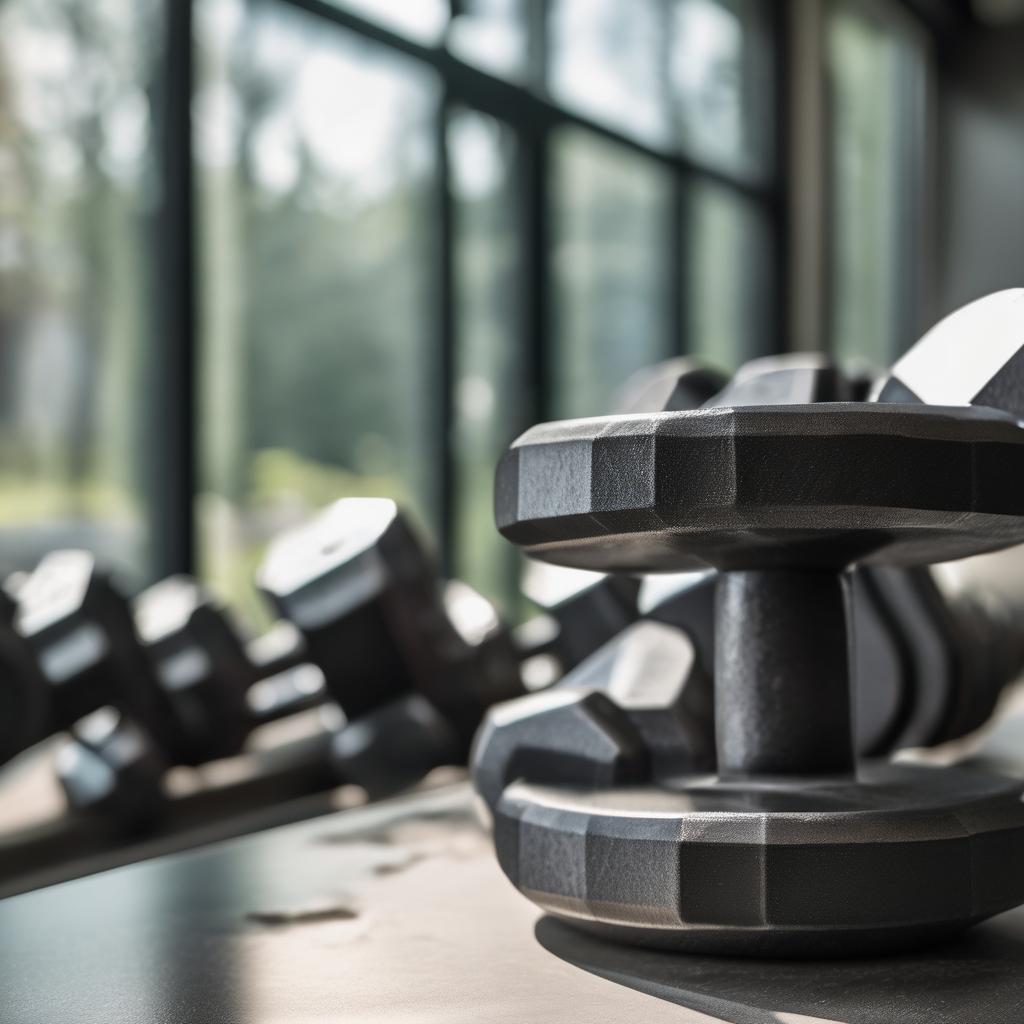} \\
    \midrule
    9 & barbell &
    \includegraphics[width=0.86\linewidth, valign=t]{figures/appendix/neuron_1255_new/iteration_9_barbell/image_1.jpeg} &
    \includegraphics[width=0.86\linewidth, valign=t]{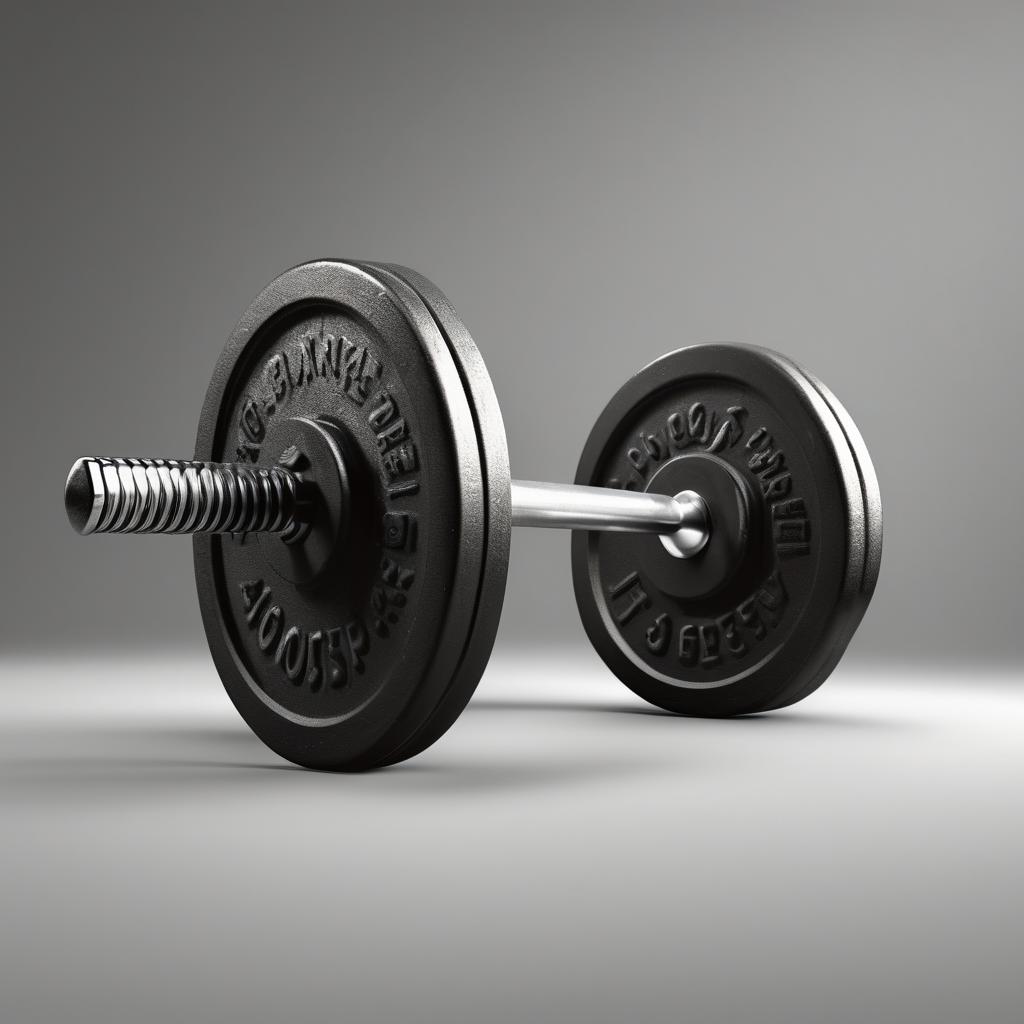} &
    \includegraphics[width=0.86\linewidth, valign=t]{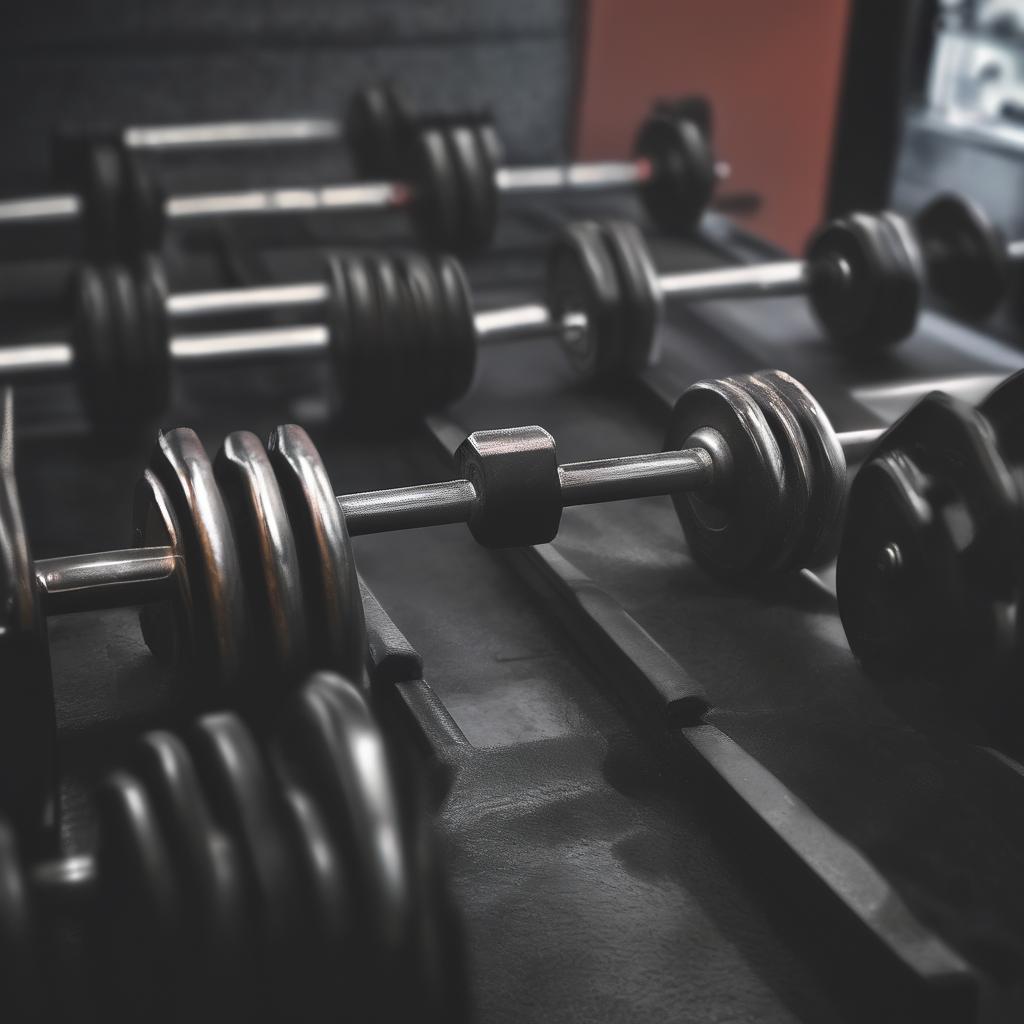} &
    \includegraphics[width=0.86\linewidth, valign=t]{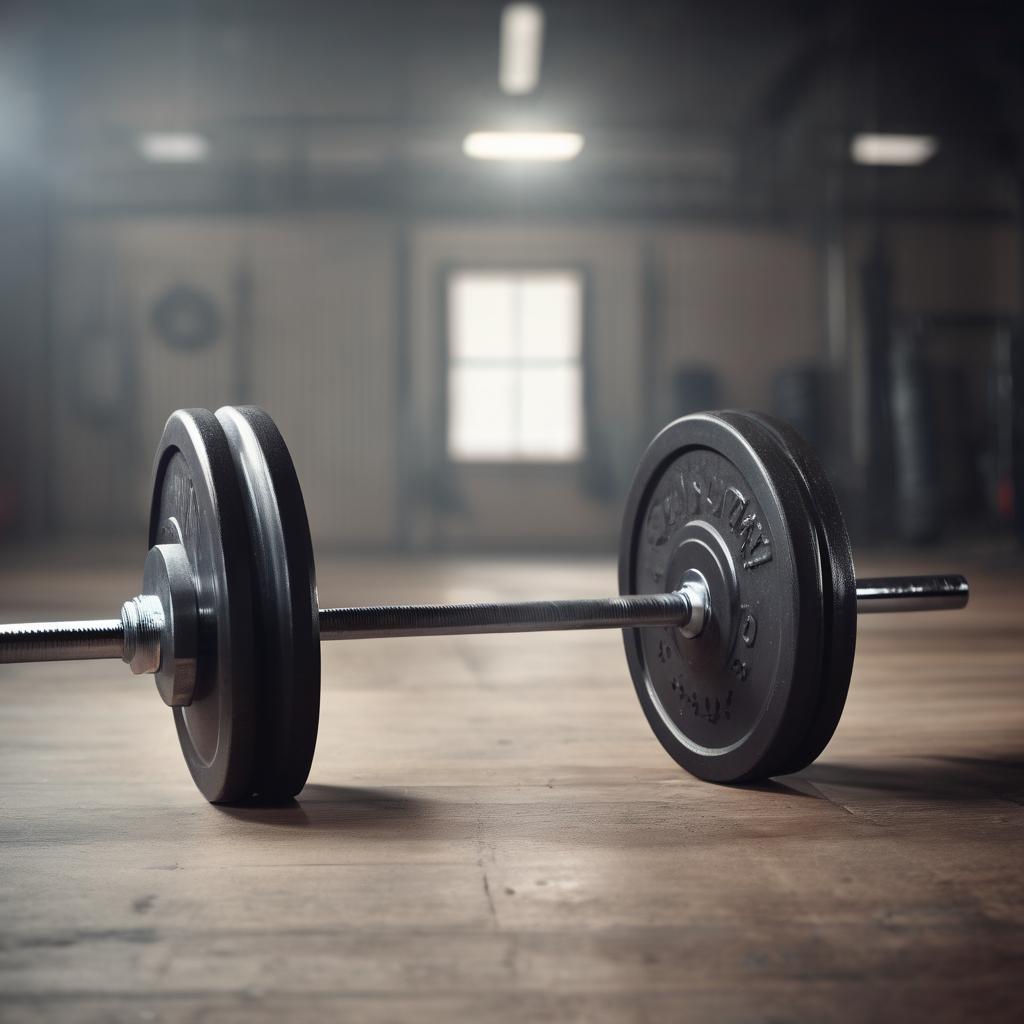} &
    \includegraphics[width=0.86\linewidth, valign=t]{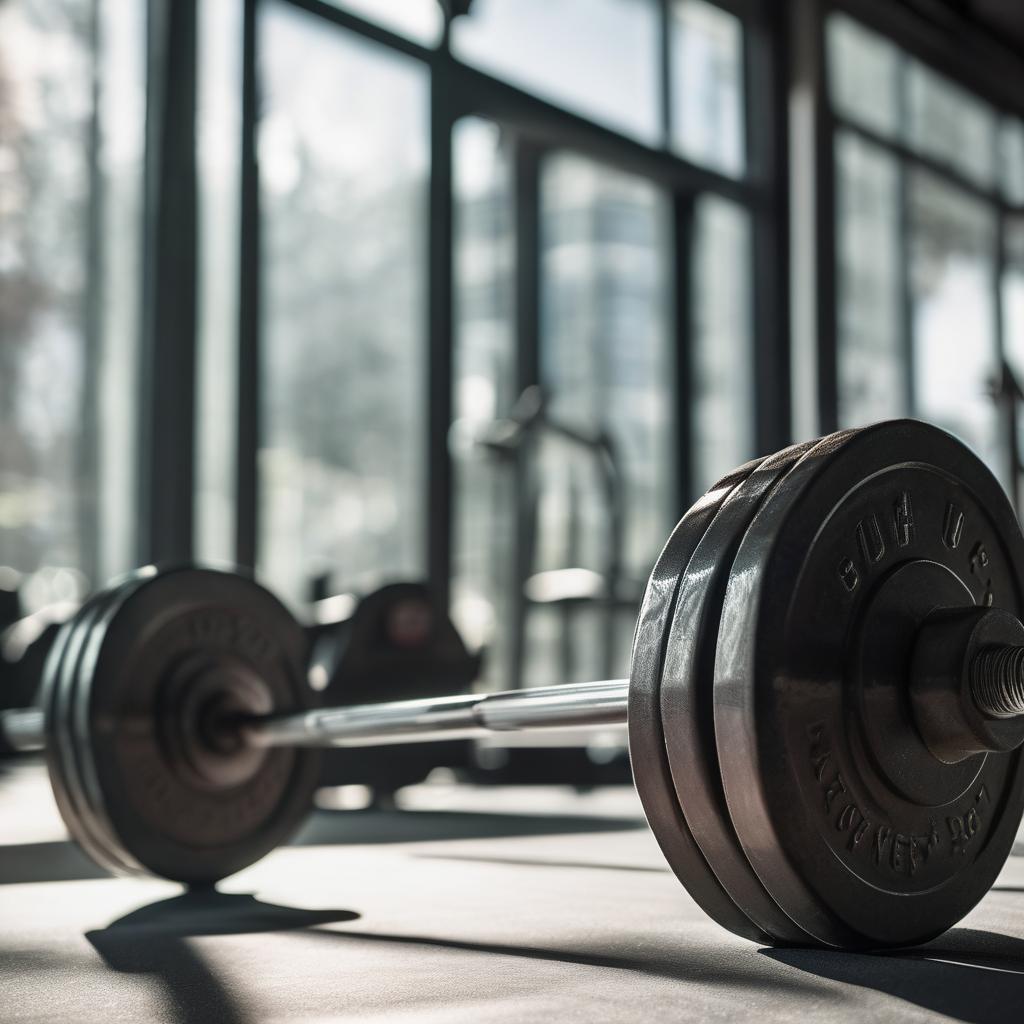} \\
    \midrule
    10 & weight bar &
    \includegraphics[width=0.86\linewidth, valign=t]{figures/appendix/neuron_1255_new/iteration_10_weight_bar/image_1.jpeg} &
    \includegraphics[width=0.86\linewidth, valign=t]{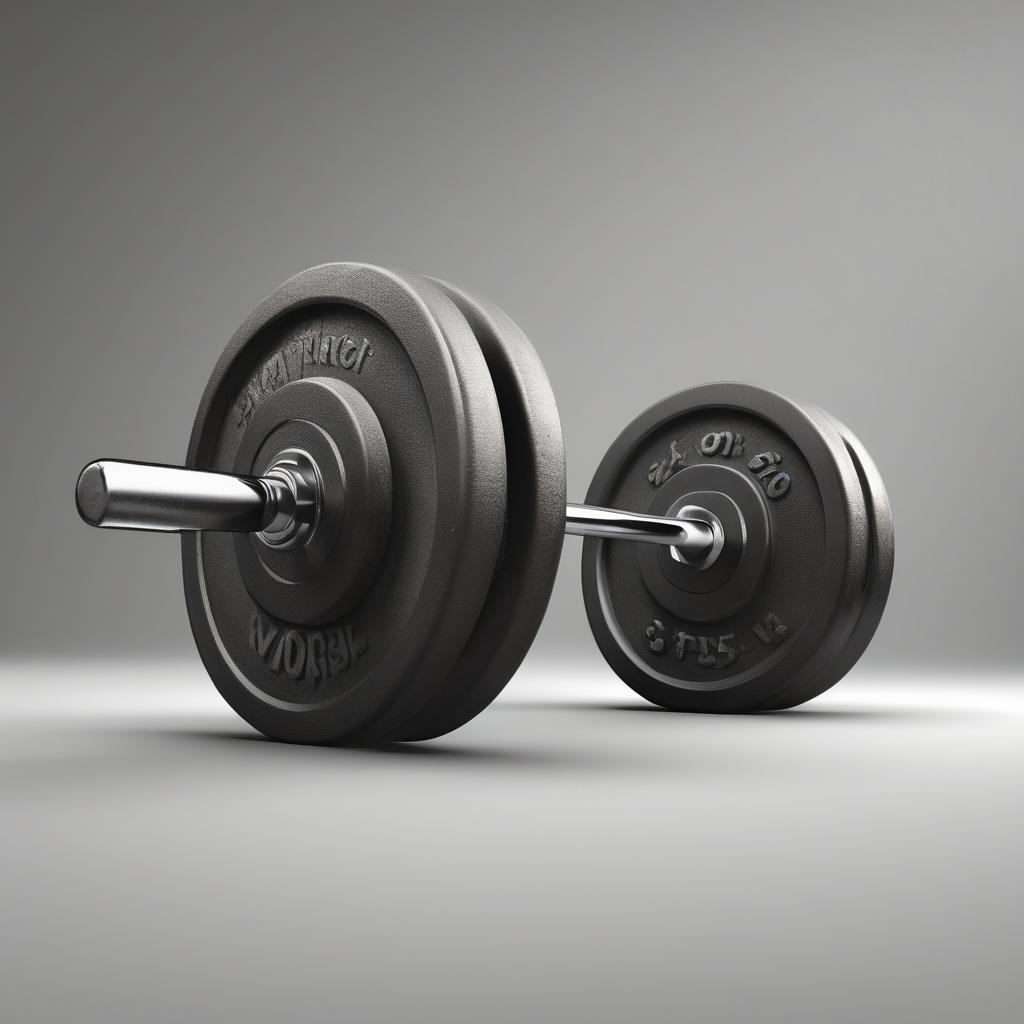} &
    \includegraphics[width=0.86\linewidth, valign=t]{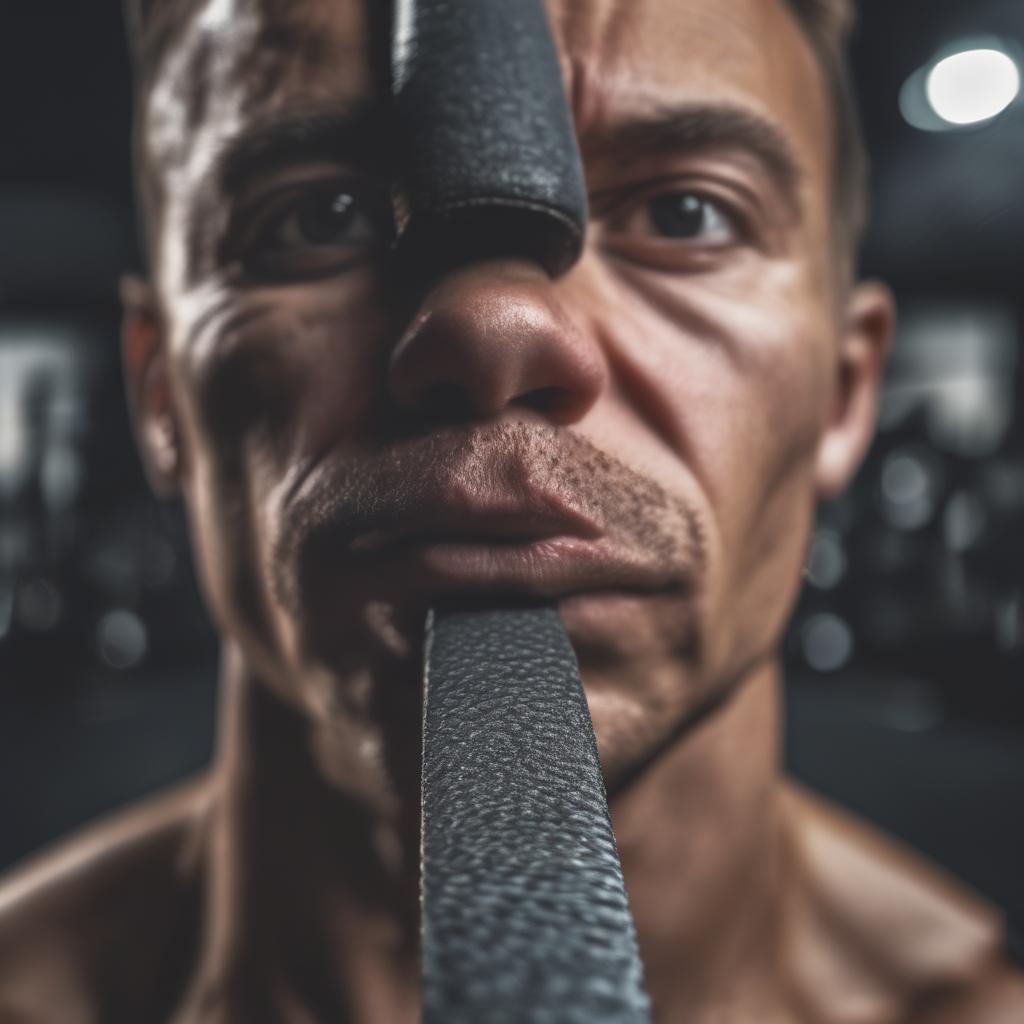} &
    \includegraphics[width=0.86\linewidth, valign=t]{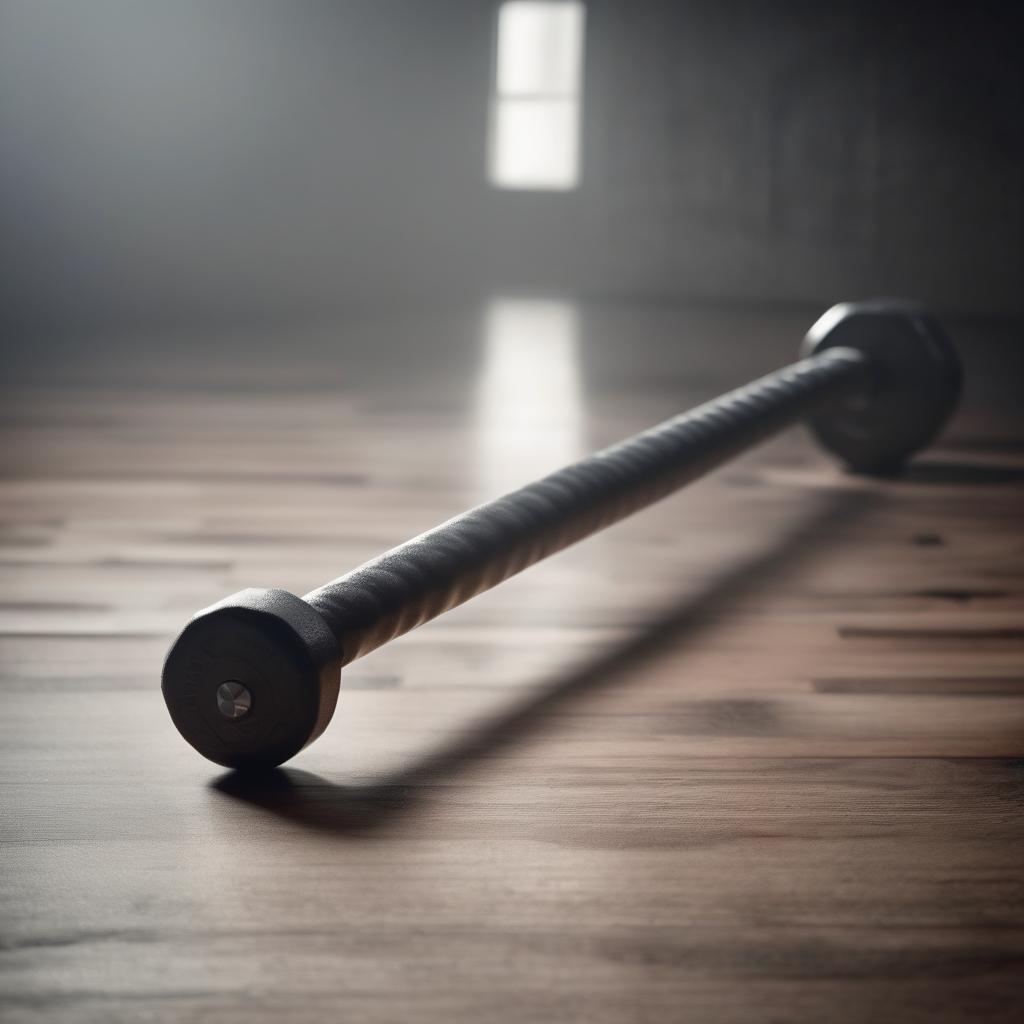} &
    \includegraphics[width=0.86\linewidth, valign=t]{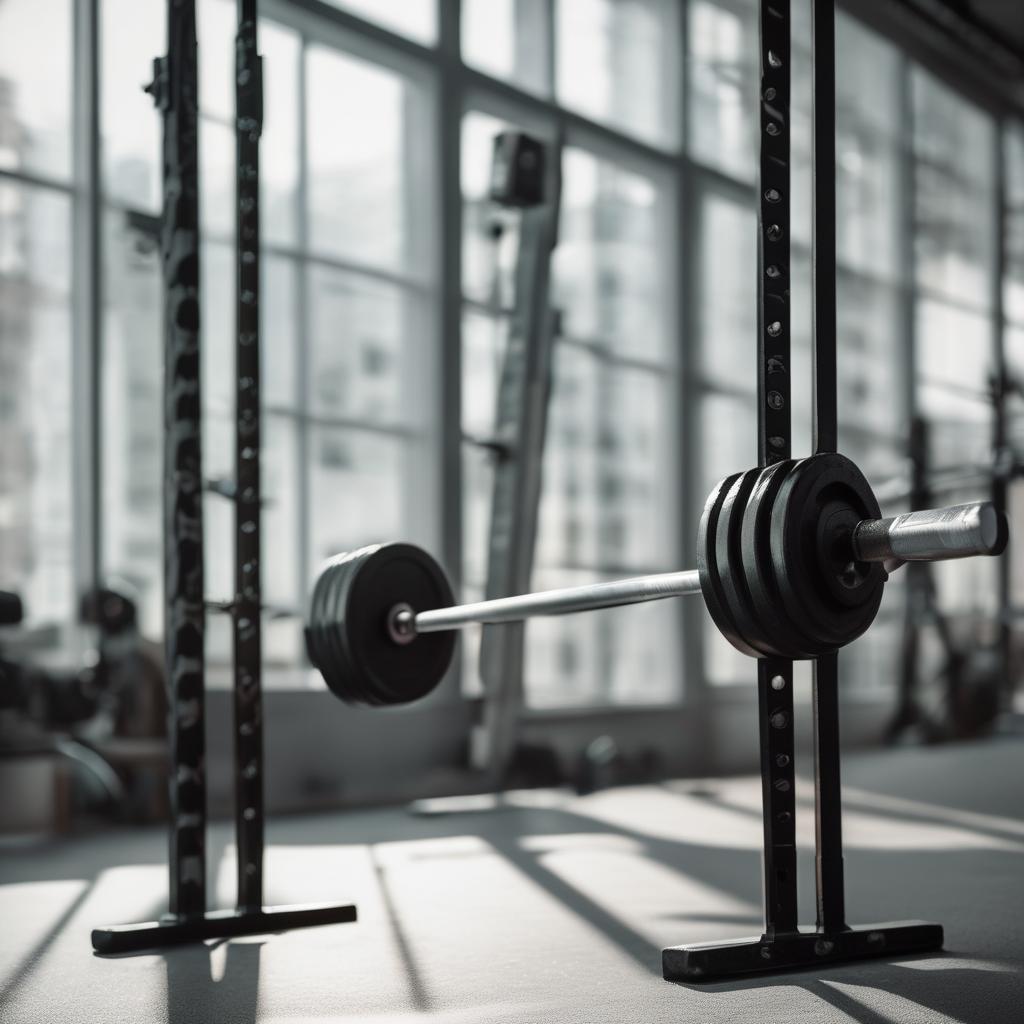} \\
    \midrule
    S & table-based exercise &
    \includegraphics[width=0.86\linewidth, valign=t]{figures/appendix/neuron_1255_new/iteration_summary_table-based_exercise/image_1.jpeg} &
    \includegraphics[width=0.86\linewidth, valign=t]{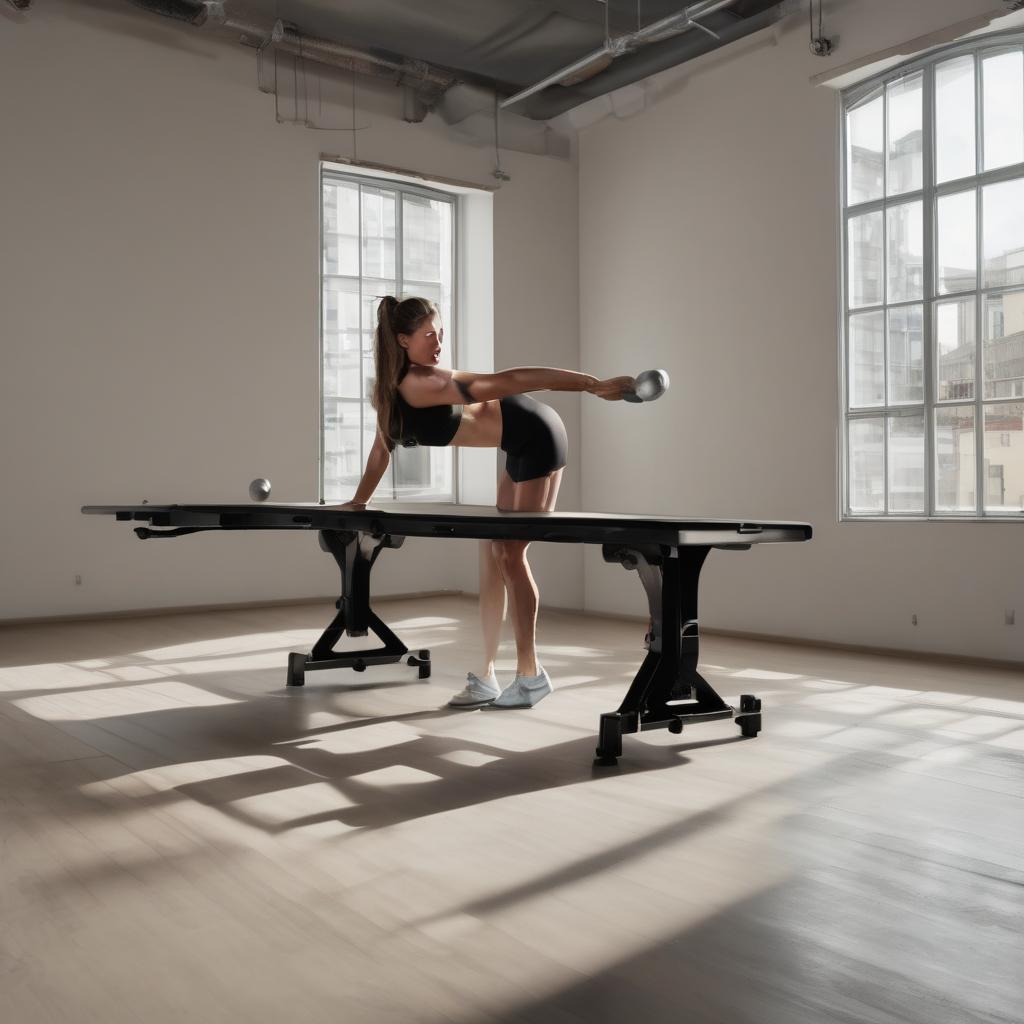} &
    \includegraphics[width=0.86\linewidth, valign=t]{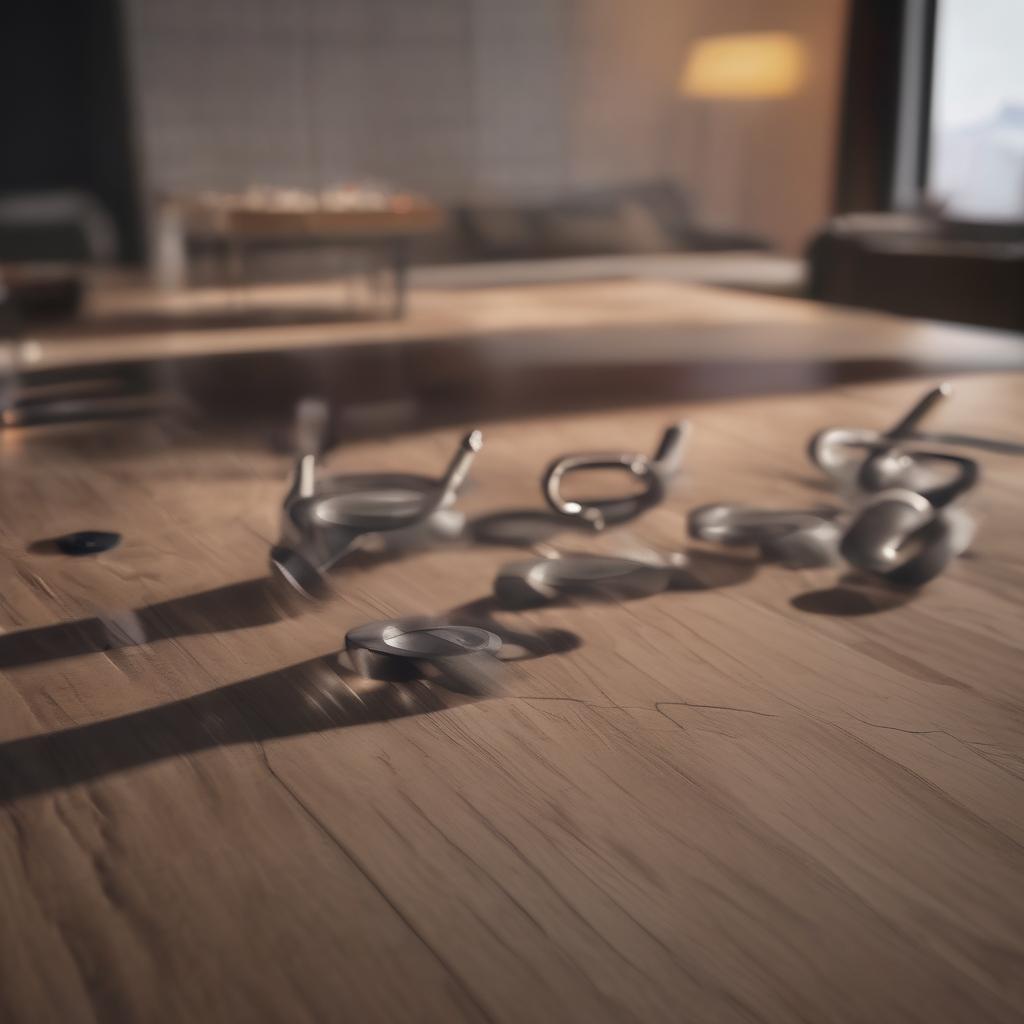} &
    \includegraphics[width=0.86\linewidth, valign=t]{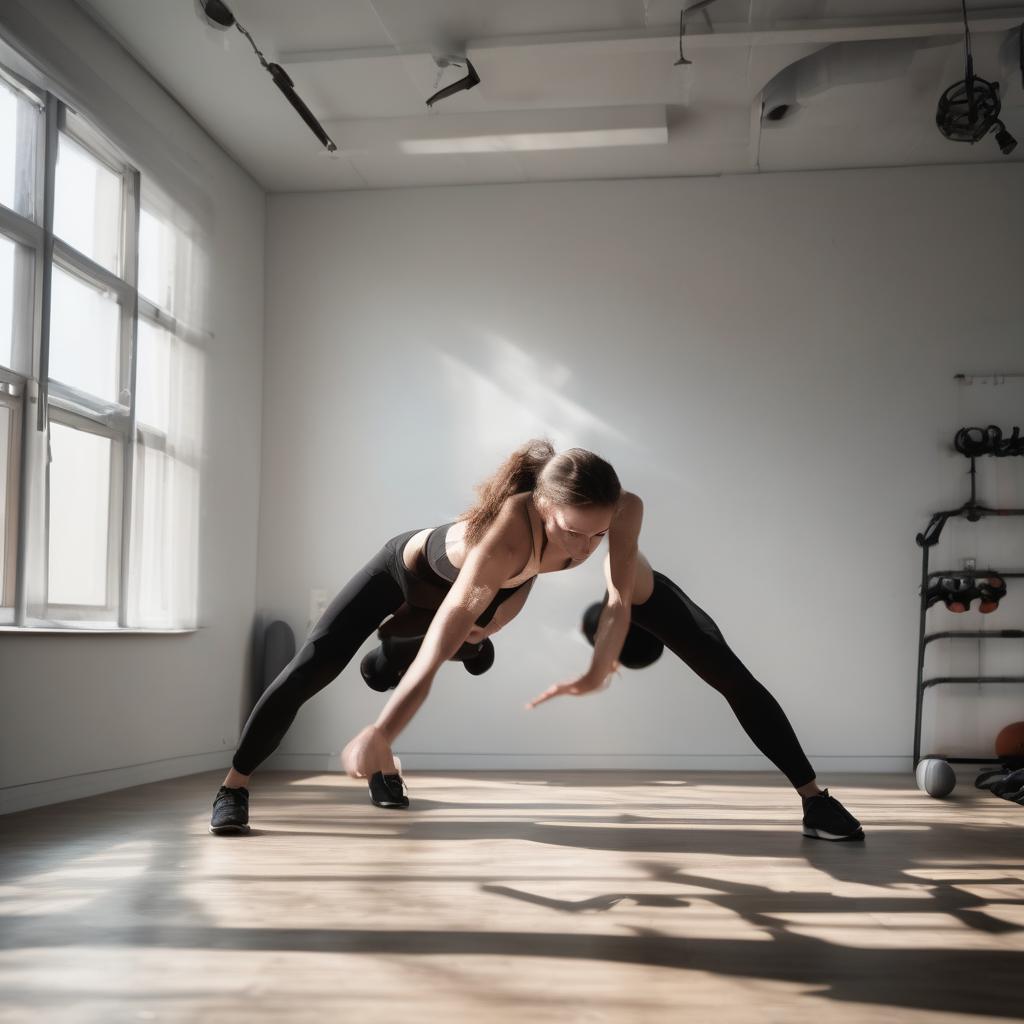} &
    \includegraphics[width=0.86\linewidth, valign=t]{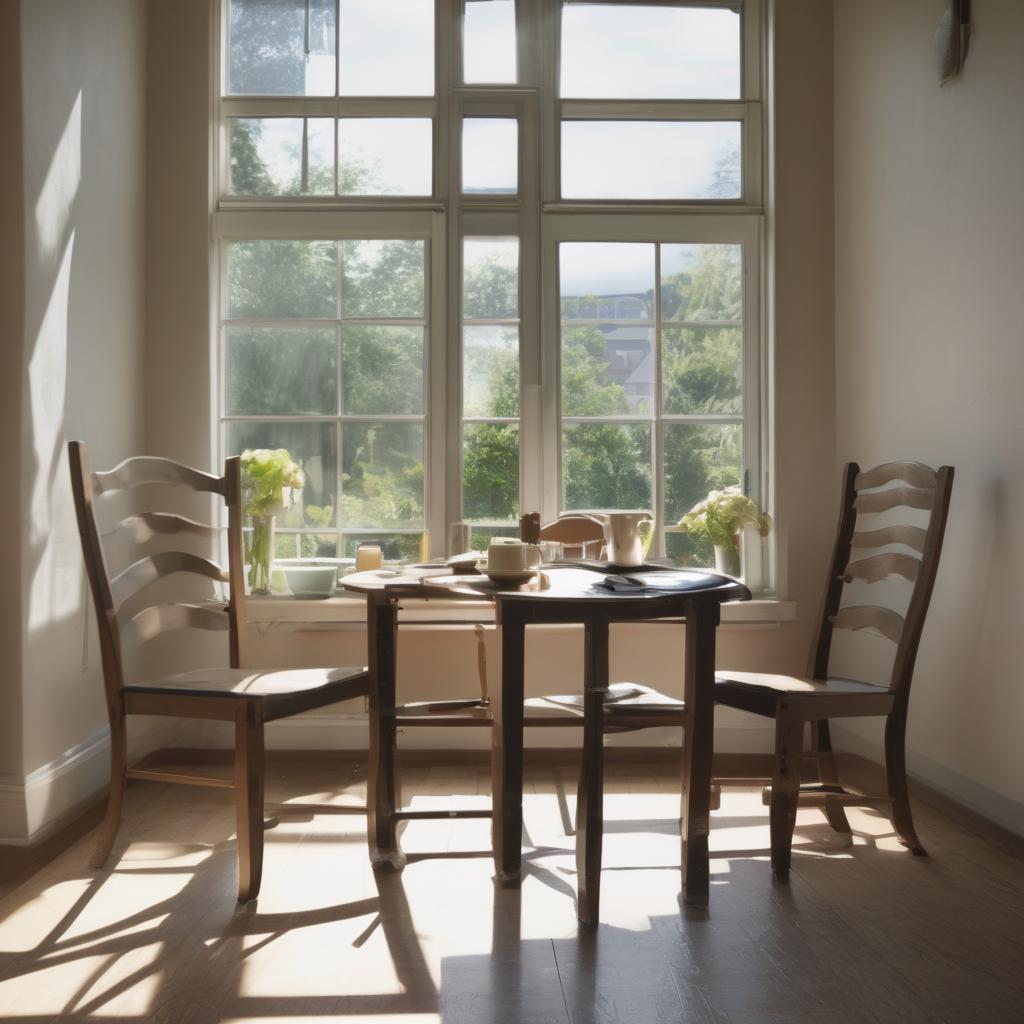} \\
    \bottomrule
    \end{tabularx}
\end{table}

\clearpage
\section{Statistical and Causal Validation}
\subsection{Statistical Validation}
\label{app:statistical}

To substantiate the claimed superiority of \LINE{} on the CoSy scores, we conduct a rigorous statistical significance analysis that also ensures consistent performance between FLUX.1[dev] and SDXL T2I models. All tests use a significance level of $\alpha = 0.05$. We apply the Wilcoxon signed-rank test, paired by neuron, to assess whether \LINE{} achieves significantly higher AUC and MAD scores than each baseline (\Cref{tab:stat_sign}). The test is applied separately within each dataset/model/layer combination. Results confirm that \LINE{} significantly outperforms all baselines.

\begin{table}[h]
    \centering
    \caption{\textbf{Statistical significance test results comparing \LINE{} and prior methods.} Expanding upon the CoSy benchmark results from \Cref{tab:comparison_table}, we calculate the statistical significance of the differences between \LINE{} and the baselines (CLIP-Dissect, INVERT, and DnD). The $p$-values confirm that \LINE{} achieves statistically significant improvements in almost all configurations, with the sole exception being the comparison against CLIP-Dissect on the ResNet18 Places365 dataset. \underline{Underlined} values indicate a lack of statistical significance ($p > 0.05$).}
    \label{tab:stat_sign}
    \vspace{0.5em}
    \begin{tabular}{l l l c c c}
    \toprule
    \textbf{Model} & \textbf{Dataset} & \textbf{Method of reference} & \textbf{AUC} & \textbf{MAD} \\
    \midrule
    \multirow{3}{*}{ResNet18} & \multirow{3}{*}{ImageNet}
    & CLIP-Dissect & $<10^{-4}$ & $<10^{-4}$ \\
    & & INVERT & $<10^{-4}$ & $<10^{-4}$ \\
    & & DnD & $<10^{-4}$ & $<10^{-4}$ \\
    \midrule
    \multirow{3}{*}{ResNet18} & \multirow{3}{*}{Places365}
    & CLIP-Dissect & $0.0125$ & $\underline{0.1472}$  \\
    & & INVERT & $<10^{-4}$ & $<10^{-4}$ \\
    & & DnD & $<10^{-4}$ & $0.0005$ \\
    \midrule
    \multirow{3}{*}{ResNet50} & \multirow{3}{*}{ImageNet}
    & CLIP-Dissect & $<10^{-4}$ & $<10^{-4}$ \\
    & & INVERT & $<10^{-4}$ & $<10^{-4}$ \\
    & & DnD & $<10^{-4}$ & $<10^{-4}$ \\
    \midrule
    \multirow{3}{*}{ResNet50} & \multirow{3}{*}{Places365}
    & CLIP-Dissect & $0.0020$ & $0.0083$ \\
    & & INVERT & $<10^{-4}$ & $<10^{-4}$ \\
    & & DnD & $<10^{-4}$ & $<10^{-4}$ \\
    \midrule
    \multirow{3}{*}{ViT-B/16} & \multirow{3}{*}{ImageNet}
    & CLIP-Dissect & $<10^{-4}$ & $<10^{-4}$ \\
    & & INVERT & $<10^{-4}$ & $<10^{-4}$ \\
    & & DnD & $<10^{-4}$ & $<10^{-4}$ \\
    \midrule
    \multirow{3}{*}{DINOv3} & \multirow{3}{*}{LVD-1689M}
    & CLIP-Dissect & $<10^{-4}$ & $<10^{-4}$ \\
    & & INVERT & $<10^{-4}$ & $<10^{-4}$ \\
    & & DnD & $<10^{-4}$ & $<10^{-4}$ \\
    \bottomrule
    \end{tabular}
\end{table}

\begin{table}[ht]
    \centering
    \caption{\textbf{Statistical significance tests across T2I models.} Building on \Cref{fig:sd_models}, we use a Wilcoxon signed-rank test to calculate the statistical significance of activation differences for 100 random ResNet50 \texttt{avgpool} neurons when replacing \LINE{}'s default SDXL model with FLUX.1[dev] or SD1.5. The $p$-values confirm no significant difference between SDXL and FLUX ($p > 0.05$), while SDXL significantly outperforms SD1.5 ($p < 0.05$).}
    \label{tab:stat_t2i}
    \vspace{0.5em}
    \begin{tabular}{l c}
    \toprule
    \textbf{Method of reference} & \textbf{\textit{p}-value} \\
    \midrule
    FLUX & 0.7965\\
    SD1.5 & 0.0375\\
    \bottomrule
    \end{tabular}
\end{table}

\newpage
\subsection{Causal Evaluations Pipeline}
\label{app:causal}

In this section, we detail the pipeline for generating ablated samples by removing the visual concepts identified by \LINE{} (as illustrated in \Cref{fig:main_causal}). To remove the visual concept $t$ (associated with neuron $n$) from an original image $x$, we utilized an image-editing generative model, specifically \emph{Qwen-Image-Edit-2511}. Given input image $x$ and a modification prompt, the model synthesizes the edited image $\hat{x}$. The prompt is constructed as follows:

\begin{tcolorbox}[
  breakable,
  enhanced,
  colback=white,
  colframe=black,
  boxrule=0.5pt,
  title=\textbf{Concept Image Removing},
  fonttitle=\bfseries
]
    Remove the \{concept\} from the image
\end{tcolorbox}

where \texttt{\{concept\}} corresponds to $t$. Using this setup, we generated ablated versions of the Places365 validation images for a random subset of neurons in the ResNet50 \texttt{avgpool} layer that strongly associate with the neuron's \LINE{} label. The resulting visual ablations, along with the corresponding changes in neuron activation, were initially presented in \Cref{fig:main_causal}.

\begin{figure}
    \centering
    \includegraphics[width=0.93\linewidth]{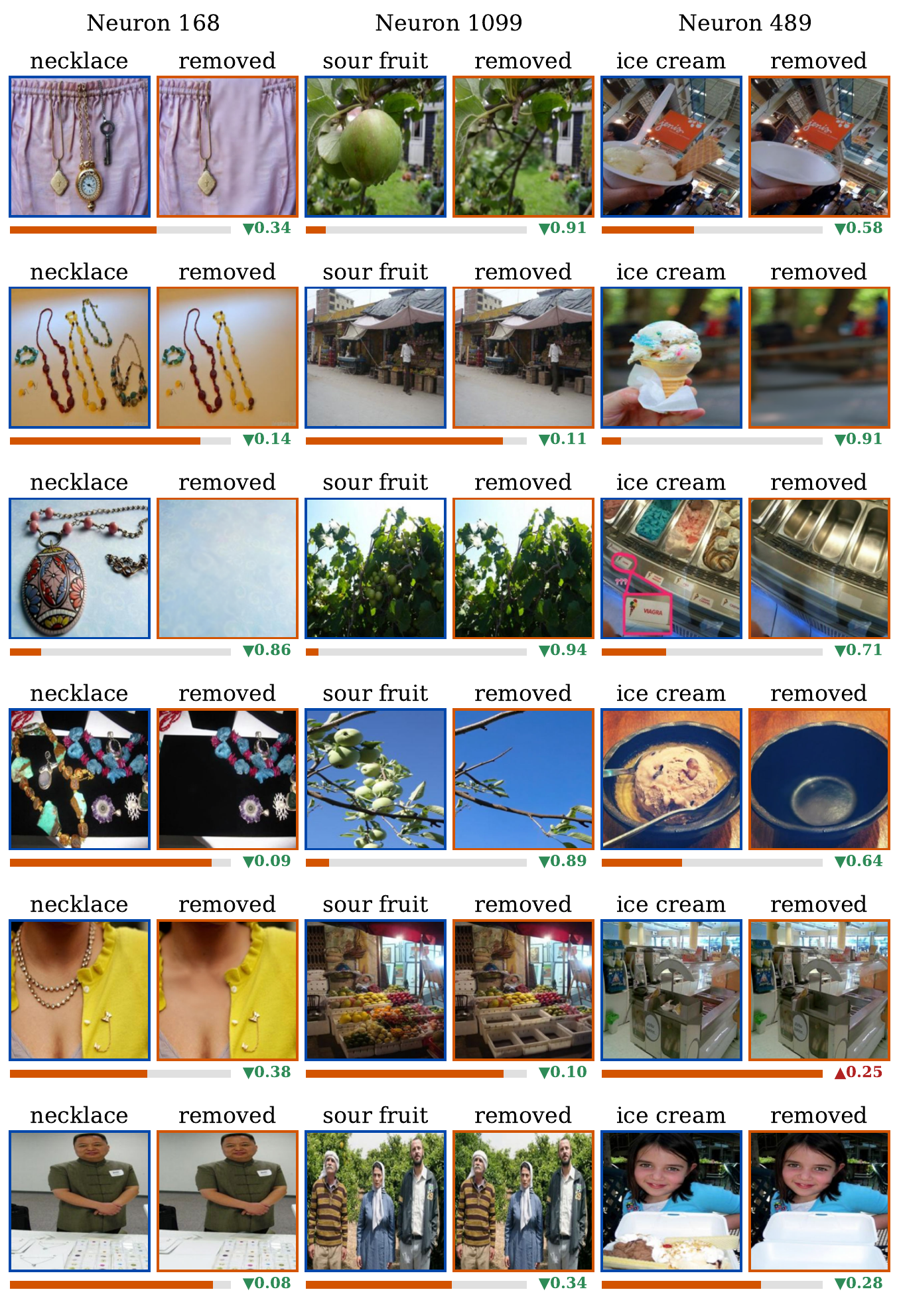}
    \caption{\textbf{Ablation of visual concepts derived from \LINE{}.} We show side-by-side comparisons demonstrating the effect of removing the \LINE{}-defined concept $t$ from highly activating images for the ResNet50 \texttt{avgpool} layer using an image-editing generative model. Concept removal via generative models is not always perfect and still requires manual inspection; for instance, removal failures can be observed in the 4th and 6th rows for neuron 168 and the 5th row for neuron 1099, whereas neuron 489 demonstrates largely successful ablations, though causing an increase in activation for a single sample out of six. Original images are outlined in \textcolor{ourblue}{\textbf{blue}}, ablated versions in \textcolor{ourorange}{\textbf{orange}}, and the resulting relative change in activation is marked \textcolor{ourgreen}{\textbf{green}} if there has been a decrease, or \textcolor{ourred}{\textbf{red}} if there has been an increase.}
    \label{fig:app_causal}
\end{figure}

\Cref{fig:app_causal} provides a broader set of edited images that highlight the current limitations of this automated ablation pipeline. Specifically, this trace reveals two primary issues: (1) the generative model occasionally fails to completely remove the targeted concept  (e.g., neuron 168) or erroneously alters unrelated features (e.g., neuron 489), requiring manual inspection; and (2) the resolution of the edited image $\hat{x}$ sometimes differs from the original image $x$, which may introduce unexpected artifacts into the model's behavior and artificially decrease activations (a phenomenon we observed when scaling up the number of ablated images).

\clearpage
\section{Extended Qualitative and Quantitative Analysis}
\label{app:extended_analysis}
In this section we extend the analysis from the main paper with: a comparison of \LINE{} variants in \Cref{app:quantitative_comparison}, multi-label descriptions in \Cref{app:multi_comparison}, the impact of vocabulary size in \Cref{app:initial_comparison}, an extended qualitative comparison in \Cref{app:extended_visual}, and evaluations on earlier layers in \Cref{app:lower_evaluations}.

\newpage
\subsection{Quantitative Comparison of \LINE{}}
\label{app:quantitative_comparison}
\begin{figure}[h]
    \centering
    \includegraphics[width=\linewidth]{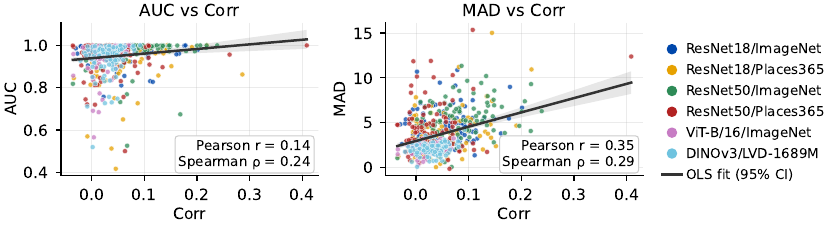}
    \caption{\textbf{Correlation between AUC/MAD and Corr metrics.} AUC and MAD scores of neurons discovered by \LINE{} correlate only weakly with the Corr metric, as indicated by low Pearson and Spearman correlation coefficients.}
    \label{fig:correlation}
\end{figure}

We perform an extended evaluation on additional variants of our method. \LINE{}~(FLUX) uses FLUX.1[dev] as a T2I model. Vocab Only takes the best concept from the initial scoreboard, without the iterative improvement. \Cref{tab:comparison_table_extended} displays the extended VLM correlation results along with CoSy benchmark results. \LINE{}~(FLUX) matches the base method on the correlation metric but performs slightly worse on CoSy, while Vocab Only occasionally falls behind the best baseline. Vocab Only occasionally outperforms \LINE{} on the correlation metric. We attribute this to the scoring function $\psi$ aligned with the AUC and MAD metrics, which correlate only weakly with Corr (see \Cref{fig:correlation}). While $\psi$ was designed to align with the CoSy benchmark, future works may find that tuning it toward the metrics proposed by \citet{oikarinen2025evaluating} yields better practical results.

\begin{table}[h]
    \centering
    \caption{\textbf{Extended Evaluation from \Cref{tab:comparison_table}.} We compare \LINE{}, \LINE{}~(Vocab Only), and \LINE{}~(FLUX) against the best baseline. We report the CoSy metrics MAD and AUC, as well as the Pearson correlation ($\times 10^{-1}$) between neuron activations and the SigLIP similarity. For all metrics, higher is better ($\uparrow$); \textbf{best} and \underline{second-best} results in each row are highlighted in bold and underlined, respectively.}
    \label{tab:comparison_table_extended}
    \vspace{0.5em}
    \begin{tabular}{l l l c c c c}
    \toprule
    \textbf{Model} & \textbf{Dataset} & \textbf{Metric} & \textbf{\LINE} & \textbf{Vocab Only} & \textbf{\LINE (FLUX)} & \textbf{Baseline} \\
    \midrule
    \multirow{3}{*}{ResNet-18} & \multirow{3}{*}{ImageNet}
      & AUC & $\mathbf{0.97}_{\pm 0.04}$ & $0.92_{\pm 0.12}$ & $\underline{0.95}_{\pm 0.10}$ & $0.90_{\pm 0.14}$ \\
     & & MAD & $\mathbf{4.76}_{\pm 1.91}$ & $3.72_{\pm 2.08}$ & $\underline{4.45}_{\pm 2.06}$ & $3.42_{\pm 2.17}$ \\
     & & Corr & $0.57_{\pm 0.49}$ & $\underline{0.63}_{\pm 0.49}$ & $\mathbf{0.66}_{\pm 0.46}$ & $0.47_{\pm 0.45}$ \\
    \midrule
    \multirow{3}{*}{ResNet-18} & \multirow{3}{*}{Places365}
      & AUC & $\mathbf{0.93}_{\pm 0.10}$ & $0.87_{\pm 0.16}$ & $\underline{0.90}_{\pm 0.16}$ & $0.89_{\pm 0.15}$ \\
     & & MAD & $\mathbf{3.77}_{\pm 2.31}$ & $2.97_{\pm 2.32}$ & $3.51_{\pm 2.42}$ & $\underline{3.63}_{\pm 2.78}$ \\
     & & Corr & $0.51_{\pm 0.54}$ & $0.41_{\pm 0.53}$ & $\underline{0.53}_{\pm 0.46}$ & $\mathbf{0.64}_{\pm 0.58}$ \\
    \midrule
    \multirow{3}{*}{ResNet-50} & \multirow{3}{*}{ImageNet}
      & AUC & $\mathbf{0.97}_{\pm 0.07}$ & $0.92_{\pm 0.14}$ & $\underline{0.95}_{\pm 0.10}$ & $0.88_{\pm 0.15}$ \\
     & & MAD & $\mathbf{5.28}_{\pm 2.10}$ & $4.47_{\pm 2.43}$ & $\underline{4.92}_{\pm 2.16}$ & $3.56_{\pm 2.52}$ \\
     & & Corr & $\mathbf{0.81}_{\pm 0.57}$ & $\underline{0.77}_{\pm 0.62}$ & $0.65_{\pm 0.45}$ & $0.66_{\pm 0.47}$ \\
    \midrule
    \multirow{3}{*}{ResNet-50} & \multirow{3}{*}{Places365}
      & AUC & $\mathbf{0.94}_{\pm 0.08}$ & $0.88_{\pm 0.17}$ & $\underline{0.93}_{\pm 0.09}$ & $0.89_{\pm 0.15}$ \\
     & & MAD & $\mathbf{4.27}_{\pm 2.81}$ & $3.38_{\pm 2.76}$ & $\underline{3.84}_{\pm 2.61}$ & $3.69_{\pm 2.90}$ \\
     & & Corr & $\underline{0.36}_{\pm 0.56}$ & $0.26_{\pm 0.41}$ & $0.35_{\pm 0.45}$ & $\mathbf{0.57}_{\pm 0.54}$ \\
    \midrule
    \multirow{3}{*}{ViT-B/16} & \multirow{3}{*}{ImageNet}
      & AUC & $\mathbf{0.94}_{\pm 0.07}$ & $0.90_{\pm 0.12}$ & $\underline{0.92}_{\pm 0.10}$ & $0.79_{\pm 0.18}$ \\
     & & MAD & $\mathbf{2.12}_{\pm 0.54}$ & $1.82_{\pm 0.71}$ & $\underline{1.90}_{\pm 0.65}$ & $1.05_{\pm 0.70}$ \\
     & & Corr & $0.26_{\pm 0.21}$ & $\mathbf{0.31}_{\pm 0.22}$ & $\underline{0.27}_{\pm 0.21}$ & $0.23_{\pm 0.21}$ \\
    \midrule
    \multirow{3}{*}{DINOv3} & \multirow{3}{*}{LVD-1689M}
      & AUC & $\mathbf{0.94}_{\pm 0.08}$ & $0.88_{\pm 0.14}$ & $\underline{0.91}_{\pm 0.12}$ & $0.76_{\pm 0.18}$ \\
     & & MAD & $\mathbf{2.05}_{\pm 0.68}$ & $1.69_{\pm 0.85}$ & $\underline{1.86}_{\pm 0.78}$ & $1.12_{\pm 1.06}$ \\
     & & Corr & $\underline{0.32}_{\pm 0.23}$ & $\mathbf{0.35}_{\pm 0.25}$ & $0.29_{\pm 0.21}$ & $0.24_{\pm 0.24}$ \\
    \midrule
    \multirow{3}{*}{\textbf{Overall}} & \multirow{3}{*}{-}
      & AUC & $\mathbf{0.95}_{\pm 0.08}$ & $0.90_{\pm 0.14}$ & $\underline{0.93}_{\pm 0.12}$ & $0.84_{\pm 0.19}$ \\
     & & MAD & $\mathbf{3.71}_{\pm 2.27}$ & $3.01_{\pm 2.24}$ & $\underline{3.41}_{\pm 2.26}$ & $2.74_{\pm 2.50}$ \\
     & & Corr & $\mathbf{0.47}_{\pm 0.50}$ & $0.45_{\pm 0.48}$ & $\underline{0.46}_{\pm 0.42}$ & $0.41_{\pm 0.49}$ \\
    \bottomrule
    \end{tabular}
\end{table}

\clearpage
\subsection{Multi-label Description}
\label{app:multi_comparison}
Inspired by INVERT, which combines multiple labels into a single neuron description, we post-hoc concatenate the top-k \LINE{} concepts from the final scoreboard using conjunctions.
In \Cref{tab:comparison_table_three} we present benchmark results for the top-3 concepts combined using '\texttt{AND}'. INVERT-3 denotes INVERT results obtained using $L=3$ and having logic operators replaced with corresponding words (e.g., \texttt{OR} instead of \texttt{|}). Although \LINE-3 scores are consistently better, we have decided not to present it as our main method, because the resulting descriptions can be overly complicated or consist of synonyms. We report this finding to show that \LINE{}'s scoreboard is a component that future work can build upon with composition methods like INVERT~\citep{bykov2024labeling,rosa2026guaranteed}.

\begin{table}[h]
    \centering
    \caption{\textbf{Evaluation of single- and multi-label descriptions.} We compare \LINE{} against INVERT, \LINE-3, and INVERT-3. We report the CoSy metrics MAD and AUC, as well as the Pearson correlation ($\times 10^{-1}$) from the VLM simulation. For all metrics, higher is better ($\uparrow$). The best and second-best results in each row are highlighted in \textbf{bold} and \underline{underlined}, respectively.}
    \label{tab:comparison_table_three}
    \vspace{0.5em}
    \begin{tabular}{l l l c c c c}
    \toprule
    \textbf{Model} & \textbf{Dataset} & \textbf{Metric} & \textbf{\LINE} & \textbf{INVERT} & \textbf{\LINE-3} & \textbf{INVERT-3} \\
    \midrule
    \multirow{3}{*}{ResNet-18} & \multirow{3}{*}{ImageNet}
      & AUC & $\underline{0.97}_{\pm 0.04}$ & $0.85_{\pm 0.14}$ & $\mathbf{0.98}_{\pm 0.03}$ & $0.92_{\pm 0.07}$ \\
     & & MAD & $\underline{4.76}_{\pm 1.91}$ & $2.16_{\pm 1.61}$ & $\mathbf{4.89}_{\pm 1.82}$ & $2.76_{\pm 1.38}$ \\
     & & Corr & $0.57_{\pm 0.49}$ & $0.47_{\pm 0.45}$ & $\mathbf{0.97}_{\pm 0.49}$ & $\underline{0.62}_{\pm 0.42}$ \\
    \midrule
    \multirow{3}{*}{ResNet-18} & \multirow{3}{*}{Places365}
      & AUC & $\underline{0.93}_{\pm 0.10}$ & $0.80_{\pm 0.18}$ & $\mathbf{0.96}_{\pm 0.07}$ & $0.87_{\pm 0.12}$ \\
     & & MAD & $\underline{3.77}_{\pm 2.31}$ & $1.85_{\pm 1.70}$ & $\mathbf{4.17}_{\pm 2.02}$ & $2.39_{\pm 1.65}$ \\
     & & Corr & $\underline{0.51}_{\pm 0.54}$ & $0.20_{\pm 0.33}$ & $\mathbf{0.73}_{\pm 0.46}$ & $0.48_{\pm 0.40}$ \\
    \midrule
    \multirow{3}{*}{ResNet-50} & \multirow{3}{*}{ImageNet}
      & AUC & $\underline{0.97}_{\pm 0.07}$ & $0.88_{\pm 0.15}$ & $\mathbf{0.98}_{\pm 0.03}$ & $0.94_{\pm 0.08}$ \\
     & & MAD & $\underline{5.28}_{\pm 2.10}$ & $2.81_{\pm 1.95}$ & $\mathbf{5.56}_{\pm 1.98}$ & $3.48_{\pm 1.83}$ \\
     & & Corr & $0.81_{\pm 0.57}$ & $0.48_{\pm 0.51}$ & $\mathbf{1.11}_{\pm 0.52}$ & $\underline{0.87}_{\pm 0.44}$ \\
    \midrule
    \multirow{3}{*}{ResNet-50} & \multirow{3}{*}{Places365}
      & AUC & $\underline{0.94}_{\pm 0.08}$ & $0.84_{\pm 0.14}$ & $\mathbf{0.96}_{\pm 0.05}$ & $0.90_{\pm 0.10}$ \\
     & & MAD & $\underline{4.27}_{\pm 2.81}$ & $2.04_{\pm 1.70}$ & $\mathbf{4.42}_{\pm 2.46}$ & $2.44_{\pm 1.54}$ \\
     & & Corr & $0.36_{\pm 0.56}$ & $0.16_{\pm 0.30}$ & $\mathbf{0.65}_{\pm 0.61}$ & $\underline{0.48}_{\pm 0.45}$ \\
    \midrule
    \multirow{3}{*}{ViT-B/16} & \multirow{3}{*}{ImageNet}
      & AUC & $\underline{0.94}_{\pm 0.07}$ & $0.79_{\pm 0.18}$ & $\mathbf{0.95}_{\pm 0.05}$ & $0.83_{\pm 0.13}$ \\
     & & MAD & $\underline{2.12}_{\pm 0.54}$ & $1.05_{\pm 0.70}$ & $\mathbf{2.24}_{\pm 0.59}$ & $1.31_{\pm 0.69}$ \\
     & & Corr & $0.26_{\pm 0.21}$ & $0.23_{\pm 0.21}$ & $\mathbf{0.36}_{\pm 0.18}$ & $\underline{0.33}_{\pm 0.17}$ \\
    \midrule
    \multirow{3}{*}{DINOv3} & \multirow{3}{*}{LVD-1689M}
      & AUC & $\mathbf{0.94}_{\pm 0.08}$ & $0.76_{\pm 0.18}$ & $\underline{0.92}_{\pm 0.10}$ & $0.80_{\pm 0.16}$ \\
     & & MAD & $\mathbf{2.05}_{\pm 0.68}$ & $0.95_{\pm 0.73}$ & $\underline{1.88}_{\pm 0.70}$ & $1.15_{\pm 0.73}$ \\
     & & Corr & $\underline{0.32}_{\pm 0.23}$ & $0.24_{\pm 0.24}$ & $\mathbf{0.47}_{\pm 0.19}$ & $0.31_{\pm 0.18}$ \\
    \midrule
    \multirow{3}{*}{\textbf{Overall}} & \multirow{3}{*}{-}
      & AUC & $\underline{0.95}_{\pm 0.08}$ & $0.82_{\pm 0.17}$ & $\mathbf{0.96}_{\pm 0.06}$ & $0.88_{\pm 0.12}$ \\
     & & MAD & $\underline{3.71}_{\pm 2.27}$ & $1.81_{\pm 1.61}$ & $\mathbf{3.86}_{\pm 2.20}$ & $2.25_{\pm 1.59}$ \\
     & & Corr & $0.47_{\pm 0.50}$ & $0.24_{\pm 0.38}$ & $\mathbf{0.72}_{\pm 0.51}$ & $\underline{0.51}_{\pm 0.41}$ \\
    \bottomrule
    \end{tabular}
\end{table}

\clearpage
\subsection{Vocabulary Size Impact}
\label{app:initial_comparison}

\begin{figure}[h]
    \centering
    \includegraphics[width=1.0\linewidth]{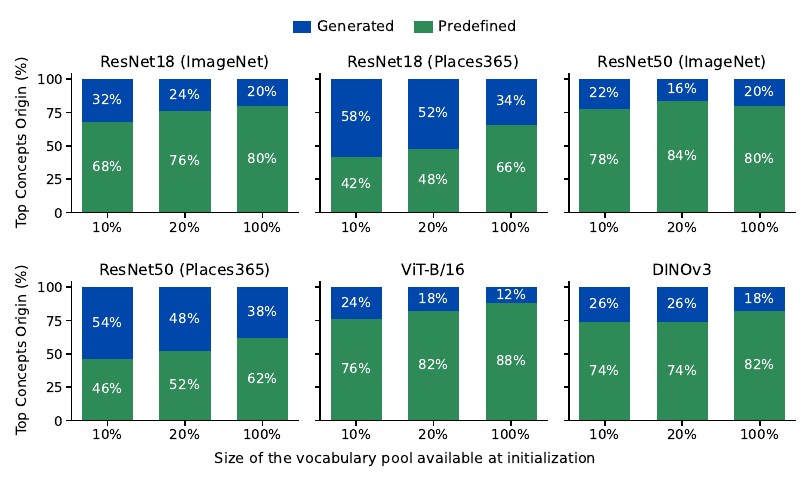}
    \caption{\textbf{Fraction of predefined vs. generated labels in reduced initial vocabulary pool.} We compare the contribution of \LINE{}'s iterative loop (\textcolor{ourblue}{\textbf{generated}}) when the initial vocabulary (\textcolor{ourgreen}{\textbf{predefined}}) is randomly reduced to 20\% and 10\% of the original pool (100\%).}

    \label{fig:comparison_table_difference_colored}
\end{figure}

 We evaluate \LINE{} on a random subset of the initial concepts and present the results in \Cref{tab:pool_ratio_ablation}. As the candidate pool shrinks from 100\% to 10\%, performance generally decreases. The effect is small in absolute terms, as removing 90\% of the vocabulary costs at most a $0.06$ drop in AUC, demonstrating \LINE's robustness even under severe computational budgets.
\Cref{fig:comparison_table_difference_colored} illustrates how the origin of the selected concept changes.\LINE{} partially compensates for the weaker initialization by generating a higher proportion of the final concepts through refinement process, reaching up to 58\%.

\begin{table}[h]
    \centering
    \caption{\textbf{Performance under a reduced initial vocabulary pool.} For 50 neurons per model and layer (sampled independently of \Cref{tab:comparison_table}), we compare \LINE{} on the CoSy benchmark when the scoreboard is initialized from a subsampled vocabulary. Values in parentheses give the change relative to the full pool.}
    \label{tab:pool_ratio_ablation}
    \vspace{0.5em}
    \begin{tabular}{l l c c c c}
    \toprule
    \textbf{Model} & \textbf{Dataset} & \textbf{Metric} & \textbf{100\%} & \textbf{20\%} & \textbf{10\%} \\
    \midrule
    \multirow{2}{*}{ResNet18} & \multirow{2}{*}{ImageNet} & AUC & $\mathbf{0.97}$ & $0.96$ \textit{(-0.01)} & $0.94$ \textit{(-0.03)} \\
     & & MAD & $\mathbf{4.72}$ & $3.74$ \textit{(-0.98)} & $3.26$ \textit{(-1.46)} \\
    \midrule
    \multirow{2}{*}{ResNet18} & \multirow{2}{*}{Places365} & AUC & $\mathbf{0.95}$ & $0.89$ \textit{(-0.06)} & $0.90$ \textit{(-0.05)} \\
     & & MAD & $\mathbf{4.12}$ & $3.13$ \textit{(-0.99)} & $2.80$ \textit{(-1.32)} \\
    \midrule
    \multirow{2}{*}{ResNet50} & \multirow{2}{*}{ImageNet} & AUC & $\mathbf{0.97}$ & $0.95$ \textit{(-0.02)} & $0.95$ \textit{(-0.02)} \\
     & & MAD & $\mathbf{5.63}$ & $4.21$ \textit{(-1.42)} & $3.74$ \textit{(-1.89)} \\
    \midrule
    \multirow{2}{*}{ResNet50} & \multirow{2}{*}{Places365} & AUC & $\mathbf{0.92}$ & $0.91$ \textit{(-0.01)} & $0.88$ \textit{(-0.04)} \\
     & & MAD & $\mathbf{4.19}$ & $3.06$ \textit{(-1.13)} & $2.93$ \textit{(-1.26)} \\
    \midrule
    \multirow{2}{*}{ViT-B/16} & \multirow{2}{*}{ImageNet} & AUC & $\mathbf{0.95}$ & $0.90$ \textit{(-0.05)} & $0.89$ \textit{(-0.06)} \\
     & & MAD & $\mathbf{2.03}$ & $1.64$ \textit{(-0.39)} & $1.61$ \textit{(-0.42)} \\
     \midrule
    \multirow{2}{*}{DINOv3} & \multirow{2}{*}{LVD-1689M} & AUC & $\mathbf{0.93}$ & $0.89$ \textit{(-0.04)} & $0.87$ \textit{(-0.06)} \\
     & & MAD & $\mathbf{2.01}$ & $1.71$ \textit{(-0.30)} & $1.50$ \textit{(-0.51)} \\
    \bottomrule
    \end{tabular}
\end{table}

\newpage
\subsection{Qualitative Comparison}
\label{app:extended_visual}

\begin{figure}[H]
    \centering
    \begin{subfigure}[t]{0.49\textwidth}
        \vspace{0pt}
        \includegraphics[width=\linewidth]{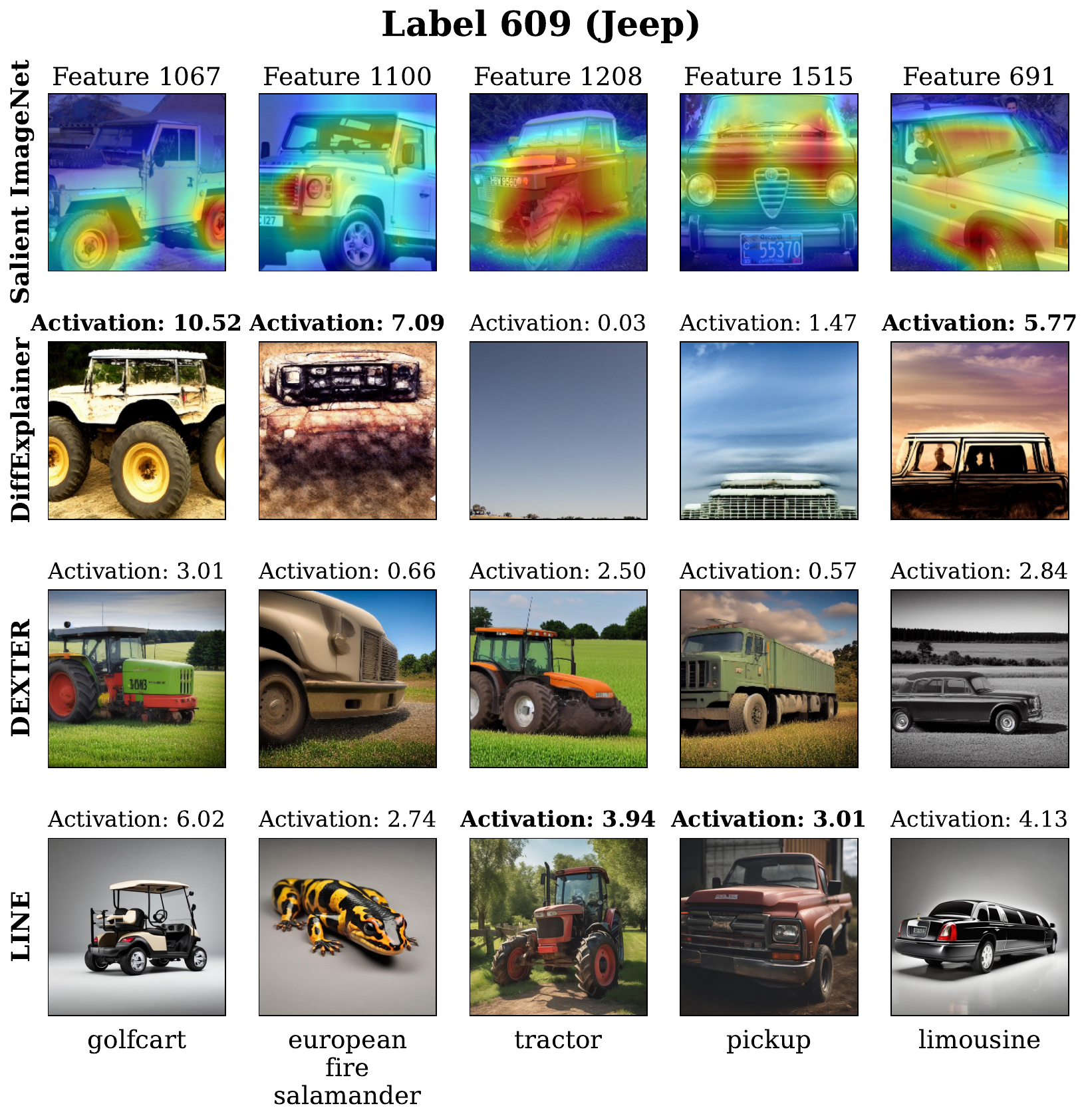}
    \end{subfigure}\hfill
    \begin{subfigure}[t]{0.49\textwidth}
        \vspace{0pt}
        \includegraphics[width=\linewidth]{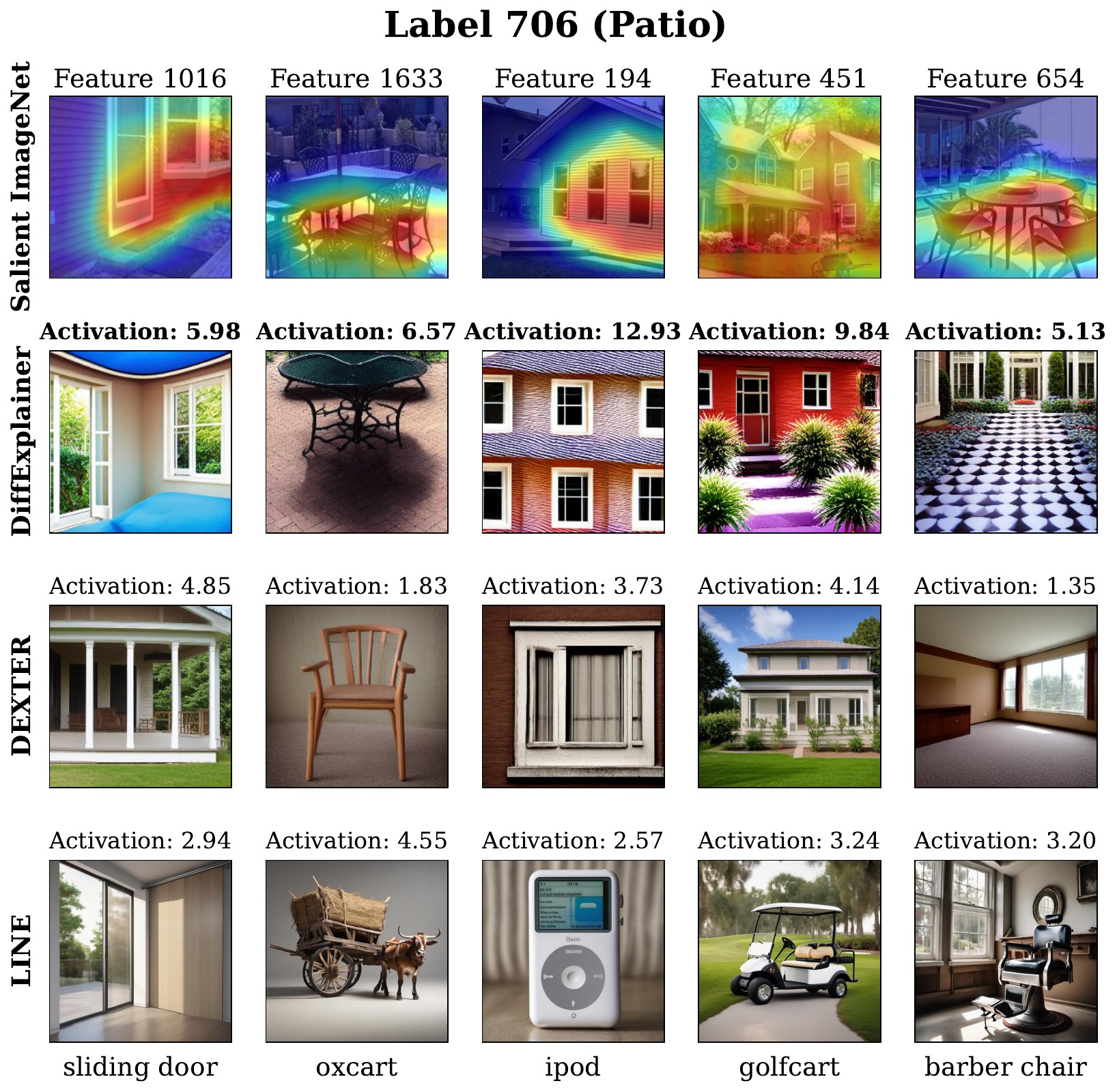}
    \end{subfigure}

    \begin{subfigure}[t]{0.49\textwidth}
        \vspace{0pt}
        \includegraphics[width=\linewidth]{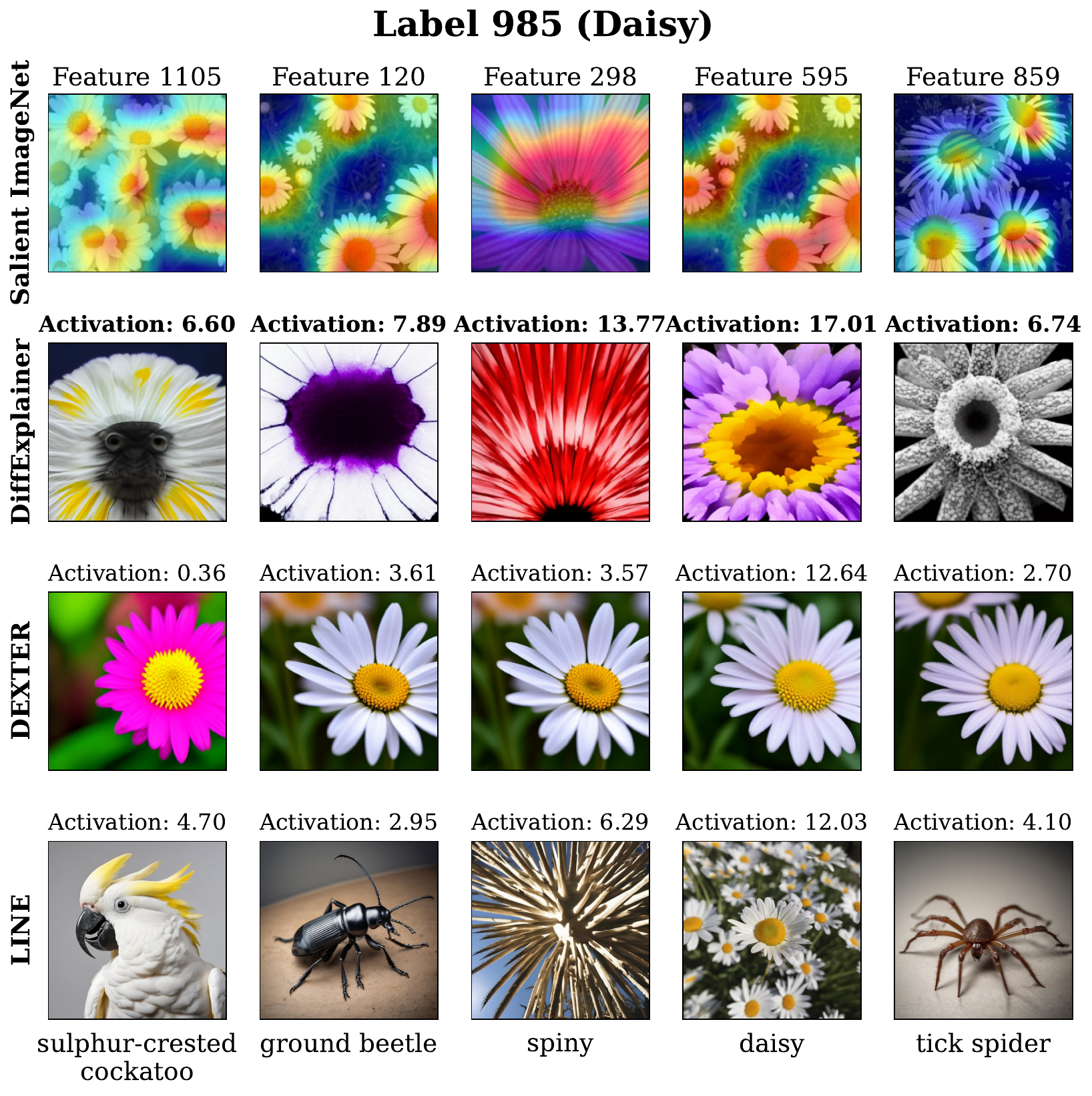}
    \end{subfigure}\hfill
    \begin{subfigure}[t]{0.49\textwidth}
        \vspace{0pt}
        \includegraphics[width=\linewidth]{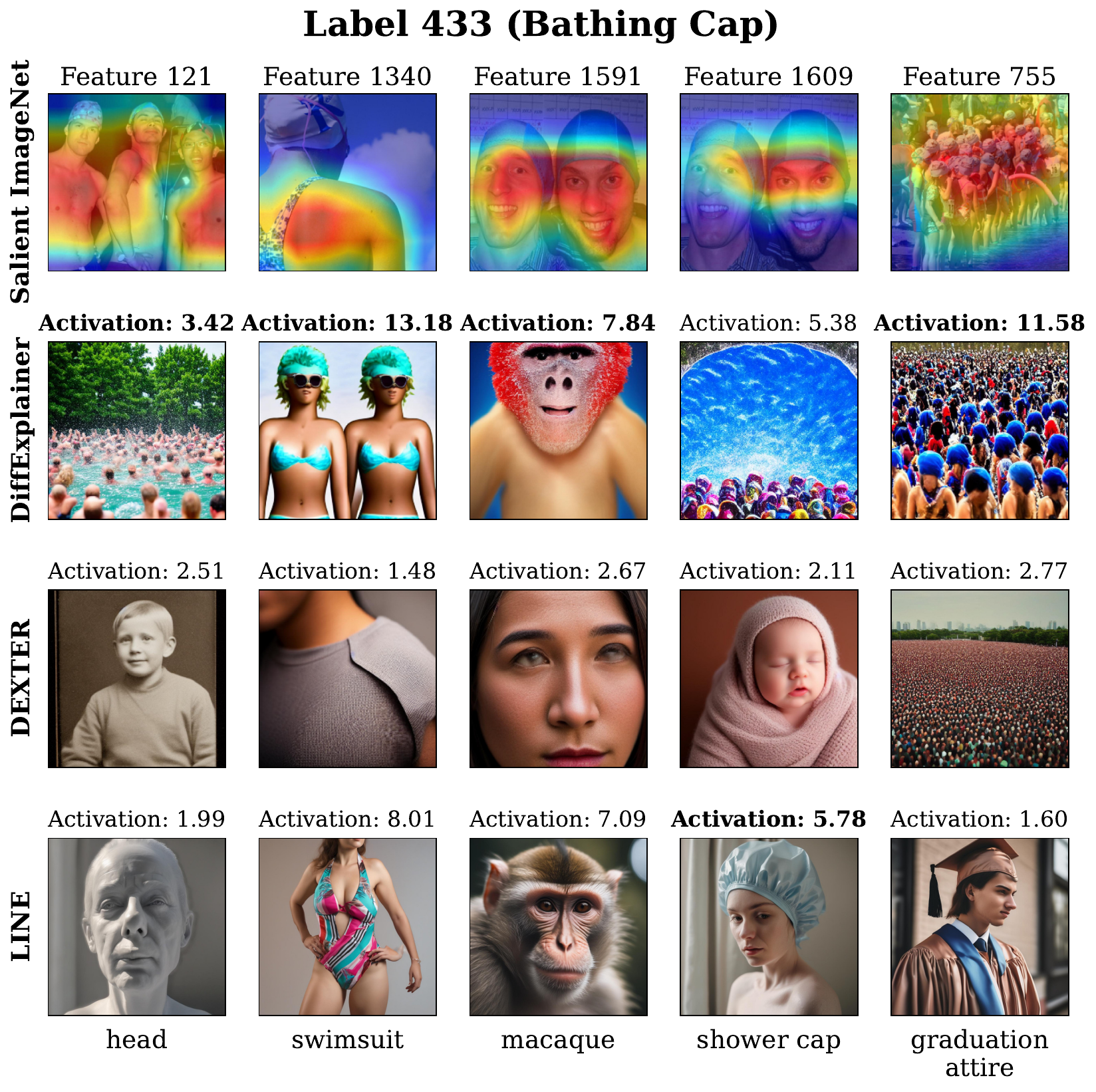}
    \end{subfigure}

    \begin{subfigure}[t]{0.49\textwidth}
        \vspace{0pt}
        \includegraphics[width=\linewidth]{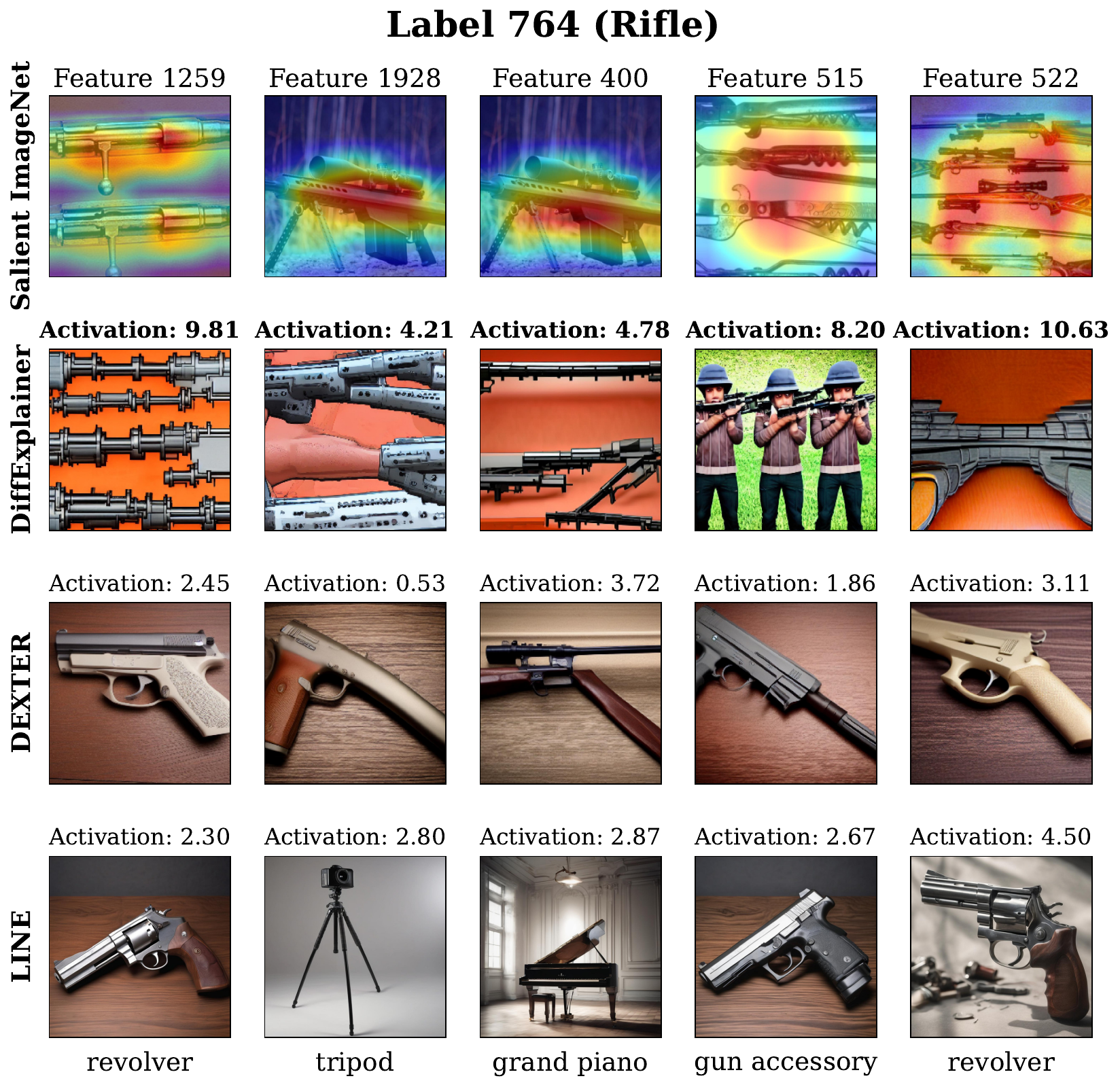}
    \end{subfigure}\hfill
    \begin{subfigure}[t]{0.49\textwidth}
        \vspace{0pt}
        \includegraphics[width=\linewidth]{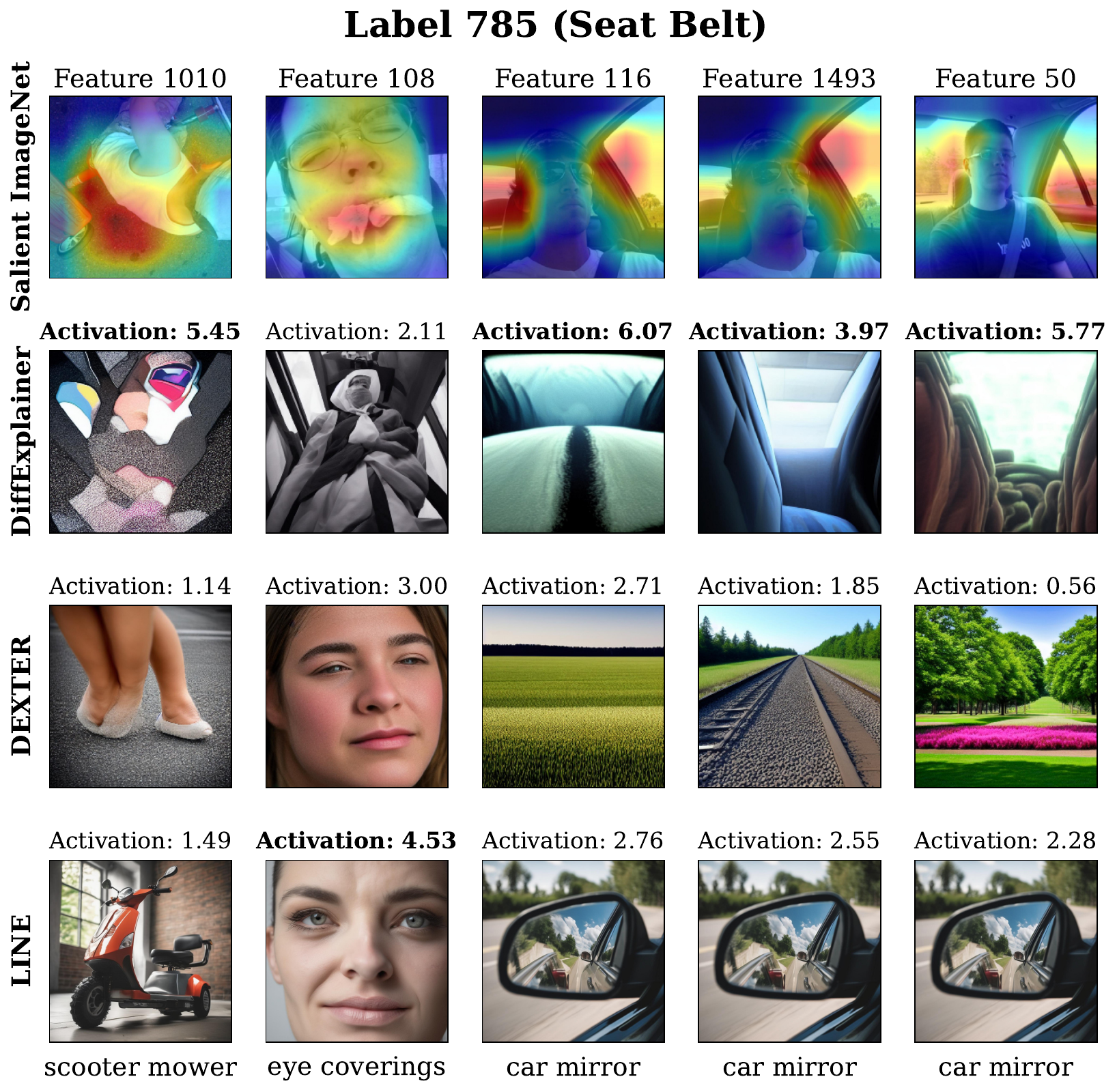}
    \end{subfigure}
    \caption{\textbf{Extended visual explanations on Salient ImageNet.} Extending the \Cref{fig:am_main}, we present results for 3 core classes (\emph{left}) and 3 biased classes (\emph{right}). Each class subfigure displays the top-5 features identified in the dataset, along with their attribution maps and explanations from DiffExplainer, DEXTER, and \LINE{}. Activation magnitudes appear above each image; \LINE{} feature descriptions are provided below.}
    \label{fig:am_full}
\end{figure}

\begin{figure}[H]
    \centering
    \includegraphics[width=1.0\linewidth]{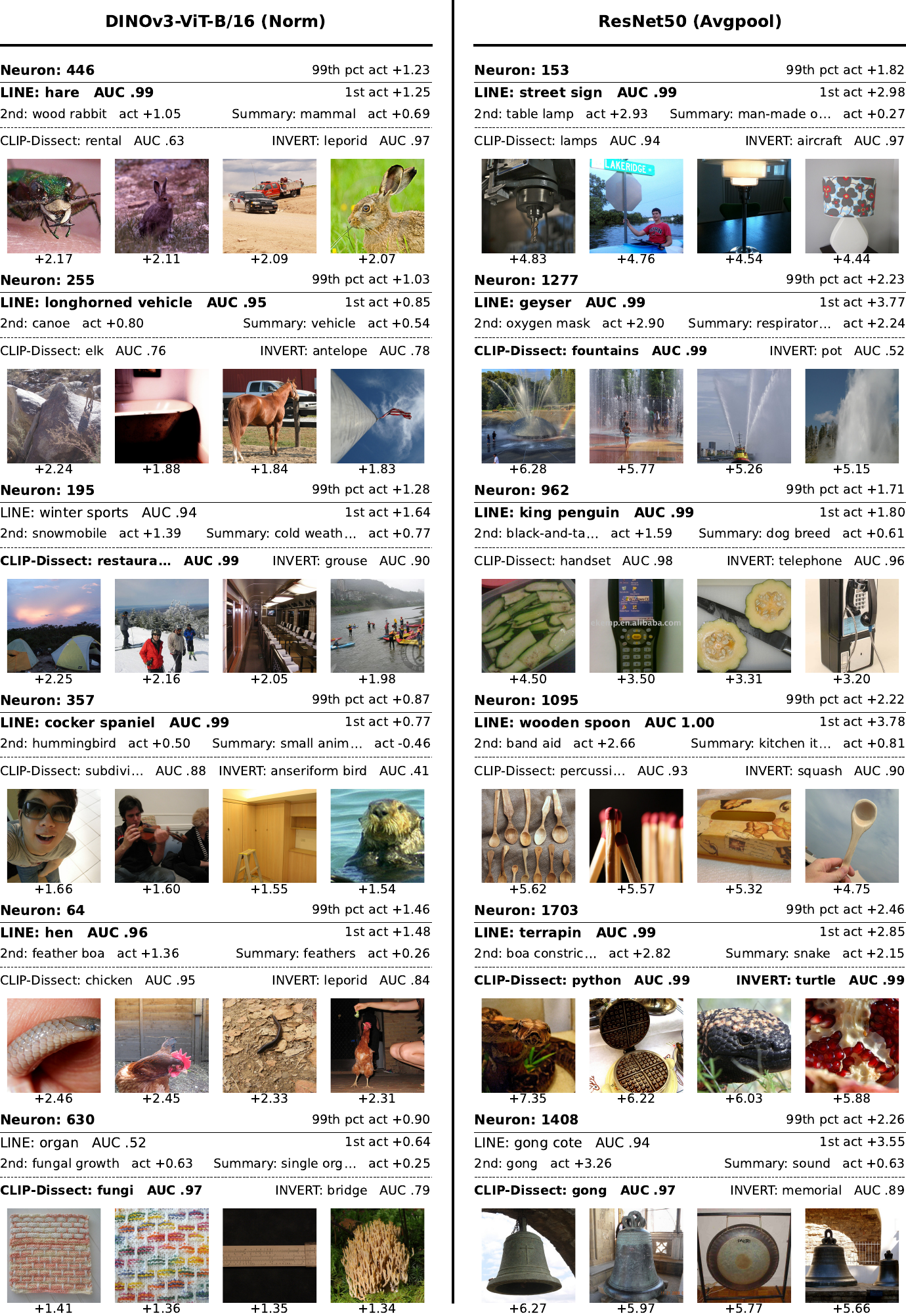}
    \caption{\textbf{Qualitative comparison of neuron descriptions.} We extend the qualitative analysis from \Cref{fig:main_qualitive} to the DINOv3 \texttt{norm} and ResNet50 \texttt{avgpool} layers. For each neuron, we show the top four activating images from ImageNet, the two highest-scoring \LINE{} descriptions together with the summary description and their activation scores, and the descriptions produced by CLIP-Dissect and INVERT, each reported with its CoSy AUC score. Every neuron is annotated with its 99th-percentile activation over the dataset, and the best method is highlighted in \textbf{bold}.}
    \label{fig:neuron_descriptions_dinov3}
\end{figure}

\begin{figure}[H]
    \centering
    \includegraphics[width=1.0\linewidth]{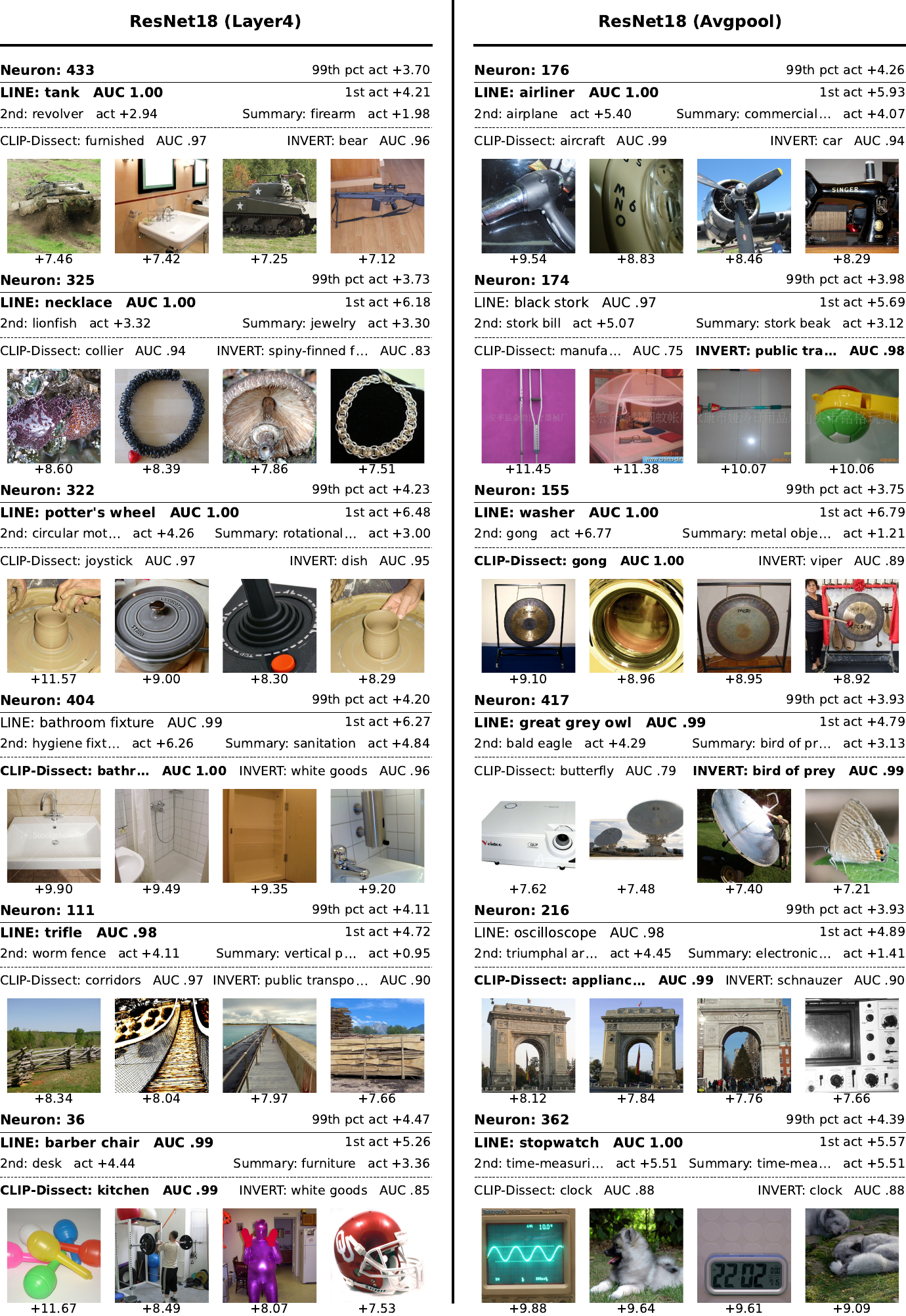}
    \caption{\textbf{Qualitative comparison of neuron descriptions in ImageNet ResNet18.} The best-performing method for each neuron, based on the AUC score, is highlighted in \textbf{bold}.}
    \label{fig:neuron_descriptions_resnet18_imagenet}
\end{figure}

\begin{figure}[H]
    \centering
    \includegraphics[width=1.0\linewidth]{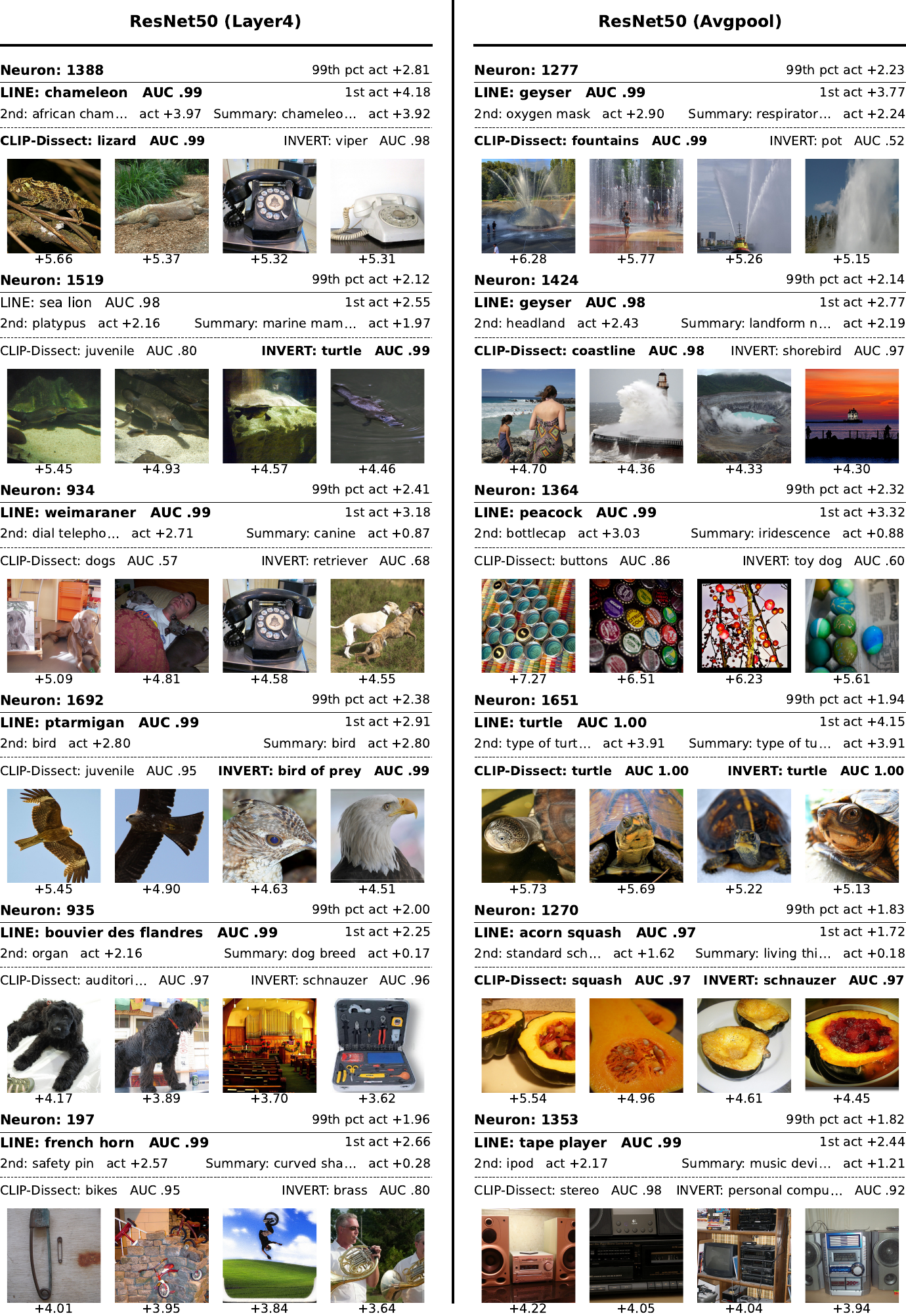}
    \caption{\textbf{Qualitative comparison of neuron descriptions in ImageNet ResNet50.} The best-performing method for each neuron, based on the AUC score, is highlighted in \textbf{bold}.}
    \label{fig:neuron_descriptions_resnet50_imagenet}
\end{figure}

\begin{figure}[H]
    \centering
    \includegraphics[width=1.0\linewidth]{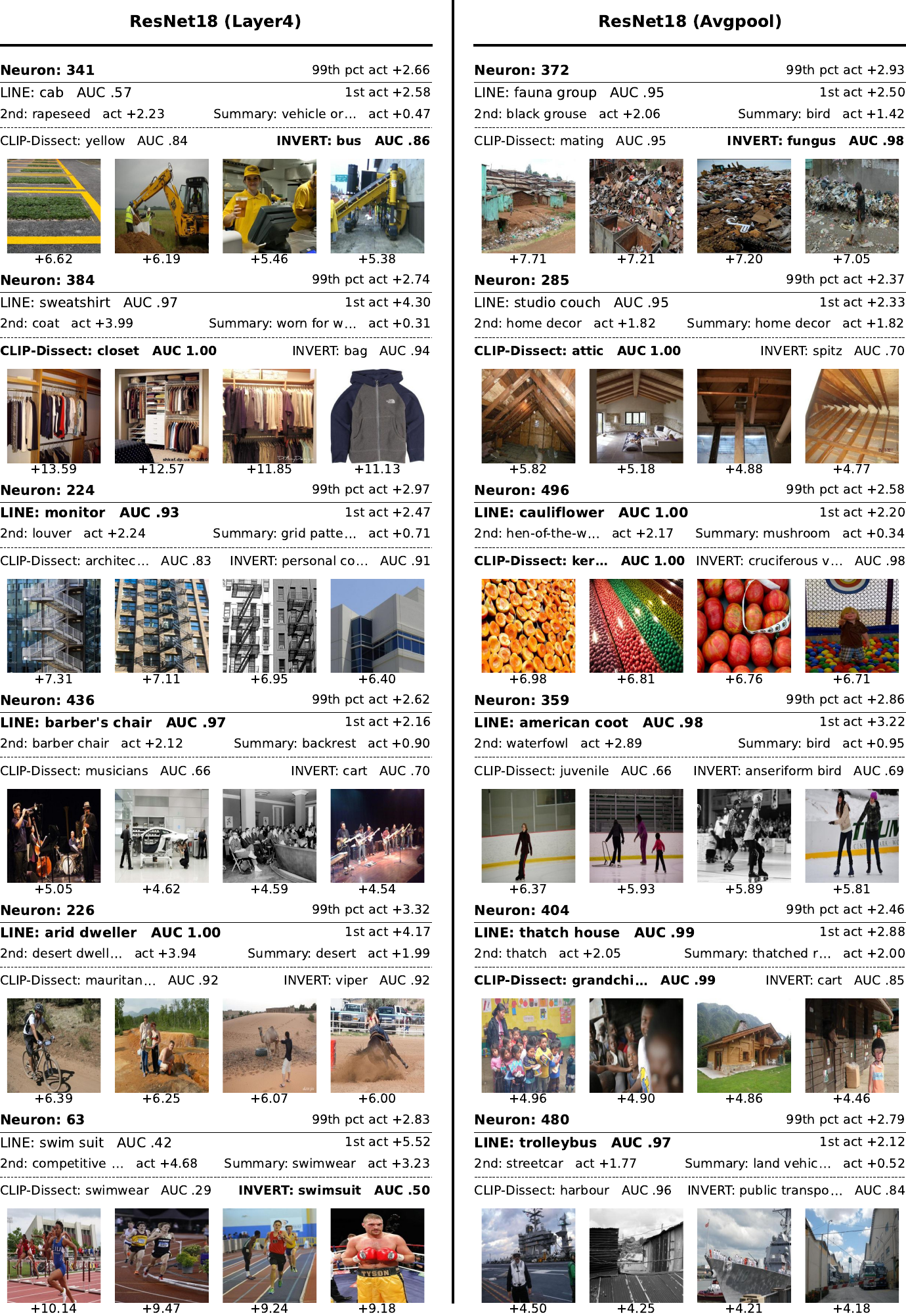}
    \caption{\textbf{Qualitative comparison of neuron descriptions in Places365 ResNet18.} The best-performing method for each neuron, based on the AUC score, is highlighted in \textbf{bold}.}
    \label{fig:neuron_descriptions_resnet18_places}
\end{figure}

\begin{figure}[H]
    \centering
    \includegraphics[width=1.0\linewidth]{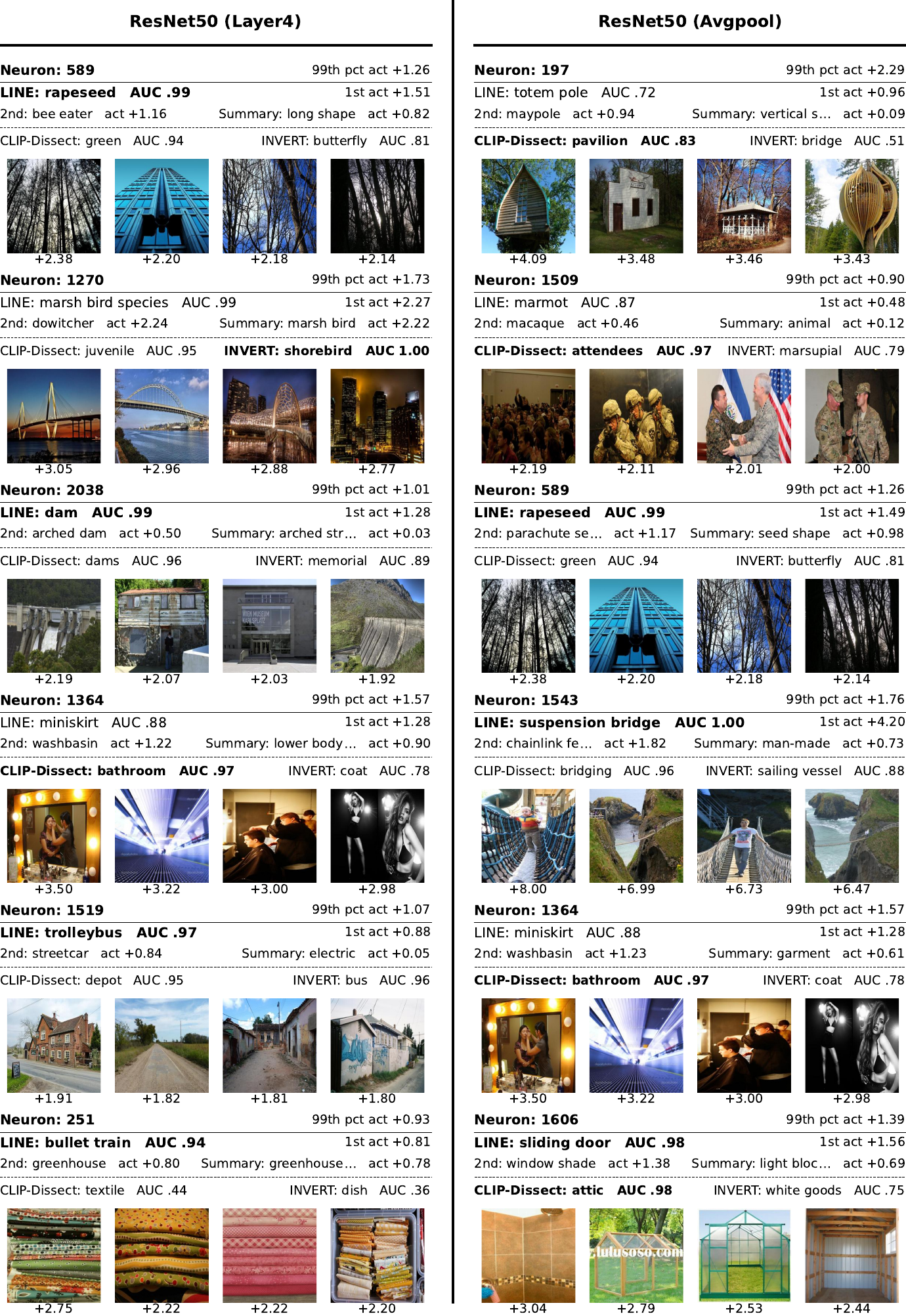}
    \caption{\textbf{Qualitative comparison of neuron descriptions in Places365 ResNet50.} The best-performing method for each neuron, based on the AUC score, is highlighted in \textbf{bold}.}
    \label{fig:neuron_descriptions_resnet50_places}
\end{figure}

To supplement the visualizations in \Cref{sec:qualitative,sec:visual_explanations}, we provide additional qualitative results across several model architectures. Specifically, \Cref{fig:am_full} presents 6 classes from a RobustResNet50 model trained on ImageNet-1K. Building on the evaluation protocol from the DEXTER~\citep{carnemolla2025dexter} and extending the results from \Cref{fig:am_main}, we selected 3 classes containing heavily spurious features and 3 classes relying mainly on core features from Salient ImageNet~\citep{singla2022salient}. For each evaluated class, we provided visual explanations of the top-5 features, categorized as either ``core'' (for the core classes) or ``spurious'' (for the spurious classes). In  \Cref{tab:am} we summarize all evaluated features from Salient ImageNet, alongside the corresponding feature labels generated by \LINE{} and DEXTER. Furthermore, we present extended comparisons of the neuron descriptions generated by \LINE{}, CLIP-Dissect, and INVERT for randomly selected neurons across different network architectures and layers in \Cref{fig:neuron_descriptions_dinov3,fig:neuron_descriptions_resnet18_imagenet,fig:neuron_descriptions_resnet50_imagenet,fig:neuron_descriptions_resnet18_places,fig:neuron_descriptions_resnet50_places}.

\begin{table}[h]
    \centering
    \caption{\textbf{Detailed feature description and label comparison on Salient ImageNet.} Expanding upon the visualizations presented in \Cref{fig:am_main}, we summarize the class and top feature descriptions while indicating whether the evaluated classes exhibit bias (\textcolor{ourred}{\ding{55}}) or not (\textcolor{ourgreen}{\ding{51}}). Alongside each evaluated feature, we report the labels produced by DEXTER (using the prompt \texttt{"a picture of a \{label\}"}) and \LINE{}. The generated labels confirm the visual explanations provided in \Cref{fig:am_main}, and notably, descriptions from both methods exhibit strong alignment for the non-biased classes.}
    \label{tab:am}
    \begin{tabular}{l c c l l}
    \toprule
    \textbf{Class (idx)} & \textbf{Non-Bias} & \textbf{\hspace{10pt}\# Feature\hspace{10pt}} & \textbf{DEXTER\hspace{20pt}} & \textbf{LINE} \\
    \midrule
    \multirow{5}{*}{Jeep (609)} & \multirow{5}{*}{\textcolor{ourgreen}{\ding{51}}} & 1067 & tractor & golfcart \\
     & & 1100 & truck & european fire salamander \\
     & & 1208 & tractor & tractor \\
     & & 1515 & truck & pickup \\
     & & 691  & ford & limousine \\
    \hdashline
    \multirow{5}{*}{Daisy (985)} & \multirow{5}{*}{\textcolor{ourgreen}{\ding{51}}} & 1105 & flower & sulphur-crested cockatoo \\
     & & 120  & daisy & ground beetle \\
     & & 298  & daisy & spiny \\
     & & 595  & daisy & daisy \\
     & & 859  & daisy & tick spider \\
    \hdashline
    \multirow{5}{*}{Rifle (764)} & \multirow{5}{*}{\textcolor{ourgreen}{\ding{51}}} & 1259 & pistol & revolver \\
     & & 1928 & gun & tripod \\
     & & 400  & rifle & grand piano \\
     & & 515  & gun & gun accessory \\
     & & 522  & pistol & revolver \\
    \hdashline
    \multirow{5}{*}{Patio (706)} & \multirow{5}{*}{\textcolor{ourred}{\ding{55}}} & 1016 & porch & sliding door \\
     & & 1633 & chair & oxcart \\
     & & 194  & window & ipod \\
     & & 451  & residence & golfcart \\
     & & 654  & room & barber chair \\
    \hdashline
    \multirow{5}{*}{Bathing Cap (433)} & \multirow{5}{*}{\textcolor{ourred}{\ding{55}}} & 121  & child & head \\
     & & 1340 & shoulder & swimsuit \\
     & & 1591 & look & macaque \\
     & & 1609 & baby & shower cap \\
     & & 755  & population & graduation attire \\
    \hdashline
    \multirow{5}{*}{Seat Belt (785)} & \multirow{5}{*}{\textcolor{ourred}{\ding{55}}} & 1010 & leg & scooter mower \\
     & & 108  & face & eye coverings \\
     & & 116  & field & car mirror \\
     & & 1493 & railway & car mirror \\
     & & 50   & park & car mirror \\
    \bottomrule
    \end{tabular}
\end{table}

\clearpage
\subsection{Evaluations on Lower Layers}
\label{app:lower_evaluations}

\begin{figure}[H]
    \centering
    \includegraphics[width=0.83\linewidth]{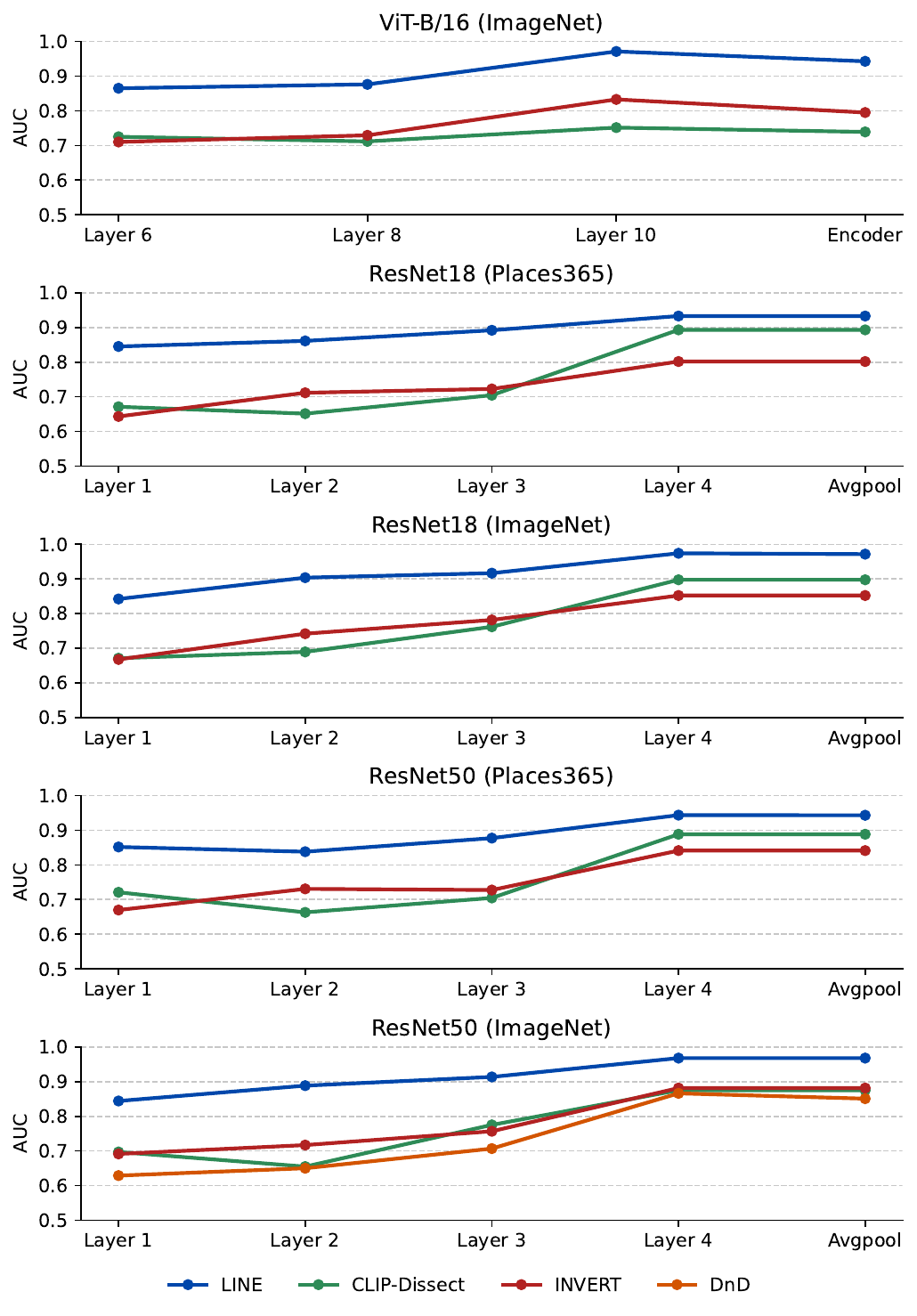}
    \caption{\textbf{Comparison of explanation methods on lower layers.} We evaluate CoSy AUC scores for \LINE{}, CLIP-Dissect, INVERT, and DnD across five models and their respective layers from shallow \texttt{Layer 1} to deep \texttt{Avgpool/Encoder}, averaged over up to 100 neurons per layer. Despite employing image cropping to better capture low-level features, DnD consistently underperforms. Notably, \LINE{} is the only method that maintains an AUC above 0.8 across all evaluated layers, including the earliest ones.}
    \label{fig:lineplot_lower}
\end{figure}

\begin{figure}[H]
    \centering
    \includegraphics[width=1.0\linewidth]{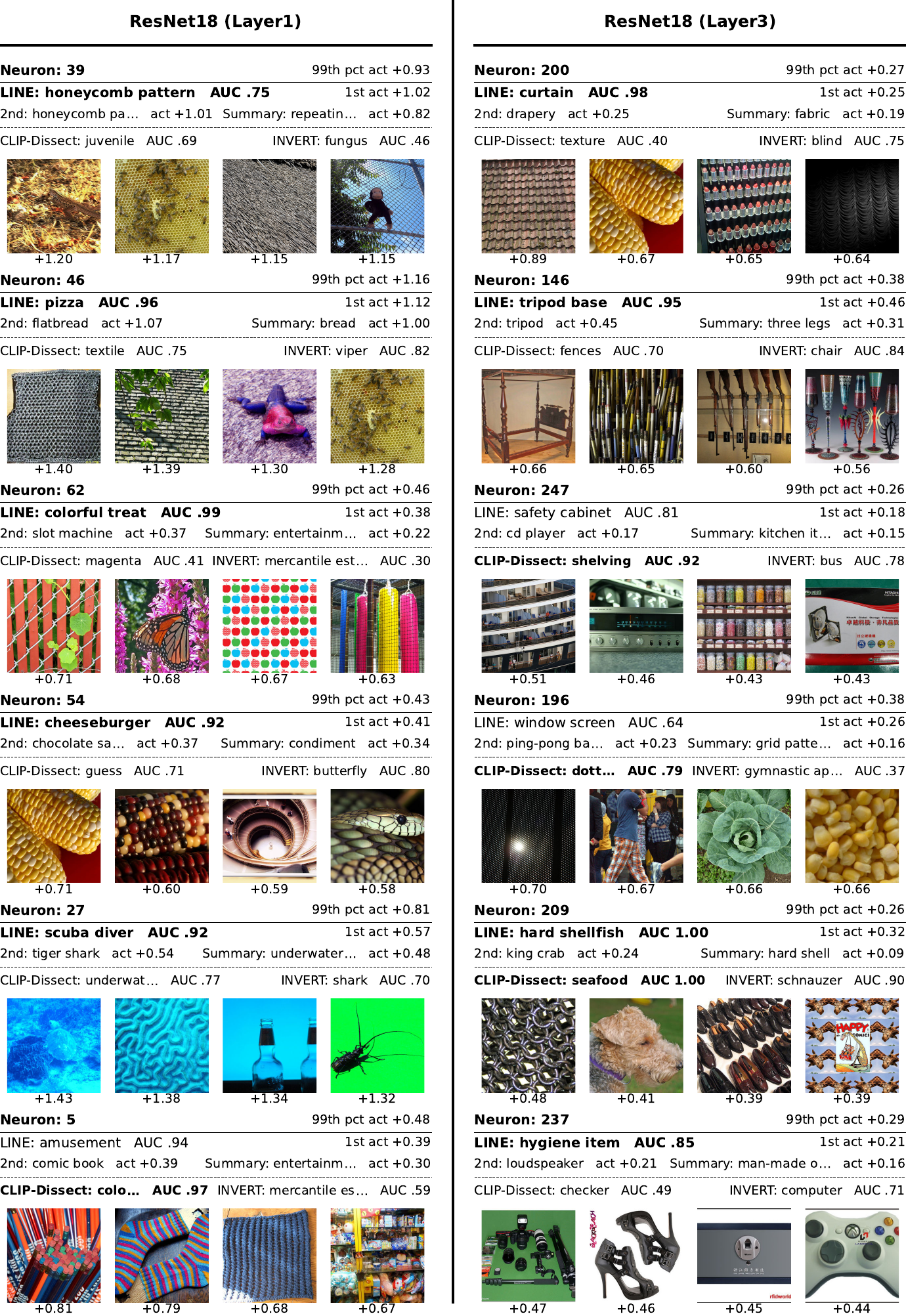}
    \caption{\textbf{Qualitative comparison of neuron descriptions in ImageNet ResNet18.} We extend the qualitative analysis from \Cref{fig:main_qualitive} and \Cref{app:extended_visual} on the lower layers of ImageNet ResNet18 \texttt{layer1} and \texttt{layer3}. For each neuron, we show the top four activating images from ImageNet-1K, the two highest-scoring \LINE{} descriptions together with the summary description and their activation scores, and the descriptions produced by CLIP-Dissect and INVERT, each reported with its CoSy AUC score. Every neuron is additionally annotated with its 99th-percentile activation over the dataset. The best-performing method for each neuron, based on the AUC score, is highlighted in \textbf{bold}.}
    \label{fig:neuron_descriptions_resnet18_imagenet_lower}
\end{figure}

\begin{figure}[H]
    \centering
    \includegraphics[width=1.0\linewidth]{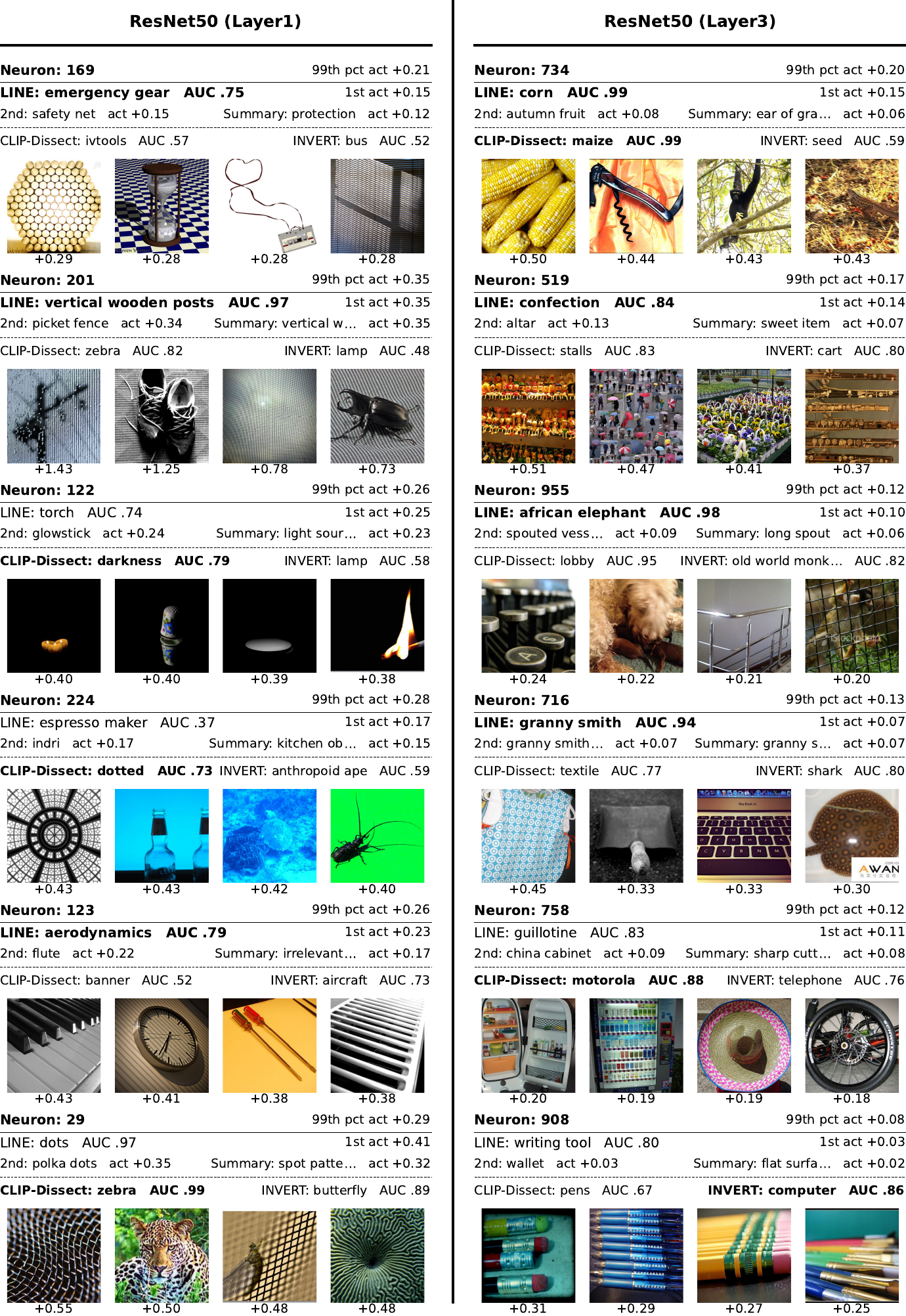}
    \caption{\textbf{Qualitative comparison of neuron descriptions in ImageNet ResNet50.} The best-performing method for each neuron, based on the AUC score, is highlighted in \textbf{bold}.}\label{fig:neuron_descriptions_resnet50_imagenet_lower}
\end{figure}

\begin{figure}[H]
    \centering
    \includegraphics[width=1.0\linewidth]{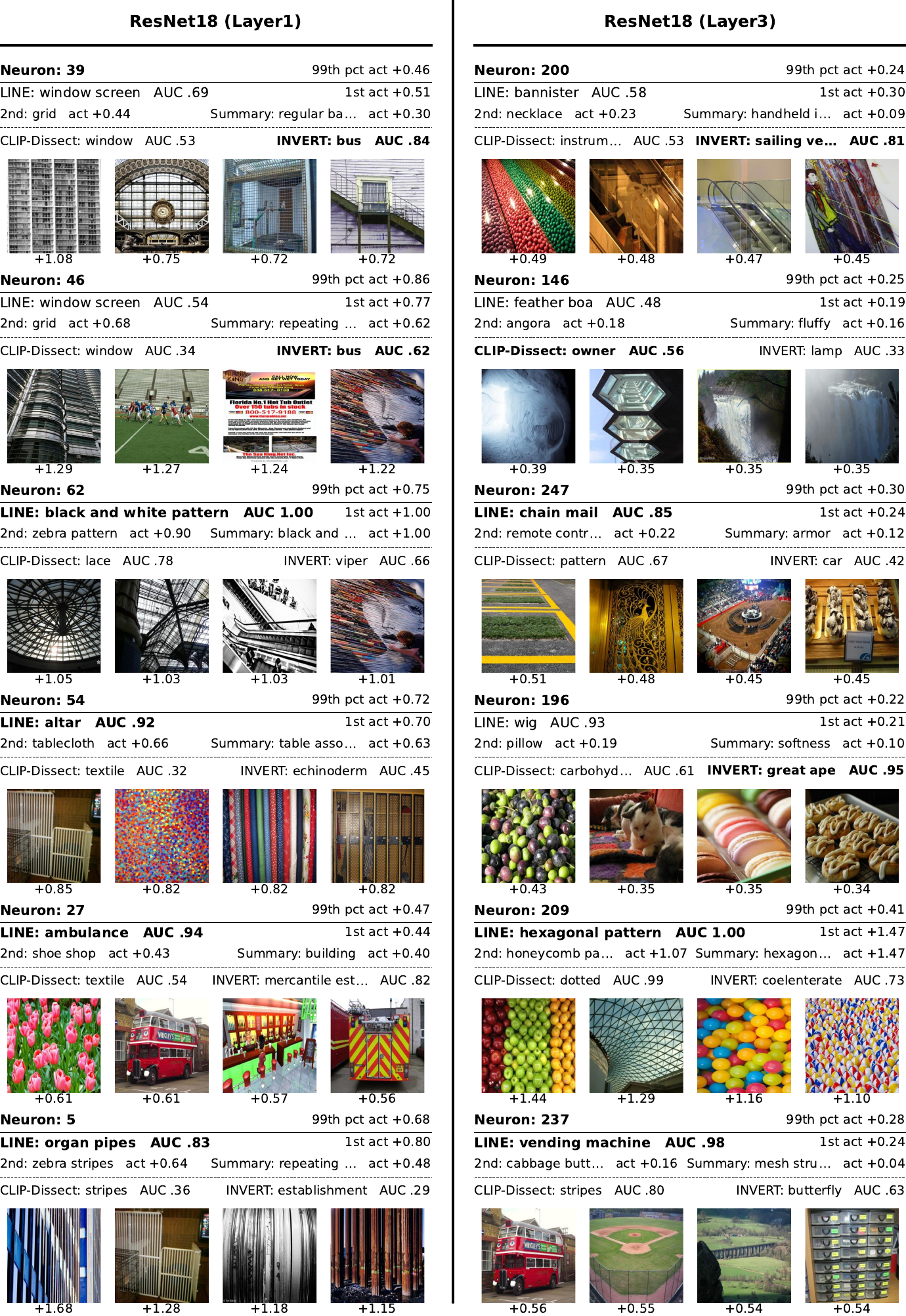}
    \caption{\textbf{Qualitative comparison of neuron descriptions in Places365 ResNet18.} The best-performing method for each neuron, based on the AUC score, is highlighted in \textbf{bold}.}\label{fig:neuron_descriptions_resnet18_places_lower}
\end{figure}

\begin{figure}[H]
    \centering
    \includegraphics[width=1.0\linewidth]{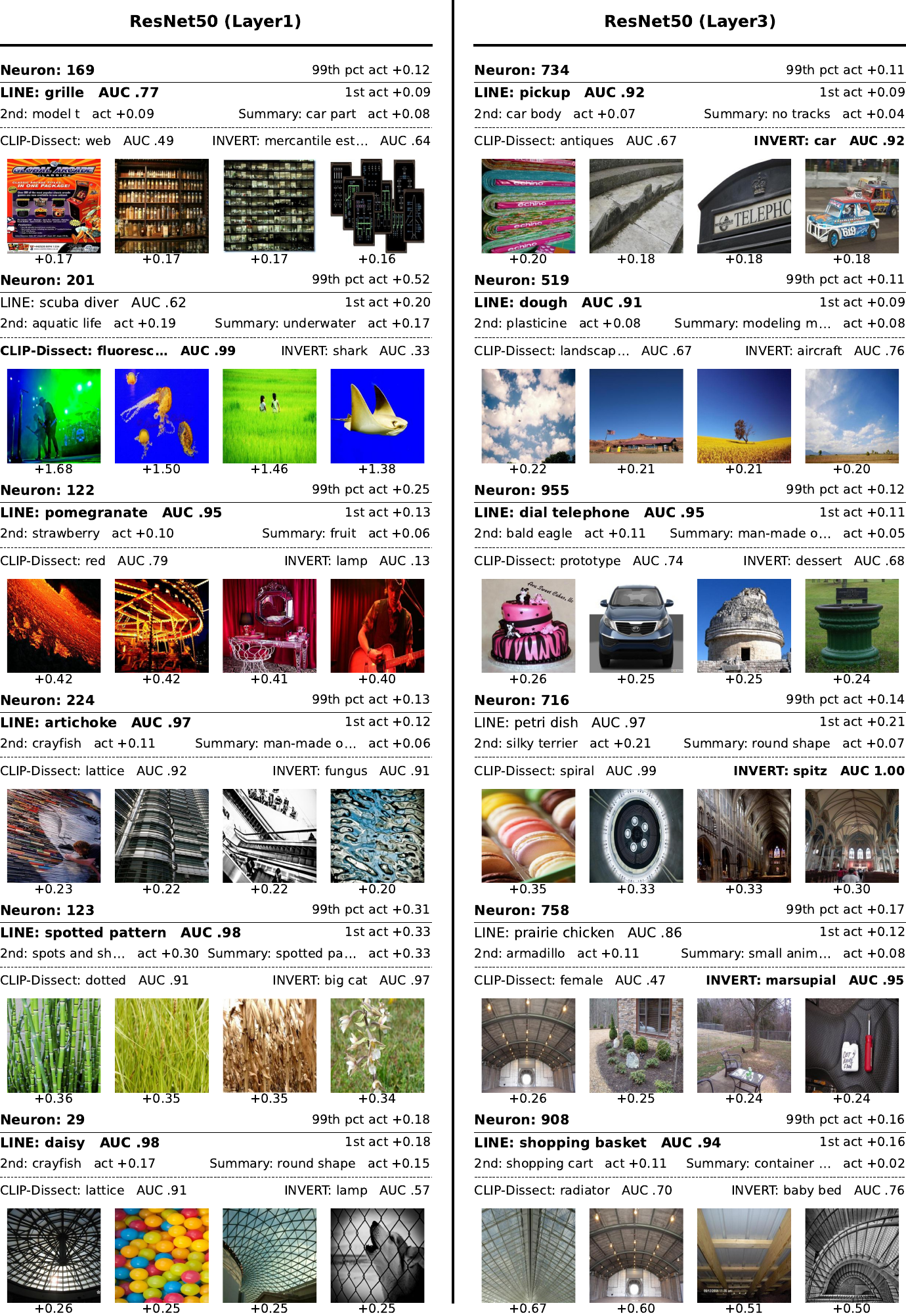}
    \caption{\textbf{Qualitative comparison of neuron descriptions in Places365 ResNet50.} The best-performing method for each neuron, based on the AUC score, is highlighted in \textbf{bold}.}\label{fig:neuron_descriptions_resnet50_places_lower}
\end{figure}

\begin{figure}[H]
    \centering
    \includegraphics[width=1.0\linewidth]{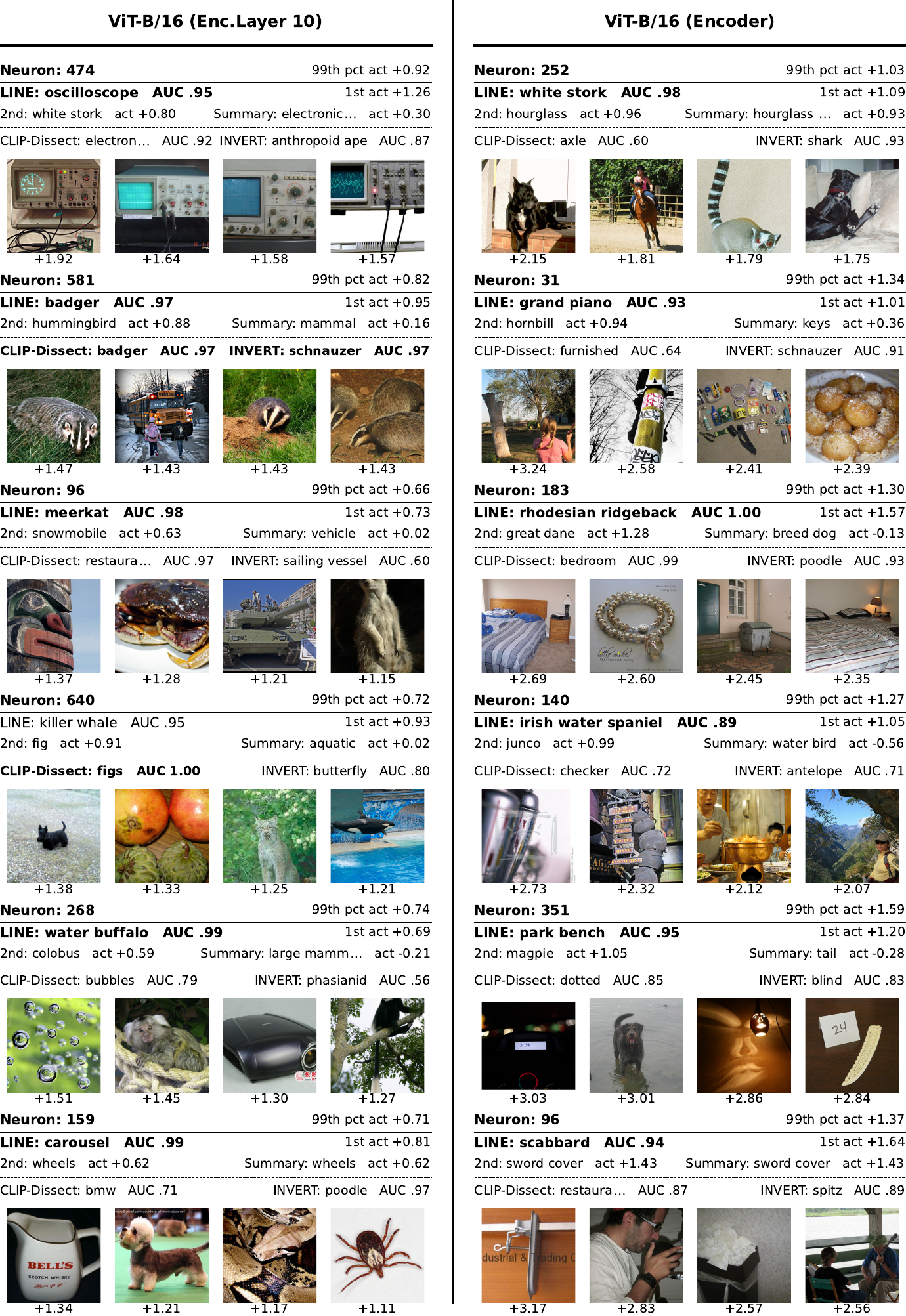}
    \caption{\textbf{Qualitative comparison of neuron descriptions in ViT-B/16.} The best-performing method for each neuron, based on the AUC score, is highlighted in \textbf{bold}.}
    \label{fig:neuron_descriptions_vit_lower}
\end{figure}

Early layers predominantly encode low-level visual features, such as colors and textures, as illustrated in \Cref{fig:neuron_descriptions_resnet18_imagenet_lower,fig:neuron_descriptions_resnet50_imagenet_lower,fig:neuron_descriptions_resnet18_places_lower,fig:neuron_descriptions_resnet50_places_lower,fig:neuron_descriptions_vit_lower}. Since \LINE{}'s scoreboard is initialized with high-level ImageNet class labels, it may occasionally perform worse than methods that are explicitly equipped with low-level concept vocabularies when interpreting neurons. Nonetheless, \Cref{fig:lineplot_lower} demonstrates that \LINE{} consistently achieves higher average AUC scores, suggesting that its iterative refinement mechanism compensates for suboptimal initialization and generalizes robustly across diverse architectures and layers.

\clearpage
\section{Details of the Experimental Setting and Reproducibility}
\label{app:reproducibility}

This section contains additional information regarding reproducibility and the models and method setup used in the experiments.

\subsection{Models and Datasets}

\textbf{Models.}
We use the following openly available vision models~\citep{he2015deep, dosovitskiy2020image} from PyTorch's \texttt{torchvision} and Hugging Face with their default hyperparameters:
\begin{itemize}
    \item ResNet-18 (Places365)~\citep{zhou2017places}: \texttt{torchvision} ResNet18 with Places365-pretrained weights (BSD-3-Clause and CC BY license) from \url{http://places2.csail.mit.edu/models_places365/resnet18_places365.pth.tar}.
    \item ResNet-18 (ImageNet)~\citep{ILSVRC15}: \texttt{torchvision} ResNet18 with \texttt{IMAGENET1K\_V1} weights (BSD-3-Clause).
    \item ResNet-50 (Places365)~\citep{zhou2017places}: \texttt{torchvision} ResNet50 with Places365-pretrained weights (BSD-3-Clause and CC BY) from \url{http://places2.csail.mit.edu/models_places365/resnet50_places365.pth.tar}.
    \item ResNet-50 (ImageNet)~\citep{ILSVRC15}: \texttt{torchvision} ResNet50 with \texttt{IMAGENET1K\_V1} weights (BSD-3-Clause).
    \item ResNet-50 (Robust)~\citep{huang2023revisiting}: \texttt{torchvision} ResNet50 with adversarially robust weights from \url{https://www.dropbox.com/s/knf4uimlqsi1yz8/imagenet_l2_3_0.pt?dl=1} (MIT License).
    \item ViT-B/16 (ImageNet)~\citep{ILSVRC15}: \texttt{torchvision} ViT\_B\_16 with \texttt{IMAGENET1K\_V1} weights (BSD-3-Clause).
    \item DINOv3 ViT-B/16~\citep{simeoni2025cijo}: \\ \texttt{facebook/dinov3-vitb16-pretrain-lvd1689m} trained on non-public dataset LVD-1689M(DINOv3 License).
\end{itemize}

For our generative pipeline experiments, we employ the following models:
\begin{itemize}
    \item Llama-3.1-8B-Instruct~\citep{grattafiori2024llama}: \\ \texttt{meta-llama/Llama-3.1-8B-Instruct} (Llama 3.1 Community License).
    \item Stable Diffusion v1.5~\citep{Rombach_2022_CVPR}: \\ \texttt{stable-diffusion-v1-5/stable-diffusion-v1-5} \\ (CreativeML Open RAIL-M License).
    \item Stable Diffusion XL~\citep{podell2023sdxl}: \\ \texttt{stabilityai/stable-diffusion-xl-base-1.0} \\ (CreativeML Open RAIL++-M License).
    \item FLUX.1-dev~\citep{flux_dev_2024}: \texttt{black-forest-labs/FLUX.1-dev} \\ (FLUX.1-dev Non-Commercial License).
\end{itemize}
For image-to-image causal ablations we use Qwen-Image-Edit~\citep{wu2025qwenimagetechnicalreport}:
\begin{itemize}
\item \texttt{Qwen/Qwen-Image-Edit-2511} (Apache 2.0 License)
\end{itemize}
For the CoSy benchmark, we use Stable Diffusion XL~\citep{podell2023sdxl}, and for VLM correlation evaluation, SigLIP-SO400M-14-384~\citep{zhai2023sigmoid}:
\begin{itemize}
\item \texttt{google/\allowbreak siglip-so400m-patch14-384} (Apache~2.0 License)
\end{itemize}

\textbf{Datasets.}
All evaluations in our experiments are conducted using the
\begin{itemize}
    \item ImageNet~\citep{ILSVRC15}: from Hugging Face \texttt{ILSVRC/imagenet-1k} \\ (ImageNet Agreement)
    \item Places365~\citep{zhou2017places}: from the official webpage \\ \url{http://places2.csail.mit.edu/download.html} (MIT Licence)
\end{itemize}

\subsection{Reproducibility}

The complete codebase is provided in the {\hypersetup{allcolors=ourlink}{\url{https://github.com/CredibleAI/LINE}}, accompanied by a comprehensive \texttt{README.md} to ensure full reproducibility of all \LINE{} results reported in this paper. T2I sampling is seeded, guaranteeing that a given concept always produces the same generated images and, consequently, identical average activations. For the LLM component, although greedy decoding is supported, we opt for sampling-based decoding, as greedy decoding tends to cause the model to repetitively generate the same concepts. We note that because concepts are processed in batches of 50 during neuron description gathering, a different ordering of sampled concepts at inference time may yield marginally different results, owing solely to the stochastic nature of LLM sampling. Furthermore, a sequence of CoSy evaluations may not be fully reproducible, as it creates images for concepts, skipping already existing ones and not reseeding the generator between them. Changing the order of the provided concepts or already generated concepts from other methods will result in different generated images and scores.

\subsection{Compute resources}
\label{app:computationl}

The \texttt{README.md} specifies a minimum of 16~GB VRAM and 32~GB RAM for processing a single neuron. For efficiency, we processed neurons in batches of 50; however, for some models it exceeded the VRAM capacity of our A100 GPUs, requiring them to be run on H100 GPUs instead. \Cref{tab:timing_stats} reports the average time \LINE{} requires to generate a neuron's final scoreboard, normalized per neuron.
\Cref{tab:time_breakdown} breaks this time down across the pipeline components for both ResNet models on the \texttt{avgpool} layer, measured on identical H100 hardware. Image synthesis dominates the total cost, while inference on the explained model is negligible.

\begin{table}[h]
    \centering
    \caption{\textbf{Timing statistics for \LINE{} across models and datasets.} Runtimes are normalized per neuron due to batched processing. DINOv3 was run exclusively on A100 GPUs; all other models had some layers processed on H100 GPUs, which accounts for their comparatively lower average runtimes. ResNet50 ImageNet timings are omitted as they were not measured under comparable conditions.}
    \label{tab:timing_stats}
    \vspace{0.5em}
    \begin{tabular}{l l c c c}
    \toprule
    \textbf{Model} & \textbf{Dataset} & \textbf{\# Neurons} & \textbf{Time/Neuron (s)} & \textbf{Total Time (h)} \\
    \midrule
    DINOv3 & LVD-1689M & $100$ & $534.09$ & $14.84$ \\
    \midrule
    \multirow{2}{*}{ResNet18} & Places365 & $564$ & $500.60$ & $78.43$ \\
     & ImageNet & $564$ & $521.39$ & $81.68$ \\
    \midrule
    ResNet50 & Places365 & $600$ & $410.64$ & $68.44$ \\
    \midrule
    ViT-B/16 & ImageNet & $600$ & $380.91$ & $63.49$ \\
    \bottomrule
    \end{tabular}
\end{table}

\begin{table}[h]
    \centering
    \caption{\textbf{Execution time breakdown.} Absolute and relative runtime of each time significant stage of the \LINE{} pipeline, evaluated on the \texttt{avgpool} layer of ResNet18 and ResNet50 for 50 neurons on the H100.}
    \label{tab:time_breakdown}
    \vspace{0.5em}
    \begin{tabular}{l c c c c}
    \toprule
    \multirow{2}{*}{\textbf{Stage}} & \multicolumn{2}{c}{\textbf{ResNet18}} & \multicolumn{2}{c}{\textbf{ResNet50}} \\
    \cmidrule(lr){2-3} \cmidrule(lr){4-5}
     & \textbf{Time (h)} & \textbf{\%} & \textbf{Time (h)} & \textbf{\%} \\
    \midrule
    Image Synthesis (T2I image generation) & 6.39 & 88.0\% & 6.34 & 88.0\% \\
    Concept Proposal (LLM concept gen) & 0.60 & 8.3\% & 0.59 & 8.2\% \\
    LLM model load & 0.14 & 1.9\% & 0.14 & 1.9\% \\
    T2I model load & 0.11 & 1.5\% & 0.11 & 1.5\% \\
    Activation Extraction & 0.02 & 0.3\% & 0.03 & 0.4\% \\
    \midrule
    \textbf{Total} & \textbf{7.26} & \textbf{100\%} & \textbf{7.20} & \textbf{100\%} \\
    \bottomrule
    \end{tabular}
\end{table}

\clearpage
\section{Broader Impact}
\label{app:broader_impact}

\LINE{} enables training-free, gray-box interpretability audits of vision models without requiring access to internal gradients or training data. However, relying on automated labeling introduces the risk of an \emph{interpretability illusion}. Specifically, compressing complex, polysemantic neural behavior~\citep{bricken2023monosemanticity} into a single human-readable label can give auditors a false sense of confidence. Furthermore, \LINE{} relies on an ImageNet-based initialization that biases the concept discovery process. The initial bias can lead to the under-representation of demographic minorities and highly specialized visual domains. \LINE{} also inherits the biases of its generative components. For example, when a T2I model fails to faithfully render a proposed concept, the evaluation pipeline might downrank it, skewing the final explanations and introducing fairness concerns.

Beyond intrinsic biases, the framework introduces external risks: adversaries could exploit \LINE{} to map sensitive neurons for targeted attacks~\citep{nanfack2024adversarial}, and the generation of synthetic images carries inherent disinformation risks. To mitigate these threats, we release \LINE{} as open-source and recommend that auditors report full scoreboards rather than relying solely on the top concept. Additionally, in fairness-sensitive or safety-critical contexts, automated labels should be paired with causal ablations, human review, and strict T2I safety filters.

\newpage
\section*{NeurIPS Paper Checklist}

\begin{enumerate}

\item {\bf Claims}
    \item[] Question: Do the main claims made in the abstract and introduction accurately reflect the paper's contributions and scope?
    \item[] Answer: \answerYes{}
    \item[] Justification: The contributions outlined in the abstract and introduction accurately reflect the results detailed in \Cref{sec:qualitative,sec:quantitative,sec:visual_explanations}.

\item {\bf Limitations}
    \item[] Question: Does the paper discuss the limitations of the work performed by the authors?
    \item[] Answer: \answerYes{}
    \item[] Justification: In \Cref{sec:conclusion}.

\item {\bf Theory assumptions and proofs}
    \item[] Question: For each theoretical result, does the paper provide the full set of assumptions and a complete (and correct) proof?
    \item[] Answer: \answerNA{}
    \item[] Justification: Not applicable, as this paper does not present theoretical results or proofs.

    \item {\bf Experimental result reproducibility}
    \item[] Question: Does the paper fully disclose all the information needed to reproduce the main experimental results of the paper to the extent that it affects the main claims and/or conclusions of the paper (regardless of whether the code and data are provided or not)?
    \item[] Answer: \answerYes{}.
    \item[] Justification: As stated in Appendix~\ref{app:reproducibility}, the code used in this work is available in {\hypersetup{allcolors=ourlink}{\url{https://github.com/CredibleAI/LINE}}}. All models and datasets used are openly available.

\item {\bf Open access to data and code}
    \item[] Question: Does the paper provide open access to the data and code, with sufficient instructions to faithfully reproduce the main experimental results, as described in supplemental material?
    \item[] Answer: \answerYes{}
    \item[] Justification: The code is available in {\hypersetup{allcolors=ourlink}{\url{https://github.com/CredibleAI/LINE}}}. Models and datasets are openly available.

\item {\bf Experimental setting/details}
    \item[] Question: Does the paper specify all the training and test details (e.g., data splits, hyperparameters, how they were chosen, type of optimizer) necessary to understand the results?
    Answer: \answerYes{}
    \item[] Justification: Detailed experimental setups and hyperparameter configurations are described in \Cref{sec:method}, with full prompt specifications provided in Appendix~\ref{app:prompts} and models/dataset setups in Appendix~\ref{app:reproducibility}.

\item {\bf Experiment statistical significance}
    \item[] Question: Does the paper report error bars suitably and correctly defined or other appropriate information about the statistical significance of the experiments?
    \item[] Answer: \answerYes{}
    \item[] Justification: In Appendix~\ref{app:statistical}, we provide statistical testing to support our claims.

\item {\bf Experiments compute resources}
    \item[] Question: For each experiment, does the paper provide sufficient information on the computer resources (type of compute workers, memory, time of execution) needed to reproduce the experiments?
    \item[] Answer: \answerYes{}
    \item[] Justification: In Appendix~\ref{app:reproducibility}.

\item {\bf Code of ethics}
    \item[] Question: Does the research conducted in the paper conform, in every respect, with the NeurIPS Code of Ethics \url{https://neurips.cc/public/EthicsGuidelines}?
    \item[] Answer: \answerYes{}
    \item[] Justification: We conform to the NeurIPS Code of Ethics.

\item {\bf Broader impacts}
    \item[] Question: Does the paper discuss both potential positive societal impacts and negative societal impacts of the work performed?
    \item[] Answer: \answerYes{}
    \item[] Justification: In Appendix~\ref{app:broader_impact}.

\item {\bf Safeguards}
    \item[] Question: Does the paper describe safeguards that have been put in place for responsible release of data or models that have a high risk for misuse (e.g., pre-trained language models, image generators, or scraped datasets)?
    \item[] Answer: \answerNA{}
    \item[] Justification: We do not release data or models.

\item {\bf Licenses for existing assets}
    \item[] Question: Are the creators or original owners of assets (e.g., code, data, models), used in the paper, properly credited and are the license and terms of use explicitly mentioned and properly respected?
    \item[] Answer: \answerYes{}
    \item[] Justification: In the codebase, we provide the MIT License, and for used models and datasets we list licenses in \Cref{app:reproducibility}.

\item {\bf New assets}
    \item[] Question: Are new assets introduced in the paper well documented and is the documentation provided alongside the assets?
    \item[] Answer: \answerYes{}
    \item[] Justification: The repository contains full documentation in the README file.

\item {\bf Crowdsourcing and research with human subjects}
    \item[] Question: For crowdsourcing experiments and research with human subjects, does the paper include the full text of instructions given to participants and screenshots, if applicable, as well as details about compensation (if any)?
    \item[] Answer: \answerNA{}
    \item[] Justification: Crowdsourcing was not used in this work.

\item {\bf Institutional review board (IRB) approvals or equivalent for research with human subjects}
    \item[] Question: Does the paper describe potential risks incurred by study participants, whether such risks were disclosed to the subjects, and whether Institutional Review Board (IRB) approvals (or an equivalent approval/review based on the requirements of your country or institution) were obtained?
    \item[] Answer: \answerNA{}
    \item[] Justification: Crowdsourcing was not used in this work.

\item {\bf Declaration of LLM usage}
    \item[] Question: Does the paper describe the usage of LLMs if it is an important, original, or non-standard component of the core methods in this research? Note that if the LLM is used only for writing, editing, or formatting purposes and does \emph{not} impact the core methodology, scientific rigor, or originality of the research, declaration is not required.
    \item[] Answer: \answerYes{}
    \item[]  Justification: LLMs are a core, non-standard component of the proposed \LINE{} method, where they are used in an iterative loop to propose new concept descriptions conditioned on a scoreboard of previously evaluated concepts. The specific LLM (Llama 3.1 8B) and its role are described in Section~\ref{sec:method}, the experimental setup in Section~\ref{sec:experiments}, and the full prompts are reported in Appendix~\ref{app:prompts}. LLMs were not used for writing or editing the manuscript in any way that affects methodology, scientific rigor, or originality, and were used strictly as productivity tools for code verification, completion, and grammatical refinement.

\end{enumerate}

\end{document}